\PassOptionsToPackage{prologue,dvipsnames}{xcolor}
\documentclass{article}
\usepackage{lmodern}
\usepackage{amsmath,amssymb,amsthm}
\usepackage{booktabs}
\usepackage{threeparttable}
\usepackage{multirow}
\usepackage{arydshln}
\usepackage{graphicx}
\usepackage[font=small]{caption}
\usepackage{subcaption}
\usepackage{adjustbox}
\usepackage{float}
\usepackage{placeins}
\usepackage{longtable}
\usepackage{makecell}
\usepackage[table]{xcolor}
\definecolor{rowblue}{RGB}{220,235,255}
\usepackage{etoolbox}
\usepackage{pifont}
\usepackage{comment}
\usepackage{tcolorbox}
\newcommand{\pkg}[1]{\texttt{#1}}
\usepackage[a-1b]{pdfx}
\hypersetup{hidelinks}
\usepackage{xr-hyper}
\usepackage[authoryear,round]{natbib}
\usepackage{xltabular} 
\usepackage{threeparttablex}
\usepackage{titlesec}
\usepackage{array}
\usepackage{ragged2e}
\titleformat{\section}{\normalfont\large\bfseries}{\thesection}{1em}{}
\titleformat{\subsection}{\normalfont\normalsize\bfseries}{\thesubsection}{1em}{}
\titleformat{\subsubsection}{\normalfont\small\bfseries}{\thesubsubsection}{1em}{}
\usepackage{arxiv}
\usepackage{makecell}

\usepackage{booktabs,array}

\newcommand{\bego}[1]{\textcolor{blue}{#1}}

\title{A reproducible and extensible framework for benchmarking competing risks survival models}

\author{
Bego\~na B. Sierra \thanks{To whom correspondence should be addressed}\\
Cancer Research UK Scotland Centre \\
Institute of Genetics and Cancer \\
University of Edinburgh \\
\texttt{B.Bolos@ed.ac.uk} 
\And
Colin McLean \\
Cancer Research UK Scotland Centre \\
Institute of Genetics and Cancer \\
University of Edinburgh \\
\texttt{Colin.D.McLean@ed.ac.uk} 
\And
Peter S. Hall \\
Cancer Research UK Scotland Centre \\
Institute of Genetics and Cancer \\
University of Edinburgh \\
\texttt{p.s.hall@ed.ac.uk} 
\And
Sarah Friedrich-Welz\\
Mathematical Statistics and AI in Medicine \\
University of Augsburg\\
\texttt{sarah.friedrich@math.uni-augsburg.de} 
\And
Catalina A. Vallejos $^*$\\
Institute of Genetics and Cancer \\
University of Edinburgh \\
\texttt{catalina.vallejos@ed.ac.uk}
}

\begin{document}

\maketitle


\begin{abstract}
{A wide range of statistical and machine learning methods have been proposed for survival analysis with competing risks, where the occurrence of one event (i.e., cancer death) precludes the occurrence of other events (i.e., cardiovascular disease death). 
Despite these methodological advances, their systematic evaluation and adoption are limited by the lack of comprehensive, reproducible and extensible benchmarking frameworks. We developed an open-source benchmarking framework for competing risks models that enables their systematic comparison across multiple datasets under different aspects of performance; calibration, discrimination, overall prediction error and clinical utility. We additionally introduce an extension of SHAP for competing risks, allowing model-agnostic interpretability of covariates contributions over time. All our code is publicly available via GitHub: \url{https://github.com/BBolosSierra/CompRisksBenchmark}.}
\end{abstract}

\section{Introduction}
\label{sec::intro}

In survival analysis, also known as time-to-event analysis, the primary goal is to model the time until an event occurs \citep{Kalbfleisch2002}. Early modelling strategies, such as the Cox Proportional Hazards (PH) model \citep{cox1972RSS}, were developed primarily for the estimation of covariate effects via hazard ratios. More recently, there has been a growing interest in using survival models in the context of risk prediction to estimate the probability of observing an adverse event within a pre-specified time window \citep{pfeiffer2017absolute}. This task is highly relevant in healthcare settings where patient-level estimation of risk for an adverse event (e.g.,~cancer death or relapse) can assist targeted and individualised clinical decision making. Widely used risk prediction models developed using a survival analysis framework include QRISK3 \citep{Hippisley-Coxj2099} and KFRE \citep{Tangri2011JAMA}.

Time-to-event data is often incomplete, as the exact event times are not always observed. This is known as censoring. Right-censoring occurs when only a lower bound for the event time is observed due to individual-level circumstances (e.g.,~loss to follow-up or withdrawn consent), or because the study follow-up ends before the event occurs (i.e.,~administrative censoring). 
Survival models explicitly account for censoring to avoid biased estimates of survival probabilities. 
Survival 
methods commonly assume non-informative censoring, meaning that conditional on observed covariates, censoring is unrelated with an individual's underlaying risk of experiencing the event. 

In healthcare contexts, individuals are often at risk of experiencing several and mutually exclusive events, known as competing risks (CR). For instance, a cancer patient can die from their disease progression 
or from other non-cancer causes such as cardiovascular events. 
Often practitioners choose an event of interest while treating subjects with CR events as censored. 
This violates the non-informative censoring assumption since those with a CR event are no longer at risk of the event of interest. 
Such treatment of CR events leads to 
an overestimation of risk for the event of interest \citep{Austin2016Circulation}, 
potentially translating into patient over-treatment or biased cost-effectiveness estimation of interventions in clinical trials \citep{Koller2012StatisticsInMedicine}. CR may also disproportionally affect specific population subgroups \citep[see e.g.~][]{BOEDDINGHAUS20262282}. In such cases, ignoring CR may lead to between-groups differences in predictive performance, with potential consequences in terms of algorithmic fairness \citep{jeanselme2025arxiv}. Appropriate handling of CR in the context of risk prediction is particularly critical 
in ageing populations with increasing prevalence of multimorbidity, which is a pressing challenge for healthcare systems worldwide \citep{WhittyBMJ2020}. Despite this, the presence of CR is largely ignored when developing risk prediction models. 

Numerous methods have been developed to account for CR \citep[see][for a review]{Monterrubio2024BiometricalJournal}. For example, the non-parametric Aalen-Johansen estimator \citep{aalenjohansen} and the Fine and Gray model 
\citep{FineGray1999JASA}. 
More recently, with the advancement of machine learning methods, flexible survival models have been proposed to capture more complex relationships between covariates and event risk; for example, via deep learning approaches \citep{Wiegrebe2024, Wang2019ACM}. Despite these methodological developments, modern CR approaches have not been widely adopted in practical applications. This gap may be attributed to lack of reproducible benchmarks, limited availability of open-source software, and challenges in terms of clinical interpretability \citep{Lillelund2025, Wiegrebe2024, Monterrubio2024BiometricalJournal}. 

First, studies introducing novel approaches carry out their own benchmarks, 
often using different datasets or unreported and non-reproducible preprocessing steps \citep{johnson17a}. Importantly, model evaluation is often inconsistent between articles, not only the range of metrics selected (often limited to discrimination), but also omitting critical details \citep[e.g.,~
whether it accounts for ties;][]{sierra2025cindexmultiverse}. In addition, as discussed in \cite{Lillelund2025}, authors' choices on evaluation metrics do not always align with the model objective; e.g.~assessing discrimination is often not the most critical aspect in clinical applications. 
As in other computational fields, this may lead to an over-optimism about new approaches, driven in part by selective evaluation and the scarcity of unbiased comparison studies which ultimately slows methodological consolidation and practical adoption \citep{Boulesteix2013PlosOne}. This has motivated 
recent work on 
benchmarking survival models in single risk settings \citep{burk2026largescaleneutralcomparisonstudy},  facilitated by software tools such as \pkg{mlr3proba} \citep{ml3proba}. In a CR context, instead, unbiased comparisons are more limited. Recent studies 
\citep{djangang2025comparativereviewmoderncompeting, Lee2026CSAM, Kantidakis2023} 
focus mainly on simulated data or single datasets, and open-source code that allows to reproduce results is often not available. This also precludes benchmarks to be extended, e.g.~to incorporate new methods or datasets. 

Secondly, novel methods are often implemented as ad-hoc code 
without integration into a formal software packages or broader machine learning frameworks for survival analysis, therefore limiting their accessibility and practical adoption. Although software packages such as \pkg{mlr3proba} \citep{ml3proba} or \pkg{pycox} \citep{pycox} include implementations for a wide range of survival methods, its extent is limited to mainly 
single risk cases, with only a few exceptions 
(e.g., \pkg{pycox} recently including DeepHit's CR extension \cite{DeepHit2018AAAI}).



Third, beyond regression approaches, which provide a direct parametrisation of covariate effects through coefficients and hazard ratios, clinical interpretability is less straightforward in other cases. 
Indeed, machine learning methods typically do not directly yield 
covariate effect estimates, and 
post-hoc steps are required to facilitate interpretability. 


To 
address these challenges,
we present a reproducible, reusable and extendable benchmark pipeline to evaluate the performance of CR approaches, focusing on risk prediction tasks. This is based on principles stated in \cite{Boulesteix2013PlosOne}. 
We consider real-world datasets with varying sample sizes and event rates,   
and methods ranging from classical survival models to deep learning-based approaches for which implementations are available in R \citep{Rsoftware} or python \citep{pythonsoftware}. We perform a comprehensive evaluation that considers several aspects of predictive performance, beyond solely assessing discrimination; whilst also considering the effect of hyper-parameter choices and interpretability. 
In addition, we introduce a new interpretability approach which extends SurvSHAP$(t)$ \citep{KRZYZINSKI2023KBS} to CR settings. 
We anticipate that our benchmarking pipeline will be a useful resource 
for those applying CR methods, facilitating model development and 
validation 
based on their own datasets. Our pipeline also aims to support developers, providing a reproducible framework to evaluate the performance of new approaches. 
All the 
results provided in this work can be reproduced from the following GitHub repository \url{https://github.com/BBolosSierra/CompRisksBenchmark}. To facilitate reproducibility, the repository also includes a Docker environment that defines the required computational environment and dependencies. 

\newpage

\section{Background} 
\label{sec::background}


Let $T \geq 0$ be a continuous random variable representing the time until an event of interest occurs. The distribution of $T$ can be defined via the survival function $S(t) =\text{Pr}(T > t)$, i.e.~the probability that the event has not yet occurred by time $t$. 
Alternatively, a survival model can also be specified by the corresponding hazard function: 
\begin{equation}\label{eq:overallhazard} 
h(t) = \lim_{\Delta t \to 0} 
\frac{P(t \le T < t + \Delta t \mid T \ge t)}{\Delta t} = \frac{f(t)}{S(t)},
\end{equation} where $f(t) = -\frac{dS(t)}{dt}$ is the density function. The hazard function represents the instantaneous risk of experiencing the event at time $t$, conditional on survival up to $t$. 
The relationship between $S(t)$ and $h(t)$ can also be written as: 

\begin{equation}\label{eq:relation-hazard-surv} 
S(t) =  1 - F(t) = \exp\!\left[-\int_{0}^{t} h(u)\,du\right]. 
\end{equation}

In the presence of CR, where $K > 1$ mutually exclusive event types can occur, the definitions above can be extended. Let $Z \in \{1,2, ... ,K\}$ indicate which of the $K$ possible event types occurred, and let $Z = 0$ denote censoring. For each cause $k$, we can define the cause-specific (CS) hazard function:


\begin{equation}\label{eq:kspecific-hazard} 
h_k(t) 
= \lim_{\Delta t \to 0} 
\frac{P(t \le T < t + \Delta t,\, Z = k \mid T \ge t)}{\Delta t},
\end{equation}

where the overall hazards in \eqref{eq:overallhazard} is the sum across all the CS hazards $h(t) = \sum_{k=1}^{K} h_k(t)$. In a risk prediction context, the quantity of interest is typically given by the cumulative incidence function (CIF) for cause $k$: 

\begin{equation}\label{eq:CIF_k} 
    F_k(t) = P(T \le t, Z = k), 
\end{equation}

that is, the probability of observing the event $k$ before time $t$ without any other competing event occurring earlier. The CIF relates to the cause-specific hazard as follows:

\begin{equation}
\label{eq:CIF_k2}
F_k(t) = \int_{0}^{t} S(u^{-}) h_{k}(u) \ du,
\quad
\text{where} \quad
S(t) =\exp\!\left[-\sum_{k=1}^{K}\int_{0}^{t}  h_{k}(u)\, du \right]. 
\end{equation}


Whilst there is a one-to-one relationship between $h(t)$ and $S(t)$ for a single event type (see \eqref{eq:relation-hazard-surv}), the latter does not hold for the cause-specific counterparts when $K>1$. 
For example, consider $K = 2$, where cancer death is denoted by $Z=1$ and cardiovascular death as $Z=2$. The CIF for cancer deaths can be written as:

$$F_{1}(t)
= P(T \le t; Z = 1)
= \int_{0}^{t} h_{1}(u)\, S(u^{-})\, du, \quad \text{where the event-free survival from any cause is} $$
\begin{equation}
\label{eq:CIF2}
S(u)
= \exp\!\left(
-\int_{0}^{u}\Big(h_{1}(s) + h_{2}(s)\Big) ds
\right). %
\end{equation}

$F_{1}(t)$ depends not only on the cancer-specific hazard function $h_1(t)$ but also on the competing event hazard $h_2(t)$ through the survival function $S(t)$ --- only those who have not experienced any event are still at risk by time $t$. 
Suppose there is a hypothetical cancer therapy that increases the risk of cardiovascular death (higher $h_2(t)$) while not reducing cancer-specific risk ($h_1(t)$ remains unchanged). Then, $S(t)$ would decrease as more individuals experience the competing event earlier. Consequently, the probability of observing a cancer death by time $t$ ($F_1(t)$) would decrease, even though the instantaneous cancer hazard ($h_1(t)$) itself has not changed. In reality, cancer therapy could have an effect on both hazards (e.g., cancer treatment lowers $h_1(t)$ but increases $h_2(t)$). In such cases, $F_1(t)$ would decrease or increase depending on the strength of those effects. This illustrative example shows how ignoring competing events could distort our interpretation about the effectiveness of the treatment. 


\section{Methods}

\begin{figure}
    \centering
    \includegraphics[width=0.85\textwidth]{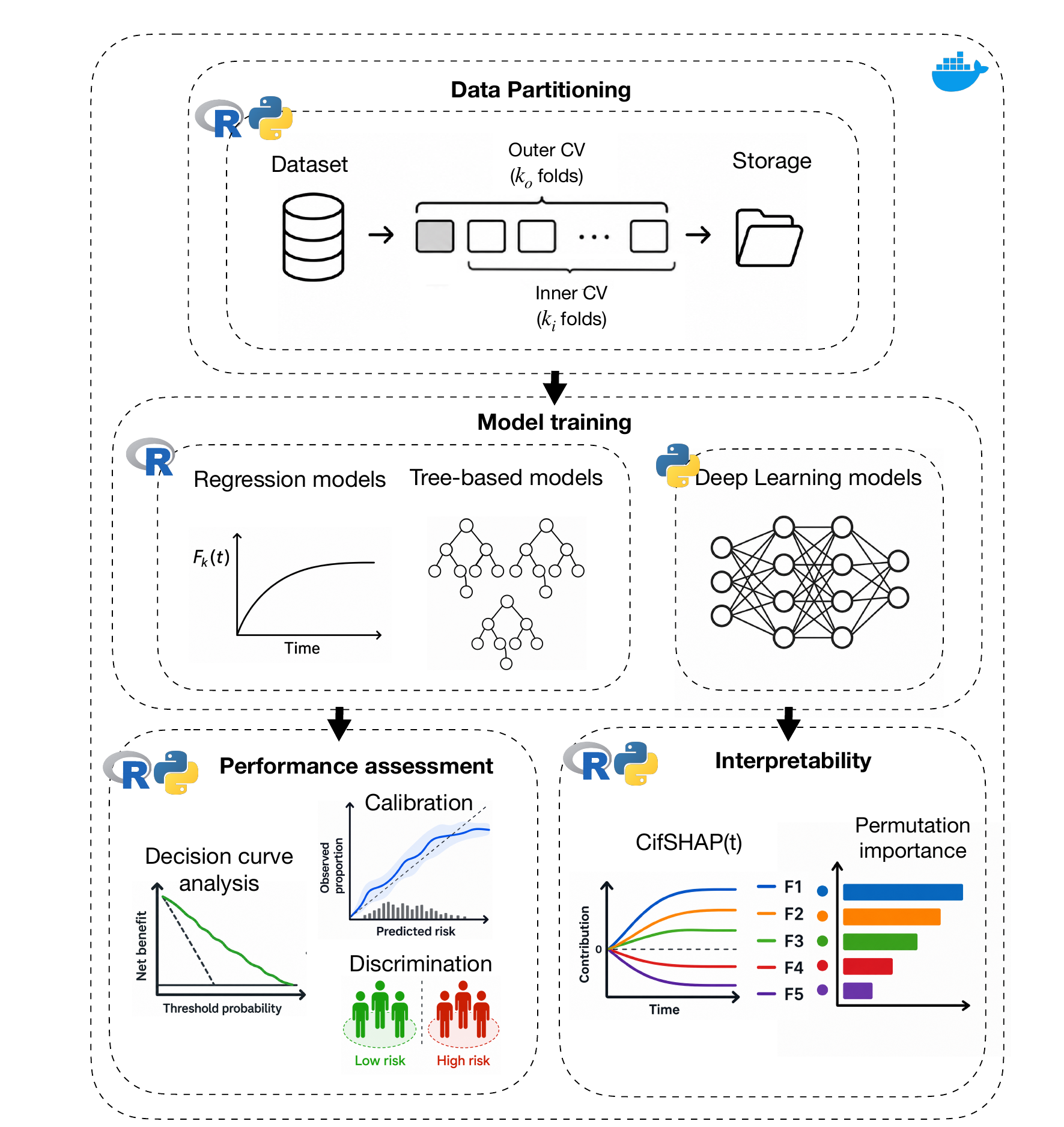}
    \caption{Overview of our benchmark pipeline for assessing risk prediction performance of CR methods. The workflow begins with nested cross-validation data partitioning in R, where datasets are split into outer and inner folds and stored into a parquet format that enables interoperability between R and python. Model training includes regression-based approaches, tree-based methods and deep learning models implemented in both programming languages. Predicted CIFs from the outer test folds are subsequently used for downstream analyses, including predictive performance assessment and interpretability. Performance evaluation comprises calibration, discrimination and clinical utility metrics. Interpretability analyses include model-dependent regression coefficients, model-agnostic time-dependent CifSHAP$(t)$ feature explanations and permutation-based variable importance.}
    \label{fig:diagCs}
\end{figure}

The aim of this study is to design a reproducible, reusable and extendable benchmarking pipeline to evaluate the performance of CR methods, focusing on risk prediction. This enables a comprehensive assessment of existing methods, whilst also facilitating further method development. Figure \ref{fig:diagCs} provides an overview for our pipeline, which we applied to datasets with varying sample sizes and event rates. We selected CR methods covering the statistical and machine learning literature, focusing on those for which software is available in either R \citep{Rsoftware} or python \citep{pythonsoftware}. By considering a common nested cross-validation procedure with a consistent approach for hyper-parameter tuning, predictive performance evaluation and interpretability, our design ensures fair comparisons across models. Throughout, we emphasise interoperability between R and python. All our code is publicly available via GitHub: \url{https://github.com/BBolosSierra/CompRisksBenchmark}.

\subsection{Datasets}

We considered datasets that are commonly used 
to benchmark new CR methods \citep[e.g.~in][]{DeepHit2018AAAI, danks2022PMLR}. 
We prioritised publicly available datasets (e.g.~contained within an R library), and those that can be easily accessed after appropriate approvals \citep[e.g.~Surveillance, Epidemiology, and End Results (SEER)][]{SEER2025}.
The datasets cover different 
sample sizes and event rates 
(Table \ref{tab:datasets}). As the impact of treating CR events as censored depends on the observed percentage of competing events, we only considered those for which the percentage of individuals with a CR event is $\approx 10\%$ or above 
\citep[the minimum threshold suggested by][]{Austin2016Circulation}. 
To reduce computational costs, we randomly selected a subset of 50,000 breast cancer patients within SEER. More information about each dataset, including preprocessing steps (e.g.~inclusion/exclusion criteria) and descriptive statistics, can be found in Supplementary Section \ref{supp:datasets}. 

All datasets represent low-dimensional settings where the number of individuals largely exceeds the number of covariates. For Molecular Taxonomy of Breast Cancer International Consortium (METABRIC), we selected the same features as in \cite{DeepHit2018AAAI} and \cite{danks2022PMLR}. For SEER, breast cancer-related and demographic features were included.
Although the selection of these variables is sufficient for benchmarking purposes, clinical domain knowledge is required for a better informed selection 
if the intention is to be used for clinical decision making.
Mayo Clinic Primary Biliary Cholangitis data (PBC) was kept with 33\% of patients with missing values.

\begin{table}[H]
\caption{Overview of datasets. $N$ denotes the number of individuals, and $p$ the number of available predictors (
excluding those dropped during data pre-processing). Cause 1 refers to primary event of interest, cause 2 refers to the competing risk, and Censor are individuals that did not experience any event during the follow-up period (right censored). The definition of event types is described in Supplementary Section \ref{supp:datasets}. Percentages are computed with respect to all subjects in the dataset and rounded to two decimals. ($^{*}$) The dataset contains missing values in the covariates, but not the outcome. ($^{**}$) The data is not open source but it can be accessed after following SEER data access request steps, specified in Supplementary Section \ref{supp:seer}.}
\label{tab:datasets}
\small
\centering
\begin{minipage}{\textwidth}
\centering
\begin{threeparttable}
\setlength{\tabcolsep}{6pt}
\renewcommand{\arraystretch}{1.15}
\begin{tabular}{lrrrrrlcc}
\toprule
Dataset &
$N$ &
$p$ &
Cause 1 (\%) &
Cause 2 (\%) &
Censor (\%) &
Source \\
\midrule
PBC ($^{*}$)      &   418 & 13 &  5.98 & 38.52 & 55.50 
& survival \tnote{a}\\
HD               &   865 &   6 & 33.64 & 15.61 & 50.75 
& CompRisksVignettes \tnote{b}\\
METABRIC         & 1,936 &   9 & 33.06 & 25.00 & 41.94 
& cBioPortalData \tnote{c} \\
Framingham       & 2,236 &  17 & 23.39 &  9.53 & 67.08 
& riskCommunicator \tnote{d} \\
SEER Breast $(^{**})$      & 50,000 & 13 & 11.63 & 21.15 & 67.22 
& SEER software \tnote{e}\\
\bottomrule
\end{tabular}

\begin{tablenotes}\footnotesize
\item[] \textsuperscript{a} \cite{survival-package}, Version 3.8.3,
\textsuperscript{b} \cite{Monterrubio2024BiometricalJournal}, \cite{monterrubio_comp_risks_vignettes},
\textsuperscript{c} \cite{Ramos2020JCO}, Version 2.23.2,
\textsuperscript{d} \cite{riskcommunicator}, Version 1.0.1, 
\textsuperscript{e} 
\cite{seerstat} SEER$^*$Stat Software Version 9.0.42.2
\end{tablenotes}
\end{threeparttable}
\end{minipage}
\end{table}

\subsection{Models}\label{models}

We 
selected a diverse 
set of CR methods that i) represent distinct formulations (e.g., covariate effects introduced via the cause-specific hazard $h_k(t)$ or directly via the CIF $F_k(t)$), 
ii) include both classical statistical models and modern machine learning approaches, and iii) for which open-source software is available in python or R. The selected methods are cause-specific Cox PH \citep[csCPH;][]{cox1972RSS}, Fine-Gray regression \citep[FGR;][]{FineGray1999JASA}, Penalised Proportional Subdistribution Hazard \citep[FGRP;][]{kawaguchi2021RJournal}, Random Survival Forests for CR \citep[RSF;][]{ishwaran2014Biostats}, DeSurv \citep{danks2022PMLR} and DeepHit \citep{DeepHit2018AAAI}. A summary of key properties for these methods is shown in Table \ref{tab:cr_methods_table1}, see also \cite{Monterrubio2024BiometricalJournal}. 
Beyond the methods selected here, we implemented the benchmark within a reproducible framework (Section \ref{sec::software}) in which new methods and datasets can be easily integrated, facilitating practical applications and method development.

csCPH, FGR and FGRP are regression based, while RSF, DeSurv and DeepHit represent more flexible machine learning methods. csCPH computes regression coefficients 
separately for each cause via $h_k(t)$, 
while $F_k(t)$ is estimated post-hoc 
using the product limit rule to prevent the sum of the cause-specific CIFs (across causes) exceeding 1 as $t \rightarrow \infty$. In contrast, FGR directly 
links covariates to the CIF. 
FGRP extends this, introducing regularisation into the estimation of the 
regression coefficients through a penalty term. 
RSF estimates $F_k(t)$ non-parametrically using an ensemble of survival trees grown using bootstrap samples of the data, with a random subset of covariates used at each node split. The key extension to the CR setting lies in the splitting rules: 
a generalised log-rank rule, which targets differences in cause-specific hazards, or a Gray's test weighted log-rank rule that targets differences in CIF. Missing covariate values can be taken into account when splitting the nodes, without the need for data imputation. 
DeepHit and DeSurv are based on neural networks. 
DeepHit formulates the problem in discrete time, and learns a joint distribution over event times and causes using a multi-task architecture, with a shared representation (across all causes) and cause-specific subnetworks. The loss function has two terms; a likelihood component, 
and a ranking loss to encourage concordance for each event type (defined similarly as in the C-index estimator by \cite{antolini2005}). 
In contrast, DeSurv operates in continuous time. 
Adopting a mixture modelling decomposition \citep{larson1985RSS}, DeSurv combines two neural networks 
to estimate the CIF: one network that maps covariates to cause-specific probabilities 
and another that models the conditional event distribution 
for each cause. The latter uses a neural network to learn how event risk changes over time. By integrating these changes through an ordinary differential equation, it estimates a smooth CIF for each individual. 

Except for csCPH and FGR, each method above requires hyper-parameter tuning. For FGRP, these control the type of penalty (e.g.~lasso, ridge) and the strength of the regularisation. We adopt a ridge penalty, controlled by a single hyper-parameter, performing regularisation without introducing sparsity. For RSF, key hyper-parameters include the number of trees and minimum node size which are used to control model complexity. For DeepHit and DeSurv, hyper-parameters include those controlling neural network architecture and the optimisation process. For DeepHit, there are additional hyper-parameters related to the loss function. 
For DeSurv, we use an extended implementation to include dropout and weight decay as additional hyper-parameters to improve regularisation and reduce overfitting, particularly for smaller datasets. A more detailed description for all model-specific hyper-parameters is provided in Supplementary Section \ref{supp:hyperp} and specific hyper-parameter values are shown in Supplementary Table \ref{tab:hp_grids}.

\begin{table}[H]
\small
\caption{Summary of the selected CR methods. “PH” denotes whether the method assumes covariate effects act multiplicatively on a cause-specific or subdistribution hazard. "Non-linear effects" refers to the models' ability to learn non-linear and interaction effects between covariates. “Hyper. tuning” refers to the presence of tuning hyper-parameters that are model dependent. “Missing values” indicates whether the implementation can natively handle missing covariate values without prior imputation. }
\label{tab:cr_methods_table1}
\centering
\begin{minipage}{\textwidth}
\centering
\begin{threeparttable}
\begin{tabular}{lccccl}
\toprule
Model & Type &
\shortstack{Non-PH or \\ Non-linear effects} &
\shortstack{Hyper.\\tuning} &
\shortstack{Missing\\values} &
\shortstack{Software} \\
\midrule
\multicolumn{6}{l}{\bf Cause-specific hazard} \\
\midrule
Cause-specific Cox PH (csCPH)
& Semiparam.
& $\times$
& $\times$
& $\times$
& riskRegression (R) \tnote{a}\\
\midrule
\multicolumn{6}{l}{\bf Subdistribution / CIF-based } \\
\midrule
Fine \& Gray regression (FGR)
& Semiparam.
& $\times$
& $\times$
& $\times$
& riskRegression (R)  \tnote{a} \\
\shortstack[l]{Penalized Proportional \\subdistribution hazard (FGRP)}
& Semiparam.
& $\times$
& $\checkmark$
& $\times$
& fastcmprsk (R)  \tnote{b}\\
\midrule
\multicolumn{6}{l}{\bf Tree-based} \\
\midrule
Random survival forests (RSF)
& Nonparam.
& $\checkmark$
& $\checkmark$
& $\checkmark$
&  randomForestSRC (R) \tnote{c} \\
\midrule
\multicolumn{6}{l}{\bf Deep learning models} \\
\midrule
DeSurv
& Nonparam.
& $\checkmark$
& $\checkmark$
& $\times$
& DeSurv (python)  \tnote{d}  \\
DeepHit
& \shortstack{Nonparam.\\(discrete-time)}
& $\checkmark$
& $\checkmark$
& $\times$
& pycox (python)  \tnote{e} \\
\bottomrule

\end{tabular}

\begin{tablenotes}
\footnotesize
\item[] \textsuperscript{a} \cite{ozenne2017}, Version 2025.09.17,
\textsuperscript{b} \cite{kawaguchi2021RJournal}, Version 1.24.10 Removed from CRAN on April 2026 available in Docker (see Section \ref{sec::software}),
\textsuperscript{c} \cite{ishwaran2014Biostats}, Version 3.4.4,
\textsuperscript{d} \cite{danks2022PMLR}, DeSurv modified that allows the user to include weight decay and dropout,
\textsuperscript{e} Version 0.2.3,  
\end{tablenotes}
\end{threeparttable}
\end{minipage}
\end{table}

\subsection{Consistency of individual-level predictions}

Beyond overall measures of predictive performance, we assessed the agreement between 
the individual-level predictions generated by each model. Despite having similar performance metrics, models might assign different risks to the same individual, which has implications at the patient-level decision-making. We therefore evaluated agreement both at a clinically relevant prediction horizon using the individual-level CIF (i.e.~$t$-year predictions), and over the entire follow-up period using the Restricted Mean Time Lost (RMTL; Equation \eqref{eq:RMTL_comp_main}), as an individual-level summary of the CIF that is not time-dependent. 
Predictions for cause of interest (cause 1 in Table \ref{tab:datasets}) were obtained from each outer test fold of the nested cross-validation procedure (Section \ref{sec:nestedcv}) and concatenated across folds.
Pairwise agreement was quantified using Pearson correlation coefficients, and the marginal distributions of the predictions were also examined.

\subsection{Predictive performance metrics} \label{section:metrics}



A comprehensive evaluation of predictive performance 
requires the assessment of complementary aspects such as calibration, discrimination, overall prediction error and clinical utility; see \cite{vanGeloven2022BMJ} for an overview. 
Some metrics evaluate performance at a fixed prediction horizon (i.e.,~$t$-year risk), whereas others provide a global assessment over the follow-up period. 
As different metrics quantify distinct properties of a model's behaviour, they might lead to different conclusions in terms of model selection. For instance, a model can have a high discriminative ability (i.e., correctly ranking individuals based on their risk) and be 
miss-calibrated (i.e.,~risk is consistently overestimated or underestimated), or vice versa \citep{Lillelund2025}. 
Here we provide a brief overview for the metrics used in this work which are also summarised in Table \ref{tab:metrics}. Supplementary Section \ref{supp:metrics} provides further details and formal definitions. 


\begin{table}[!htbp]
\small
\caption{Summary of performance metrics included in the benchmark.  Direction indicates whether lower ($\downarrow$) or higher ($\uparrow$) values of the metric represent better performance, or whether estimates should be closer to a reference value or curve (e.g., the diagonal for calibration plots, closer to the value of 1 in O/E ratio, or higher net benefit than treat-all/treat-none strategies in decision curve analysis). Censoring is marked as ($\checkmark$) when there is adjustment for censoring, via pseudo-observations or  inverse probability of censoring weights  inverse probability of censoring weights (IPCW), ($\times$) when there is no adjustment for censoring (via pseudo-observations or IPCW). ($^{*}$) The attainable maximum depends on the outcome prevalence and censoring distribution, and is typically below 1. ($^{**}$) pycox code version is extended to CR with minor modifications, and the method used throughout this benchmark is \textit{adj\_antolini}.}
\label{tab:metrics}
\centering
\centering
\begin{threeparttable}
\begin{tabular}{llllll}
\toprule
Metric & Category & \shortstack{Direction\\(better)} & Range & Censoring & Software\\
\midrule

\multicolumn{6}{l}{\textbf{$ t $-year predicted risk metrics (evaluated at a fixed horizon $t$)}} \\
\midrule

Calibration plots 
& Calibration 
& $\approx \text{diagonal}$ & [0, 1]
&  $\checkmark$  & 
ValidationCompRisks (R) \tnote{a}\\

Integrated Calibration Index (ICI)
& Calibration 
&  $\downarrow$ & $[0,1]$ 
&  $\checkmark$  & 
ValidationCompRisks (R) \tnote{a}\\

Observed to Expected (O/E) ratio
& Calibration 
&  $\rightarrow 1 $ & $[0,\infty)$ 
&  $\checkmark$ &  
ValidationCompRisks (R) \tnote{a}\\

Time-dependent AUC (tdAUC)& Discrimination
& $\uparrow$ & $[0,1]$ 
& $\checkmark$  &  
riskRegression (R) \tnote{b}\\


Weighted Brier Score (BS) & Prediction error
& $\downarrow$ & $[0,1]$ $(^{*})$ 
& $\checkmark$ &  
riskRegression (R) \tnote{b}\\

Net benefit  (decision curve)& Clinical utility
&  $>$ treat-all/none & $(-\infty, 1]$
&  $\times$  
& ValidationCompRisks (R) \tnote{a}\\

\midrule
\multicolumn{6}{l}{\textbf{Global measures}} \\
\midrule

C-index $C^{td}$& Discrimination
& $\uparrow $ & $[0,1]$ 
& $\times$ &  
\pkg{pycox} (python) ($^{**}$) \tnote{c}  \\

C-index $C^{\tau}$ & Discrimination
& $\uparrow $ & $[0,1]$ 
& $\checkmark$ &  
\pkg{pec} (R) \tnote{d}\\

Integrated Brier score (IBS)
& Prediction error 
& $\downarrow $ & $[0,1]$ $(^{*})$
& $\checkmark$ &  
riskRegression(R) \tnote{b}\\

\bottomrule
\end{tabular}
\begin{tablenotes}
\footnotesize
\item[] \textsuperscript{a} \cite{vanGeloven2022BMJ} \cite{van_geloven_validationcomprisks},
\textsuperscript{b} \cite{ozenne2017}, Version 2025.09.17,
\textsuperscript{c} Version 0.2.3,
\textsuperscript{d} \cite{pec}, Version 2025.06.24
\end{tablenotes}
\end{threeparttable}
\end{table}

\subsubsection{Calibration measures: plots and numerical summaries}

Calibration is examined visually for each event type, comparing the predicted risk at a time point $t$ against the observed event proportions. Deviations from the diagonal line indicate miss-calibration. 
Due to censoring and the presence of CR, the observed event proportions in the calibration plot are approximations that can be computed with different methods. 
We use pseudo-observations as a model-agnostic approach, 
using an estimated subject-specific proxy derived from a jackknife estimator 
\citep{Gerds2014SM}. 
We
complement calibration curves with the Integrated Calibration Index (ICI; lower is better) and the observed to expected ratio (O/E ratio; deviations from 1 indicate miss-calibration) as numerical summaries. These are calculated as in \cite{vanGeloven2022BMJ}. 

\subsubsection{Discrimination}
We use the time-dependent Area Under the Curve (tdAUC) for $t$-year predictions \citep{Saha2010Biometrics}, and the C-index as an overall concordance metric \citep[note that the C-index is not a proper metric to assess $t$-year risk;][]{Blanche2019Biostatistics}.  Several C-index definitions exist \citep{sierra2025cindexmultiverse}. Here, we use CR extensions for Antolini's $C_{td}$ \citep{antolini2005} and Uno's $C_{\tau}$ \citep{Uno2011}. Whilst $C_{td}$ is directly calculated from the CIF,  
$C_{\tau}$ requires summarising the CIF into a single measure. 
We propose to use the Restricted Mean Time Lost (RMTL): 

\begin{equation}
\label{eq:RMTL_comp_main}
M_k(\mathbf{x}_i) =
\operatorname{RMTL}_k(\mathbf{x}_i;T^*)=\mathbb{E}\!\left[ (T^*-T_i)
\,\mathbf{1}\{T_i\leq T^*,\,Z_i=k\}
\mid \mathbf{x}_i
\right]=
\int_{0}^{T^*}
F_k(t\mid\mathbf{x}_i)\,dt .
\end{equation}

where $F_k(t \mid \mathbf{x}_i)$ is the CIF for cause $k$ for individual $i$, $\mathbf{x}_i$ denotes a vector of observed covariate values, and $T^*$ is a pre-specified truncation time. 
Analogous to the use of Restricted Mean Survival Time (RMST) for a single event type \citep{sierra2025cindexmultiverse}, the RMTL summarizes the CIF into an interpretable quantity in a CR setting. Specifically, 
it quantifies the expected time lost due to cause $k$, relative to an event-free (immortal) cohort, up to $T^*$ \citep{Lyu2020BMCMRM, Wu2022AJE}. Larger RMTL values indicate a greater cumulative burden of cause-specific risk over $[0, T^*]$.
The RMTL has been used to quantify treatment effects in a CR setting, where hazard-based summaries can be difficult to interpret clinically \citep{Zhao2018JAMA}.

Among these discrimination metrics, only $C_{\tau}$ accounts for the censoring distribution, 
up-weighting those individuals that have had an event despite a high probability of being censored. This is done using Inverse Probability of Censoring Weights (IPCW) calculated using a Kaplan-Meier estimator for the censoring distribution \citep{Kaplan1958JASA}. 


\subsubsection{Overall prediction error} \label{subsec:IBS}
Combining both calibration and discrimination, overall prediction error quantifies how close $t$-year predictions and observed outcomes are for each event type. 
The weighted Brier Score (which, for simplicity, we refer to as BS) is a common choice, based on the squared difference between the predicted CIF and the observed event status by a time point $t$, adjusted by censoring via IPCW.  Integration of BS over a time grid results in the Integrated Brier Score (IBS), a global measure of overall prediction error across time. Lower values indicate better performance: varying from 0 (perfect prediction), to a data-specific maximum that depends on the event rate 
(i.e., when event occurrence proportion is lower, the highest achieved score also decreases) \citep{vanGeloven2022BMJ}. 


\subsubsection{Clinical utility via decision curve analysis}

Beyond assessing discrimination and calibration, it is important to assess whether a model is clinically useful. 
Decision curve analysis addresses this by comparing the model net benefit with two default strategies: \textit{treating all} or \textit{treating none}. 
Net benefit is calculated across a range of risk thresholds, where each of them represents the predicted risk above which an intervention would be recommended. 
Lower thresholds correspond to more permissive intervention strategies where overtreatment is less concerning, whereas higher thresholds reflect more conservative decision-making where unnecessary treatment is considered harmful. A model is considered clinically useful when its net benefit exceeds both default strategies over a range of clinically relevant thresholds. Note, however, that this only provides retrospective \emph{in silico} evidence of clinical utility and does not guarantee gains if the model were to be deployed in real-world settings. A prospective evaluation (e.g.~randomised trial) would be required for that purpose.





\subsection{Interpretability} \label{subsec:interpretability}


For statistical approaches (csCPH, FGR and FGRP),  regression coefficients provide direct insights into the associations between covariates and the cause-specific hazard or CIF. 
However, this is generally not available for more complex machine learning methods (RSF, DeSurv and DeepHit), where interpretability typically relies on post-hoc approaches.

First, we consider 
permutation-based feature importance, quantifying the change in predictive performance (e.g.~IBS) after randomly permuting the values of each covariate individually. 
This is conceptually similar to the feature importance measure implemented in {\tt randomForestSRC} \citep{ishwaran2014Biostats}. By using identical permutations across models, our implementation enables fair comparisons. 
Directly informed by predictive performance, this provides a global measure of variable importance, aggregated across all individuals. By choosing IBS as the target, this also represents an aggregated measure over time. If the goal is $t$-year prediction, alternative target metrics (e.g.~BS or tdAUC) can be chosen. This could be used to monitor how variable importance varies across prediction time frames.   

Beyond aggregated metrics, analysts might also be interested in variable importance at an individual-level; that is, how does each variable contribute to the corresponding predicted risk. 
Recent work by \cite{KRZYZINSKI2023KBS} have expanded SHAP values \citep{lundberg2017shaplibrary} to obtain time-dependent variable explanations for survival models on single event type settings, named SurvSHAP$(t)$. Here, we further extend this concept to CR contexts, introducing CifSHAP$(t)$. In brief, we assume a linear decomposition for the cause-specific CIF at each time point $t$; that is, 
\begin{equation} \label{eq:cifshap}
F_k(t \mid x) = \phi_0^{(k)}(t) + \sum_{j=1}^p \phi_j^{(k)}(t),
\end{equation}
where $\phi_0^{(k)}(t)$ denotes a baseline prediction, which is an average across individuals within a background dataset of size $M$. The latter is usually defined by a random subset of the data for which the explanations are to be calculated (e.g.~individuals in the test set). In Equation \eqref{eq:cifshap}, $\phi_j^{(k)}(t)$ represents the contribution of feature $j$ to the CIF of cause $k$ at time $t$. 
The latter is estimated utilising Kernel SHAP \citep{lundberg2017shaplibrary} in python and its equivalent in R \citep{kernelshapR}. Methodological details on CifSHAP($t$) are provided in Supplementary Section \ref{cifhsap}. 

We use $\phi_j^{(k)}(t)$ to assess cause-specific variable importance over time, which is visualised as an average across individuals. We also consider absolute covariate contributions integrated over time to obtain a global importance measure, akin to permutation-based summaries. This is used to calculate an overall ranking across covariates. 

\subsection{Nested cross-validation (CV)}\label{sec:nestedcv}
\enlargethispage{1.5\baselineskip}



Our benchmarking pipeline was designed to avoid data leakage and ensure consistency between R and python. This was applied separately for each of the datasets in Table \ref{tab:datasets}. 
Outer CV folds are used for model training and out-of-sample evaluation, whilst the inner loop is used for hyper-parameter tuning where required (Supplementary Figures \ref{fig:supp_partitioning}, \ref{fig:supp_model_workflow}). Inner and outer folds are generated using stratified sampling to match the overall event distribution (i.e.~across all event types). As a default, datasets were split into 5 outer and 3 inner folds, but the number of folds was reduced (3 outer; 2 inner) when the proportion of events of a given type is low (PBC and Framingham). 
After variable types (e.g.~continuous versus categorical) are assigned, categorical variables are 
converted into one-hot encoding (i.e.~dummy variables). This avoids potential inconsistencies related to the choice of baseline category. Model training was done in R (csCPH, FGR, FGRP, RSF) or python (DeSurv, DeepHit) depending on software availability. To produce consistent results, the sampling of outer and inner folds was done once (in R), with results stored into parquet and json files to support interoperability \citep{parquetApache}. 

For each test set in the outer CV loop, out-of-sample individual-level CIFs were estimated for a user-defined and dataset-dependent time grid ($\mathcal{T}$) with equally spaced time points (Supplementary Table \ref{datasets_time_grid}). When supported by the model implementation, CIFs were directly predicted at the given time grid. For methods that naively return predictions on a fixed time grid (i.e., RSF), the predicted CIFs were linearly interpolated on the defined time grid consistent across methods. Subsequently, the CIFs were stored as parquet files to enable the computation of predictive performance metrics (Section \ref{section:metrics}), while the corresponding model fits were stored as RDS (R models) and pt (python models) file formats  to compute interpretability measures (Section \ref{subsec:interpretability}). 

The outer test CIF predictions from each CV fold are subsequently loaded in R and python to calculate performance metrics. Computed independently for each outer test fold, metrics were then summarised across folds. Specifically, we report the mean, median and approximate 95\% confidence intervals across the outer test folds. Confidence intervals were estimated using the fold-level standard error and a Student's $t$-distribution with $k_o - 1$ degrees of freedom, where $k_o$ is the number of outer folds. Note that this does not take into account for the overlap in the training datasets across CV folds. Consequently, the reported confidence intervals should be interpreted as a rough proxy for fold-to-fold variability. 
CifSHAP($t$) covariate contributions were computed at the individual-level based on cause-specific CIFs, and subsequently averaged across individuals within each outer fold, with optional stratification by the observed event type (only possible in retrospective evaluation based on labelled datasets with known outcomes). Fold-level summaries were then aggregated across folds to obtain mean trajectories and approximate the 95\% confidence intervals over time (calculated as above).

Hyper-parameter optimization was applied to FGRP, 
RSF, DeSurv and DeepHit. 
To make fair comparisons, all tuned models were optimised using a common criterion. Specifically, we preformed a grid-search, and selected the hyper-parameter configuration that minimised 
the IBS associated to a pre-specified event type (Section \ref{subsec:IBS}).  
By default, 
this is based on the primary event of interest (cause 1 in Table \ref{tab:datasets}). We choose the IBS because it jointly reflects calibration and discrimination. This criterion was applied irrespective of the model-specific defaults, which often differ across model implementations (e.g., deep learning models frequently are tuned to optimise model-specific loss functions). Consequently, differences in predictive performance are attributable to the models themselves rather than to inconsistent hyper-parameter optimisation objectives. Regardless of the software used to fit each model (R and python), IBS was computed using a common implementation (i.e., \pkg{riskRegression}). 
Overall, there is 
little guidance on how to define the hyper-parameter space, particularly for machine learning models that depend on a large number of inputs --- any performance claims are bound to the chosen optimisation grid. 
In this work, we aim to consider a wide range of hyper-parameters values (Supplementary Table \ref{tab:hp_grids}) to ensure that each method can 
achieve competitive performance. For deep learning models, the tuning grids explicitly include regularisation, such as dropout and weight decay to mitigate overfitting in smaller datasets. 

The training of deep learning models (DeepHit, DeSurv) is based on gradient-based optimisation with the 
loss function serving as the optimisation objective. 
Early stopping is typically used to terminate training once the validation loss no longer improves. By default, we do not include early stopping when performing hyper-parameter tuning within the inner CV. 
This is to ensure the validation data is used exclusively for hyper-parameter selection based on the IBS, rather than also influencing the training duration and the fitted model. 
To maintain consistency across methods, we 
choose a fixed optimization budget (i.e., 300 epochs) as in \cite{danks2022PMLR}. We further explore the effect of early stopping during inner validation  as a sensitivity analysis. 
Note that all deep learning models are trained deterministically for reproducibility, and fixed 
seeds are used throughout study.

\subsection{Handling of missing values}
 
Except for RSF, all of the methods considered in this benchmark require missing values to be either removed (e.g.~complete case analysis, or covariate removal) or imputed prior to model fitting. We consider  
univariate imputation and multiple imputation via MICE \citep{mice}. These are implemented under the nested cross-validation setup to avoid data leakage. That is, for each outer CV split, the imputation model is defined using the training set, and subsequently applied to both training and test set partitions (for model fitting and out-of-sample performance evaluation, respectively). The same approach is applied to the inner validation folds. 

For each outer CV iteration, univariate imputation was performed using the median for continuous covariates and the mode for categorical covariates, calculated from the training set. 
For MICE, 
$m$ imputed datasets are generated 
(we used $m=3$). Models are fitted 
separately on each imputed dataset, and predicted CIFs for individuals in the test set are subsequently aggregated by mean pooling across imputations to estimate a single CIF per individual. As the objective is risk prediction rather than inference on model parameters, the focus is on aggregating individual-level predicted CIFs across imputations rather than pooling model parameters.

Performance metrics are estimated directly from the aggregated CIF. For CifSHAP$(t)$, SHAP values are first aggregated across imputations at the individual level; that is, for a given individual, cause $k$, and time point $t$, the corresponding SHAP contributions from the $m$ imputations are averaged. Similarly, permutation-based importance values are first aggregated across imputations prior to fold-level summarisation. Subsequent aggregation across outer CV folds is then performed as described in Section~\ref{sec:nestedcv}. 

Note that, the models used by MICE are dependent on the variable types specified for each covariate. 
These are predictive mean matching 
for continuous variables, logistic regression 
for binary variables, and multinomial logistic regression 
for categorical variables with more than two categories. 
The benchmarking framework further supports the inclusion of auxiliary variables (not used for model fitting) solely for the purpose of imputation. 

\section{Results}
\label{sec::results}


\subsection{Consistency of individual-level predictions}

\begin{figure}[htbp]
    \centering
    \includegraphics[width=0.8\textwidth]{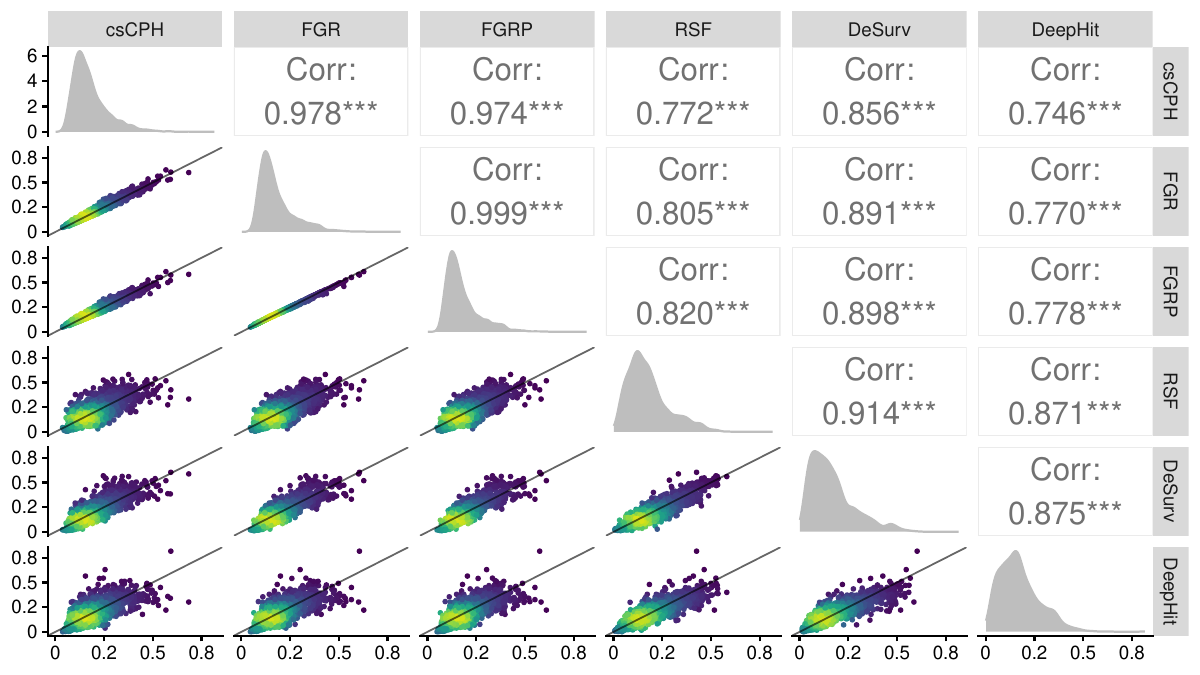}
    \caption{Pairwise comparison of individual-level predicted CIF at a specific time $t =5$ years across benchmarked models for cause 1 on the METABRIC dataset. Lower panels show scatter plots of the predicted risks for individuals from the outer test folds, while diagonal panels display the marginal distributions of predictions for each model. Upper panels report the pairwise Pearson correlation coefficients between model predictions; significance stars indicate the corresponding p-value (i.e., $*** \text{ $p$-value} < 0.001)$}
    \label{fig:cif_correlation}
\end{figure}

Overall, the strength of the correlation in individual-level predictions varied across methods and datasets (Figure \ref{fig:cif_correlation}; Supplementary Figures \ref{fig:sup:consistency_cifs}-\ref{fig:sup:consistency_rmlt}). Figure \ref{fig:cif_correlation} shows representative pairwise comparisons between the individual-level CIF predicted by the different models for METABRIC at $t = 5$ years (a popular prediction time frame in clinical contexts). 
The strongest agreement was observed between classical regression approaches (csCPH, FGR, FGRP), with nearly perfect correlations. While machine learning methods (RSF, DeSurv and DeepHit) produced more distinct predictions, with DeepHit 
exhibiting the lowest correlations, particularly with the regression-based models. This indicated more frequent deviations in individual risk assignment for DeepHit relative to the other approaches. 
Supplementary Figures \ref{fig:sup:consistency_cifs}--\ref{fig:sup:consistency_rmlt} show that these patterns are dataset dependent. For example, in the HD dataset, the largest discrepancies occur among higher-risk individuals, whereas in the PBC dataset predictions were concentrated close to zero for most individuals. For SEER, RSF, DeSurv and DeepHit exhibited a rightward shift relative to those from the classical regression models, indicating systematically higher predicted risks. Similar patterns were observed for the RMTL predictions, suggesting that these differences were not limited to a single prediction horizon but persisted when predictions were summarised over the entire follow-up period.

\subsection{Predictive performance assessment}


\subsubsection{Calibration plots and related metrics}

\begin{figure}[H]
    \centering
    \vspace{-5mm} \includegraphics[width=\textwidth]{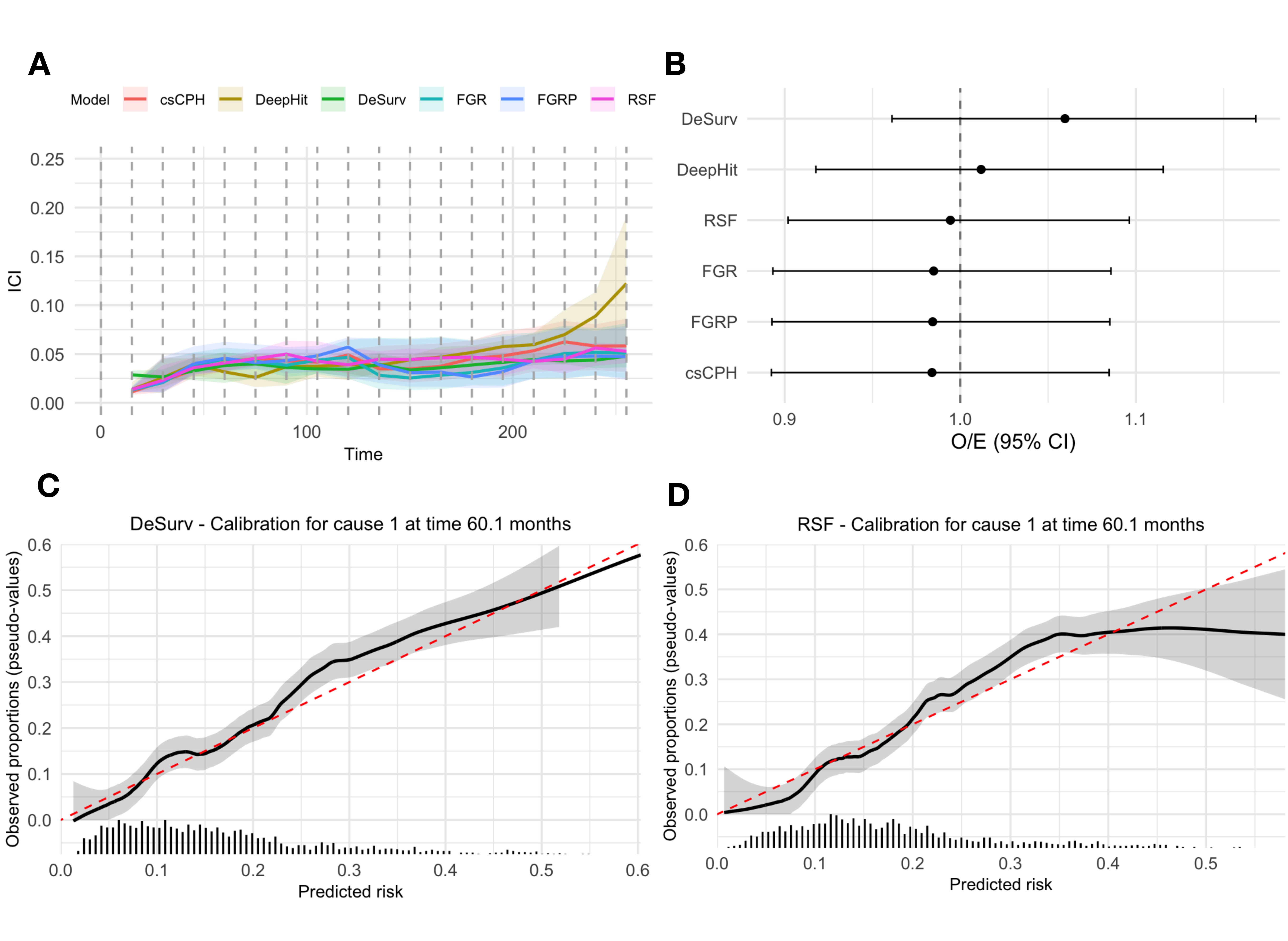}
    \vspace{-8mm}
    \caption{Evaluation with $t$-year metrics of benchmarked models on METABRIC dataset for cause of interest (cause 1). A) Integrated Calibration Index (ICI) over time, where dotted vertical lines indicate the time grid at which the calibration plots were computed, and subsequently ICI derived from. B) O/E ratio across models for cause of interest at $t = 5$ years, where deviations from 1 indicate miscalibration at that specific time point. Calibration plots for at $t = 5$ years, where the red dotted line indicates perfect calibration at the specific time for C) DeSurv and D) RSF.   
    }
    \label{fig:metabric_results}
\end{figure}

For METABRIC, assessment of calibration across multiple time points through the ICI showed broadly similar calibration between methods for small and large values of $t$, deteriorating at later time points particularly for DeepHit (Figure \ref{fig:metabric_results}A). The $O/E$ ratios at $t =5$ years for all methods were close to 1, indicating broadly good calibration (i.e., on average, the predicted risk was close to observed event proportions; Figure \ref{fig:metabric_results}B). 
Calibration curves 
provide a more granular assessment across the whole range of predicted risks. For instance, DeSurv largely followed the diagonal, 
but it underestimated risk around the middle (Figure \ref{fig:metabric_results}C). 
In contrast, RSF tended to overestimate risk at higher predicted probabilities (although with high statistical uncertainty; Figure \ref{fig:metabric_results}D). 
This illustrates how a model can appear well calibrated according to $O/E$ ratio, while exhibiting miscalibration within specific risk ranges, coupled with varying levels of statistical uncertainty. Whilst aggregated metrics derived from the calibration curve (e.g.~ICI, $O/E$ ratio) can be useful when assessing calibration across multiple time points (as shown in Figure \ref{fig:metabric_results}A), solely relying on them could mask more granular patterns. Calibration curves for all datasets are shown in Supplementary Figures \ref{fig:calpbc}-\ref{fig:calseer}.

Across datasets, CR methods exhibited broadly similar calibration. The $O/E$ ratios were generally close to 1, indicating good overall agreement between predicted and observed event probabilities, with higher miscalibration for deep learning models (Supplementary Figure \ref{fig:sup:eos}). Differences between methods were more apparent in the ICI over time, with DeepHit showing greater deterioration in calibration at later time points (Supplementary Figure \ref{fig:sup:icis}).

\subsubsection{Discrimination and overall prediction error}

For METABRIC, when predicting risk at $t = 5$ years, DeSurv outperformed the alternative model approaches based on BS, and RSF reached a comparable performance based on tdAUC (Table \ref{tab:table_metabric}). However, as illustrated in Supplementary Figure \ref{fig:sup:aucs}, the model ranking was not preserved when considering different prediction time frames. This is consistent with the findings by \cite{sonabend2022avoiding} and \cite{sierra2025cindexmultiverse} in the context of the C-index; that is, failure to pre-specify $t$ based on the domain context may open opportunities for metric-\emph{hacking}.

\begin{figure}[H]
    \centering
    \includegraphics[width=1.1\textwidth]{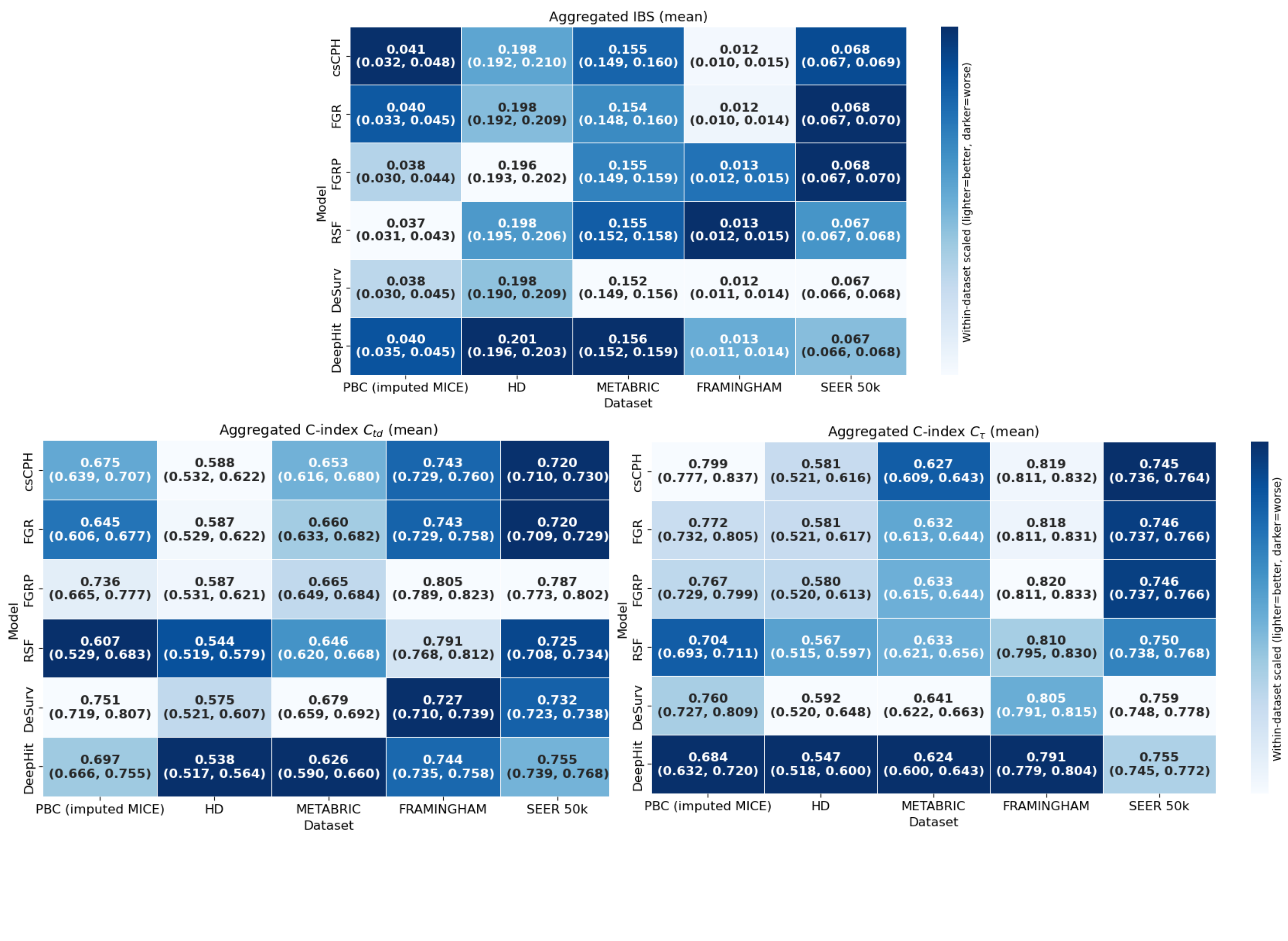}
    \vspace{-18mm}
    \caption{Model benchmark results across datasets for cause 1 for global performance metrics: IBS, $C^{td}$ and $C^{\tau}$. The values shown represent the mean across outer folds and confidence intervals. Values were scaled within each dataset, where lighter colours represent better performance. Note that lower IBS values indicate better overall prediction error, whereas higher values of $C^{td}$ and $C^{\tau}$ indicate better discrimination.}
    \label{fig:heatmaps_global_measures}
\end{figure}

Figure \ref{fig:heatmaps_global_measures} summarises model performance across all datasets under global measures, namely $C_{\tau}$, $C_{td}$ and IBS, showing that model ranking varied depending on the chosen metric --- even when comparing the two C-index definitions. We observed that the spread in IBS values for models was small. 
This might be partly explained by the benchmarking set up, as hyper-parameter optimisation was based on IBS, potentially driving tunable methods towards similar IBS values. Consequently, IBS based comparisons may favour these methods, and highlights the importance of using complementary metrics. Within the same dataset, methods achieving the lowest IBS, did not always achieved the highest C-index ($C_{\tau}$ or $C_{td}$) or vice versa. 
This pattern was most pronounced for PBC, where different global measures favoured different models (RSF under IBS, csCPH under  $C_{\tau}$, and DeSurv under $C_{td}$). Whilst this may be partly explained by the limited sample size ($N = 418$), model ranking differences were also observed for bigger datasets. DeSurv was found to outperform based on IBS and $C_{\tau}$ in three out of the five datasets. Indeed, for METABRIC, DeSurv outperformed other models based on all global measures (Table \ref{tab:table_metabric}). In contrast, DeepHit was often ranked as the worst performing model.
We observed that simpler regression-based approaches remained highly competitive across datasets, particularly FGRP where $C_{td}$ clearly distinguished it as a strong candidate (e.g.~leading on larger datasets such as Framingham and SEER). 




\begin{table}[H]

\caption{\label{tab:table_metabric}Out-of-sample performance (cause = 1) in METABRIC for 5 outer-folds and 3 inner-folds. Global measures are calculated over a time grid ($\mathcal{T}$), where the last time point is $T^*$. IBS, $C^{td}$ and $C^{\tau}$ are calculated up until $T^* = 21$ years, where $\tau = 5$ years.}
\centering
\small

\begin{minipage}{\textwidth}
\centering
\begin{threeparttable}

\makebox[\textwidth][c]{
\begin{tabular}{lcccccc}
\toprule

\multicolumn{7}{c}{\bf $t$-year prediction measures ($t=5$ years)} \\

\midrule

& \multicolumn{3}{c}{BS} 
& \multicolumn{3}{c}{tdAUC} \\

\cmidrule(lr){2-4} \cmidrule(lr){5-7}

Model & Mean & Median & 95\% CI & Mean & Median & 95\% CI \\

\midrule

csCPH  & 0.1334 & 0.1337 & [0.1265, 0.1404]
        & 0.7059 & 0.7050 & [0.6594, 0.7447] \\

FGR    & 0.1329 & 0.1331 & [0.1260, 0.1393]
        & 0.7157 & 0.7098 & [0.6853, 0.7489] \\

FGRP   & 0.1330 & 0.1334 & [0.1260, 0.1393]
        & 0.7188 & 0.7121 & [0.6886, 0.7509] \\

RSF    & 0.1305 & 0.1304 & [0.1233, 0.1379]
        & \cellcolor{rowblue}\bf{0.7378} & \cellcolor{rowblue}0.7473 & \cellcolor{rowblue}[0.7094, 0.7675] \\

\rowcolor{rowblue}
DeSurv & 0.1291 & 0.1272 & [0.1202, 0.1390]
        & 0.7376 & \bf{0.7490} & [0.6987, 0.7759] \\

DeepHit& 0.1311 & 0.1315 & [0.1255, 0.1368]
        & 0.7279 & 0.7319 & [0.6987, 0.7461] \\

\bottomrule
\end{tabular}

\vspace{0.6cm}

}
\begin{tabular}{lccccccccc}
\toprule

\multicolumn{10}{c}{\bf Global measures} \\

\midrule

& \multicolumn{3}{c}{IBS } 
& \multicolumn{3}{c}{$C^{\tau}$}
& \multicolumn{3}{c}{$C^{td}$} \\

\cmidrule(lr){2-4} 
\cmidrule(lr){5-7}
\cmidrule(lr){8-10}

Model 
& Mean & Median & 95\% CI
& Mean & Median & 95\% CI
& Mean & Median & 95\% CI \\

\midrule

csCPH  
& 0.1550 & 0.1564 & [0.1487, 0.1599]
& 0.6269 & 0.6230 & [0.6092, 0.6426]
& 0.6534 & 0.6589 & [0.6157, 0.6804] \\

FGR    
& 0.1544 & 0.1551 & [0.1481, 0.1596]
& 0.6320 & 0.6336 & [0.6128, 0.6443]
& 0.6599 & 0.6625 & [0.6333, 0.6817] \\

FGRP   
& 0.1545 & 0.1558 & [0.1486, 0.1592]
& 0.6326 & 0.6353 & [0.6146, 0.6435]
& 0.6648 & 0.6622 & [0.6491, 0.6836] \\

RSF    
& 0.1551 & 0.1558 & [0.1516, 0.1582]
& 0.6326 & 0.6274 & [0.6207, 0.6558]
& 0.6464 & 0.6552 & [0.6196, 0.6683] \\

\rowcolor{rowblue}
DeSurv 
& 0.1519 & 0.1521 & [0.1486, 0.1561]
& 0.6411 & 0.6408 & [0.6215, 0.6626]
& 0.6790 & 0.6893 & [0.6594, 0.6919] \\

DeepHit
& 0.1557 & 0.1548 & [0.1523, 0.1593]
& 0.6239 & 0.6222 & [0.6001, 0.6426]
& 0.6263 & 0.6346 & [0.5900, 0.6598] \\

\bottomrule
\end{tabular}

\end{threeparttable}
\end{minipage}
\end{table}

\subsubsection{Clinical utility via decision curve analysis}

Overall, differences in clinical utility between methods were modest, with no single model consistently demonstrating superior performance across all datasets.
For SEER, all methods had similar net benefit curves that laying above \textit{treat all} and \textit{treat none} until a threshold probability of $ \approx 0.50$ (Figure \ref{fig:clinical_ut}; similar findings were observed for METABRIC, see Supplementary Figure \ref{fig:clinical_utilities}). 
Therefore, utilising model predictions up to this threshold (i.e., $0.50$) yielded a higher net benefit than indiscriminate intervention or no intervention. In contrast, model-specific net benefit curves differed for FRAMINGHAM dataset, where RSF and DeSurv achieved a maximum net benefit of $\approx 0.020$, meaning that about 20 more patients per 1000 were 
correctly identified for intervention, without increasing the number of unnecessary interventions compared with treating none. Alternative methods showed a lower net benefit at the same probability threshold. 
Clinical utility was of limited use for model-guided clinical decision making when applied to the PBC and HD datasets. Here, greater variability was observed between all methods, along with a narrower threshold range over which the models could outperform the default strategies (Supplementary Figure \ref{fig:clinical_utilities}).

\begin{figure}[H]
    \centering
    \vspace{-5mm}
    \includegraphics[width=\textwidth]{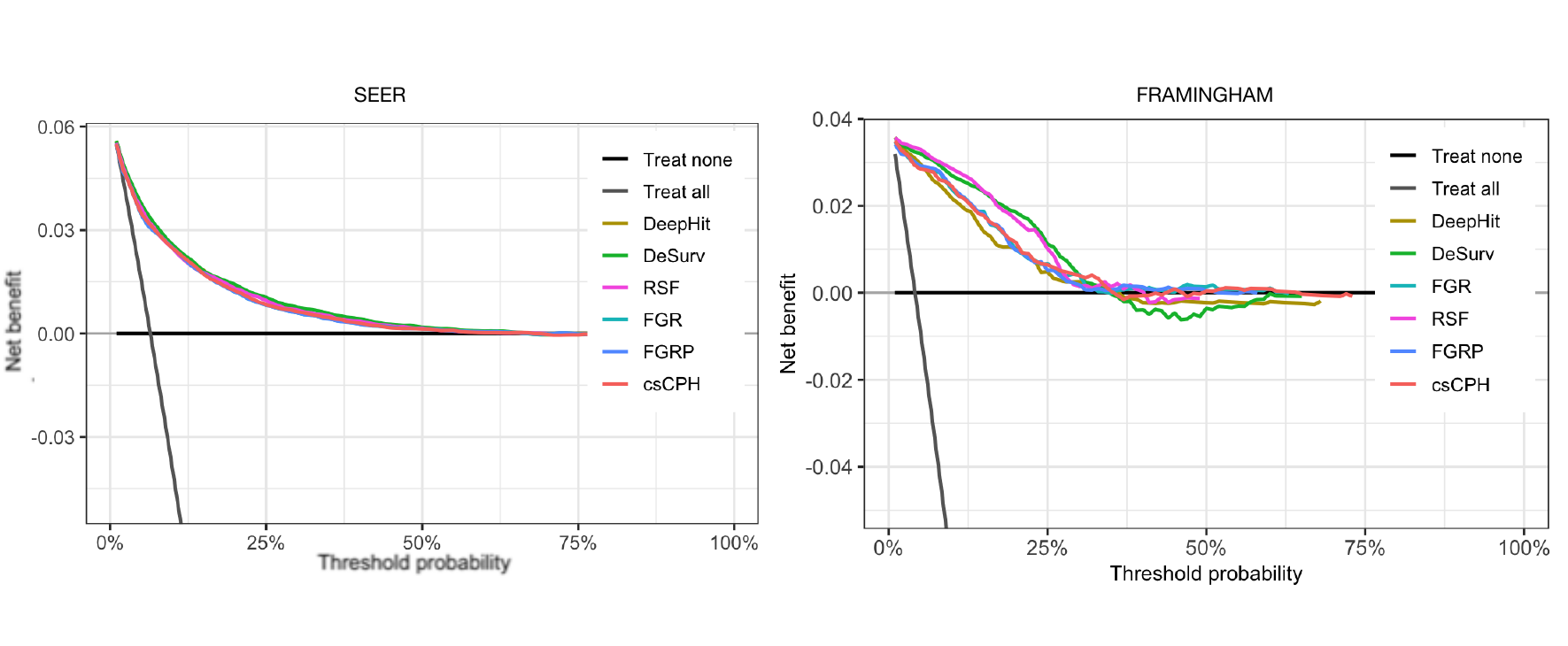}
    \vspace{-10mm}
    \caption{Evaluation of clinical utility at $5$-year for benchmarked models on FRAMINGHAM and SEER datasets for cause of interest (cause 1). Net benefit curves over \textit{treat all} and \textit{treat none} indicate positive net benefit at the given threshold probability. 
    }
    \label{fig:clinical_ut}
\end{figure}

\newpage

\subsection{Interpretability}

\begin{figure}[H]
    \centering
    \includegraphics[width=0.8\textwidth]{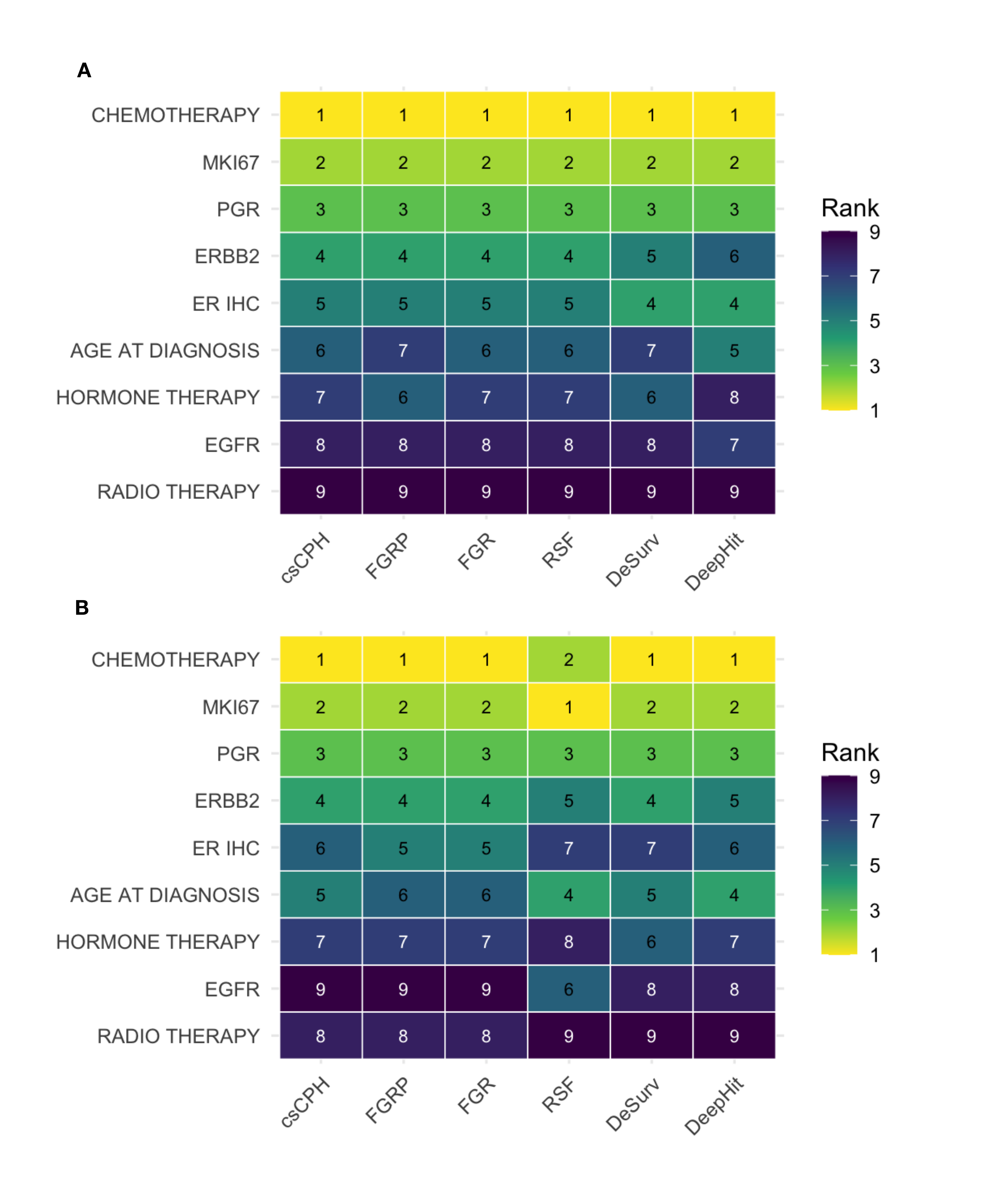}
        \vspace{-10mm}
        \caption{Covariate ranks across models on METABRIC dataset (rank $=1$ indicates the most important variable or higher contribution). A) Average permutation importance across folds based on IBS for cause of interest (cause 1). B) Covariate ranks across models based on the absolute values of CifSHAP($t$) for cause of interest (cause 1) computed as the mean across individuals, integration across time and subsequent mean across folds.}
    \label{fig:ranks}
\end{figure}

Based on permutation importance computed with respect to IBS for primary cause of interest (i.e., breast cancer death) in METABRIC, Figure \ref{fig:ranks}.A shows highly consistent ranking of the top three covariates across models; that is chemotherapy, followed by the gene indicator MKI67, and subsequently, PGR. 
Subsequent covariate ranks were similarly stable, with ERBB2 and ER-IHC consistently appeared in the next tier and only minor reordering for DeSurv and DeepHit. Radiotherapy was consistently less influential across methods.

When CifSHAP($t$) values were integrated over time and aggregated by absolute magnitude (Figure \ref{fig:ranks}.B), the resulting global ranking was comparable to permutation importance. Some ranking reordering of covariates between these two methods occurred since permutation importance reflects how much permuting a covariate degrades IBS, while SHAP reflects how covariates shift the predicted CIF over time. However, the resulting covariate rankings were highly aligned between both approaches for METABRIC. Chemotherapy was the most persistent contribution across models based on SHAP, indicating a higher effect to the final predicted CIF for the primary cause of interest. Similar alignment between both methods was observed on alternative datasets (Supplementary Figure \ref{fig:addinterp}).

\begin{figure}[H]
    \centering
    \includegraphics[width=\textwidth]{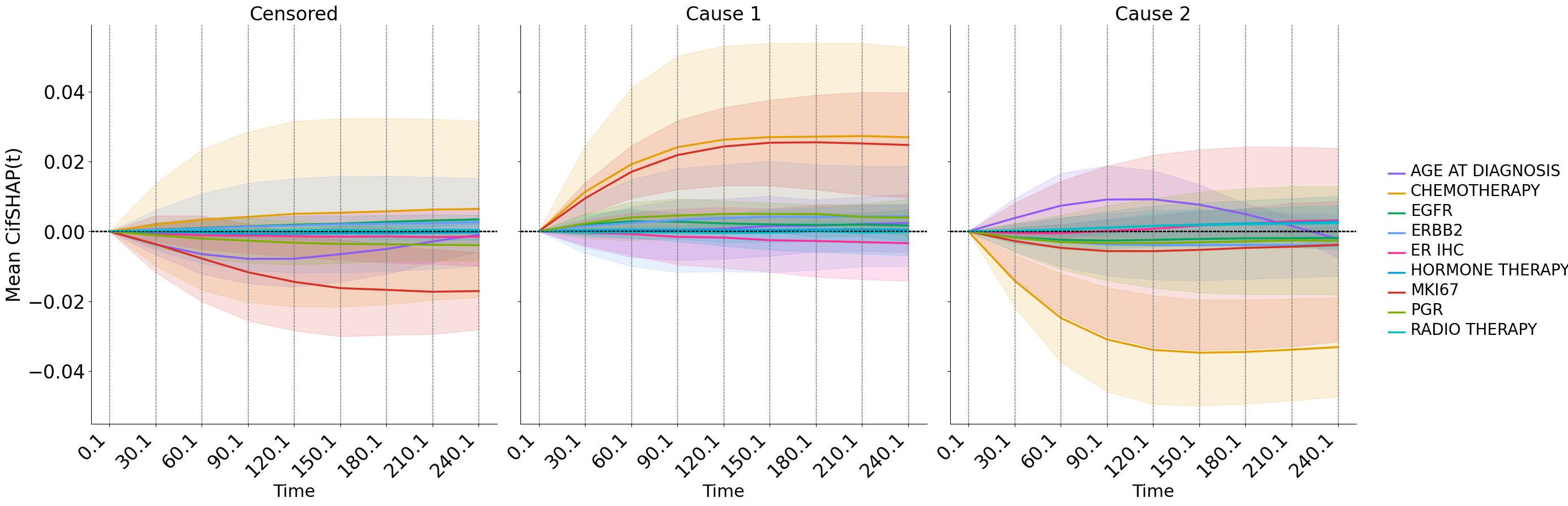}
    \vspace{0.5cm}
    \includegraphics[width=\textwidth]{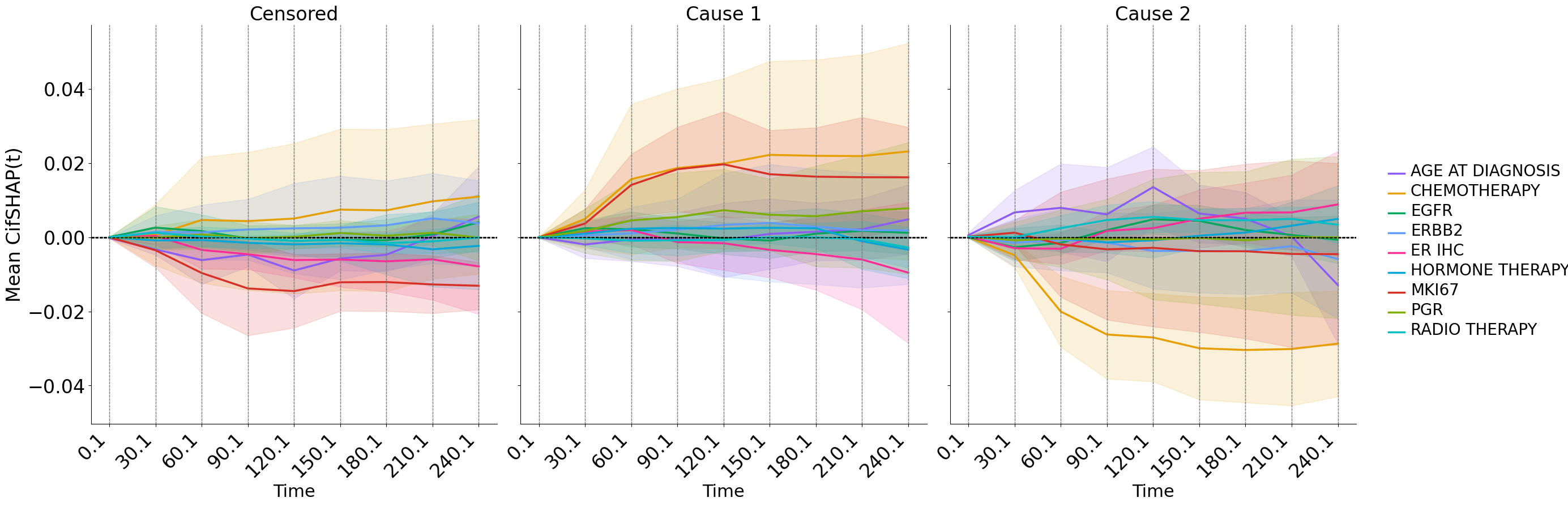}
    \caption{CifSHAP($t$) values for cause 1 stratified by event type on METABRIC dataset for DeSurv (top) and DeepHit (bottom). The SHAP values are calculated per each outer fold, with a background dataset of 400 individuals from the outer training sets, and 50 individuals to be explained derived from the respective outer test. The confidence intervals are calculated across the outer folds. Proportions of events (i.e., censoring, cause 1, and cause 2) from the training set are maintained on the background set, and test set's event proportions are kept on the individuals to be explained. The mean SHAP values are computed at a discretised time grid (in months) represented by each dotted vertical line, and averaged across individuals. The sign of each covariate contribution is interpreted to have an effect on the CIF; a positive contribution indicates a CIF increase, while a negative contribution indicates a CIF decrease.}
    \label{fig:cifshapstratify}
\end{figure}

Based on CifSHAP($t$), covariate contributions presented different time-dependent patterns in METABRIC based on the observed event type as shown in Figure \ref{fig:cifshapstratify} (note that this is only possible in retrospective evaluations based on labelled datasets). For instance, for both  DeSurv and DeepHit, chemotherapy contributed to an increase in CIF for cause 1 (i.e., breast cancer death) for individuals that experienced that event, while the same covariate contributed to a decrease in CIF for cause 1 on individuals that experienced the cause 2 (i.e., death of other causes), and censored individuals had covariate contributions closer to zero. Therefore, CifSHAP($t$) captures not only the temporal evolution of covariate effects, but also how the same covariate can contribute differently across competing events. 
Despite DeSurv and DeepHit representing the strongest and weakest performing methods, respectively, under several metrics outlined in Section \ref{section:metrics}, both models exhibited broadly similar temporal contribution patterns across covariates. This suggested that lower predictive performance does not necessarily imply fundamentally different covariate attribution dynamics.

\subsection{Missingness}

To evaluate the impact of missing data handling strategies, RSF models were trained on the PBC dataset using RSF's internal imputation, univariate imputation and MICE. Overall, the three approaches resulted in high calibration and discrimination, but low on clinical utility (Supplementary Figure \ref{fig:pbc_missigness}). Calibration was comparable with low ICI values that increased over time and overlapping $O/E$ confidence intervals centred around 1. Univariate imputation showed consistently lower calibration performance based on ICI. Likewise, discriminative performance, evaluated with tdAUC, was high for all imputation strategies and decreased over time, and univariate imputation showing the lowest performance. Decision curve analysis at $t=5$ years further demonstrated nearly overlapping net benefit curves, indicating minimal differences in clinical utility between the three imputation strategies. Note, however, that the limited sample size of the PBC dataset led to high statistical uncertainty for all metrics. This may have limited the generalisability of these conclusions. 

\subsection{Optimisation criterion effects}\label{tuning_effects}

Throughout this benchmark, we carried out the hyper-parameter tuning on the inner CV based on the IBS, therefore optimising based on overall prediction error (i.e., calibration and discrimination). Alternatively, the optimisation criterion could be modified to select the hyper-parameters based on other criterion, such as the model's own loss, $C_{td}$ or any other criterion (i.e., another metric). To assess the effect of different criterion on performance metrics, we compared three DeepHit models where hyper-parameter tuning was based on IBS, $C_{td}$ and DeepHit's loss. 

Using METABRIC, DeepHit IBS-based tuned model yielded improved calibration, higher discriminative performance and clinical utility compared to tuning DeepHit based on the its loss or $C_{td}$ (Supplementary Figure \ref{fig:deephit_loss_comparison} and Table \ref{tab:table_deephit_losses}). The last two tuning approaches resulted in highly similar performance estimates across all metrics. This similarity may be explained by 
the composition of DeepHit's loss function, which incorporates a ranking element closely related to the discrimination metric $C^{td}$, effectively resulting in optimisation of hyper-parameters for a similar objective. Consistent with their respective optimisation objectives, the models tuned based on model loss and $C_{td}$ achieved higher performance under $C_{td}$. Instead, the IBS-optimised model outperformed in calibration, discrimination ($C_{\tau}$ and tdAUC) and clinical utility, suggesting that optimisation based on IBS may produce a more balanced model fit. 
These results emphasise the importance of clearly reporting the criteria used to optimise hyper-parameters, as this is critical when interpreting performance metrics and model ranking as this can be strongly influenced by the hyper-parameter optimisation process. 


\subsection{Early stopping effects}

Early stopping is a regularisation method commonly used in deep learning models to avoid overfitting. This method prompts an early termination of the training process when the validation loss (i.e.~the loss function evaluated on a set of data that is not used for model training) is no longer improving. Allowing early stopping during the inner validation, results in a hyper-parameter selection based on the coupled effect of both IBS and validation loss. 

By default, the benchmark pipeline disables early stopping for deep learning models (DeepHit and DeSurv) to ensure a fair comparison across alternative modelling strategies. Nevertheless, the benchmark pipeline allows an option for users to include early stopping during inner CV loop. As a sensitivity analysis, we carried out a comparison between a ``fixed" budget of epochs (i.e., $e = 300$ iterations) against early stopping to assess the impact of both approaches on predictive performance on METABRIC dataset. 

Overall, models trained under the fixed budget achieved  superior calibration, discrimination and clinical utility than their early stopping counterparts (Supplementary Figure \ref{fig:early_stopping} and Table \ref{tab:table_earlystop_fixed}), including lower BS and IBS together with higher tdAUC and $C^{\tau}$. DeepHit with early stopping achieved slightly higher $C^{td}$ than its fixed counterpart. This may be expected due to the composition of its loss function Section \ref{models}. However, enabling early stopping generally reduced predictive performance, particularly for DeSurv under this benchmark settings.

\subsection{Runtime}

Computational runtime was evaluated by fitting the CR models on progressively larger subsets derived from the SEER dataset (i.e., five samples per each size). To ensure that the measured runtime reflected the cost of model fitting, hyper-parameter optimisation was not performed. Instead, models with hyper-parameters (i.e., FGRP, RSF, DeSurv and DeepHit) were evaluated using a fixed pre-specified set. For instance, FGRP was evaluated with $\lambda = 0.5$, while RSF, DeSurv, and DeepHit were evaluated using high-capacity configurations corresponding to the largest tree ensemble configuration or the largest neural network architectures respectively, as indicated in Supplementary Table \ref{tab:hp_grids}. Runtime experiments were conducted on the University of Edinburgh high-performance computing cluster \citep{ECDF2026}, with each model fitted as a separate job requesting a single shared-memory CPU core (64 GB RAM) and runtime was measured as elapsed wall-clock time.

\begin{figure}[H]
    \centering
    \includegraphics[width=0.75\textwidth]{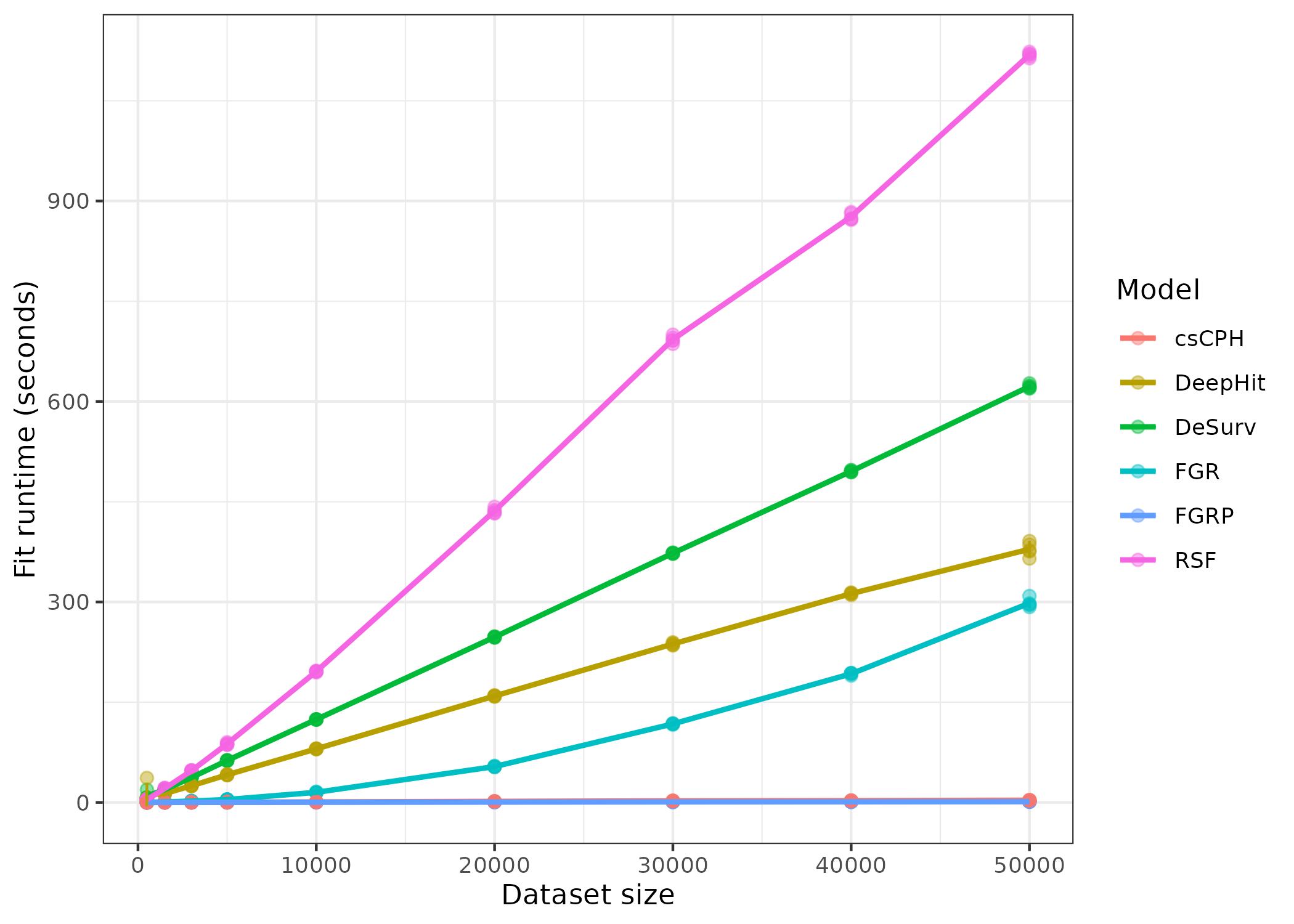}
    \caption{Training runtime (seconds) of each model on progressively larger samples of SEER dataset. Each point corresponds to the observed runtime for fitting a model on a single dataset of the indicated size. Runtime was measured over five independent samples for each dataset size using fixed hyperparameter settings (i.e., without hyperparameter optimisation). }
    \label{fig:runtime}
\end{figure}

As shown in Figure \ref{fig:runtime} computational requirements increased with dataset size for all methods. Machine learning methods, exhibited the longest runtimes, with RSF showing the greatest computational burden. DeSurv also required substantial computation, consistent with the repeated numerical solution of ordinary differential equations during model fitting. 
csCPH and FGRP displayed comparatively short runtimes, with FGR having longer running times. The differences between FGRP and FGR may be attributed to the optimised implementation in \pkg{fastcmprsk}.  These runtimes should be interpreted as illustrative of single model fitting costs under fixed hyper-parameter configurations. In practice, the total computational cost of the benchmarking pipeline would include hyper-parameter optimisation, which may alter  the relative computational burden of these methods. 

\vspace{-5mm}


\section{Discussion}
\label{sec::discussion}

In this work, we established and executed a benchmarking framework for survival models with competing risks (CR), designed to compare heterogeneous modelling approaches under a common experimental setting. We evaluated classical regression models (i.e., csCPH, FGR, FGRP), and more flexible machine learning methods (i.e., RSF, DeSurv, DeepHit) across multiple open source datasets (or easily accessed such as SEER). The benchmark pipeline allows for different strategies to handle missing data and uses nested cross-validation with user-specified number of folds. The hyper-parameter space is pre-specified, but it is modifiable to allow for more extensive tuning exercises. Model performance is assessed capturing complementary criteria covering calibration, discrimination and clinical utility, to account for global measures and $t$-year metrics. Beyond performance, the benchmark pipeline provides 
model-agnostic (permutation importance and CifSHAP$(t)$) and model-specific (regression coefficients) interpretability analysis to assess the association between different covariates and event risk, including permutation importance based on IBS degradation, time-dependent covariate contributions based on the idea of SurvSHAP(t). Key learnings from our experiments are summarised in Box 1, providing practical recommendations to prospective users and developers of CR methods. 

The empirical findings presented here are conditional on the specific set up used in our experiments, including the choice of datasets, event types, random seeds, the resulting nested cross-validation fold assignments, and the hyper-parameter space. Therefore, performance assessment and model rankings based on the collected metrics should be interpreted solely as comparative evidence under this specific benchmark set up and datasets, rather than generalisable statements on which models have superiority. 

Likewise, conclusions regarding missing data handling are based on a single dataset and should be regarded as exploratory rather than general recommendations. Furthermore, these results are not intended to support clinical conclusions, as covariates might not be necessarily relevant beyond its use as practical examples for benchmarking. The benchmarking framework nonetheless remains appropriate for practitioners in the clinical setting, supporting model selection on their own cohorts based on clinically meaningful covariates and endpoints.

Model selection within a dataset depends on the evaluation metrics, as different metrics assess distinct aspects of model performance. 
As a result, a single model rarely achieves the highest performance across all metrics, with METABRIC being a notable exception. Therefore, when using these metrics for model selection, the criterion must be fixed to match the intended deployment objective, as discussed in \cite{Lillelund2025}. For instance, applications that require ranking patients according to their risk may prioritise discrimination, whereas applications that rely on accurate estimates of absolute risk at clinically relevant times are better assessed using calibration or overall prediction error. 
Furthermore, model selection varies depending on the C-index variant (i.e., $C_{\tau}$ or $C_{td}$). These results also reinforce that C-index variants matters for model selection in survival models with CR, in line with the work in \cite{sierra2025cindexmultiverse} on single risk survival models.  



The optimisation criterion used for hyper-parameter tuning is critical, as it determines which aspects of model performance are prioritised during training. Different optimisation objectives (i.e., IBS, $C_{td}$) lead to substantially different model behaviour. For instance, although DeepHit incorporates a discrimination-based component in its loss function, its performance depends strongly on the metric used for hyperparameter tuning. When tuned using IBS, the model achieves a balanced performance across calibration, discrimination and clinical utility. In contrast, tuning based on discrimination (i.e., its loss function or $C_{td}$), unsurprisingly leads to high discrimination but poor calibration. 
This highlights the need to align model deployment objective with both performance evaluation metrics, and the hyperparameter optimisation criterion.
It also emphasises the importance of avoiding circular evaluation frameworks, in which loss function, hyperparameter tuning criterion and final evaluation metric all target the same performance aspect, favouring certain models by construction, as seen in \cite{Lee2026CSAM}. However, this concern is only substantive when conclusions are framed beyond that single metric or objective. If discrimination is the sole pre-specified deployment target, then using a discrimination-aligned loss and tuning and evaluation criterion can be appropriate, provided resulting claims are limited to that aspect of performance (e.g., global discrimination). 



In practise, deep learning models often include early-stopping, which prompts training termination based on improvements in the optimisation objective (i.e., loss function). 
Although early stopping is often used as a regularisation strategy to reduce overfitting, our results showed worse performance when it was enabled under the current set up, suggesting the early stopping might be biasing hyper-parameter tuning towards the loss function (which varies across models) rather than a uniform benchmark selection criteria (i.e., IBS). Early stopping is therefore disabled in our implementation by default to drive hyper-parameter optimisation solely on IBS-based criterion. 

Interpretability results show that models rely on very similar covariate signal for METABRIC dataset, consistent with high agreement in global importance metrics, with strong correlations in individual-level prediction outputs, and overall agreement with integrated CifSHAP($t$) across time rankings. Overall, this suggest that the compared methods rely on very similar covariate signal, and differences in performance are more likely to be due to models' functional form or regularisation, rather than which predictors are used. CifSHAP($t$) results should be interpreted as an association, not as a causal effect. For example treatment variables, such as chemotherapy, 
can act as proxies of disease severity (i.e., patients receiving chemotherapy would have a higher disease severity than those who do not receive it). 

The observed runtime patterns are consistent with the computational bottlenecks of each method. RSF exhibited the highest computational burden, reflecting the cost of fitting large ensembles of trees and evaluating split criteria across splits and event times, which impacts negatively on the sample size scaling. DeepHit was considerably faster than DeSurv in this experiments for the given discretised time grid. However, its discrete-time formulation implies that model complexity is sensitive to the chosen time discretisation \citep{Kvamme2021Lifetime}. In turn, this may affect runtime and memory requirements. In contrast, regression-based approaches show comparatively low computational demands, as model fitting involves estimating a relatively low set of parameters using efficient procedures implemented in highly optimised software (e.g., \pkg{fastcmprsk}).

The current framework has limitations, it targets tabular covariates under right-censoring and does not address time-varying covariates, recurrent events, counterfactual outcomes, interval/left censoring, or high-dimensional settings. The framework is dependent on a nested cross-validation set up, which can reduce the effective training samples size within each outer split, increasing variability in performance estimates, particularly in smaller cohorts or low event proportion datasets. For the deep learning models, this resampling setting further motivates a more controlled analysis of training related effects. In particular, separating the contributions of the architecture, loss function and training set up, to attribute differences in performance to the modelling approach rather than training configuration. Additionally, although deterministic training was adopted to maximise reproducibility, future work should investigate the stability of deep learning methods by quantifying variability across repeated model fits, for example using bootstrap resampling of the training data.  
Further extensions include evaluating missing data handling through controlled missingness experiments on larger datasets, such as SEER, and incorporating recently proposed calibration assessment methods for CR models, as proposed in \cite{alberge2026calibrationsurvivalmodelscompeting}. 
Despite these limitations, the proposed benchmarking framework provides a reproducible and extensible foundation for the systematic evaluation of heterogeneous CR models, which can facilitate the development, validation and adoption of future CR prediction models.

\newpage
\section{Software}
\label{sec::software}
To facilitate reproducibility and adoption, all code required to reproduce the benchmark results is publicly available in GitHub at \href{https://github.com/BBolosSierra/CompRisksBenchmark}{CompRisksBenchmark}. The repository includes the full benchmark pipeline, covering data download, data processing, data partitioning, model training under nested cross-validation, performance assessment and post-hoc interpretability methods. In the particular case of SEER, the steps followed to download the dataset after receiving access to the SEER*Stat Software are provided in Supplementary Section \ref{supp:seer}, and subsequent processing steps covered in GitHub repository. The benchmark provides R markdown workflows to analyse each dataset and visualise results, while Jupyter notebooks are provided to reproduce the python-based analysis.
In addition, Docker images are provided, in R and python to ensure a reproducible computational environment and to minimise issues with dependencies. The benchmarking framework is currently being formalised into an R package.

\vspace{0.4em}
\begin{tcolorbox}[
    title=\textbf{Box 1: Key learnings},
    colback=gray!5,
    colframe=black,
    boxrule=0.6pt,
    arc=2mm,
]
\textbf{Performance evaluation according to the intended deployment objective.}
This benchmark demonstrates that calibration, discrimination, overall prediction error and clinical utility capture distinct aspects of predictive performance, and that metric selection should be driven by the intended clinical use rather than by convention.

\vspace{0.4em}

\textbf{Hyperparameter tuning substantially influences benchmark conclusions.}
The choice of tuning criterion can materially change model rankings. IBS emerged as a robust general-purpose optimisation target, highlighting the importance of transparently reporting tuning strategies alongside predictive performance.

\vspace{0.4em}

\textbf{Model interpretation requires careful clinical context.}
Permutation importance and CifSHAP($t$) identify influential predictors and their time-dependent contributions. However, their interpretation should be guided by clinical knowledge, as they characterise model behaviour rather than causal effects.

\vspace{0.4em}

\textbf{Model complexity should be justified by meaningful performance gains and considering dataset size.}
Across this benchmark, simpler regression-based methods frequently achieved competitive performance while requiring substantially fewer computational resources than machine learning models, making them a strong baseline against which more complex machine learning and deep learning approaches should be evaluated, particularly since the latter tended to show greater potential on larger datasets.

\end{tcolorbox}

\section*{Contributions \& Acknowledgements}
Following the Contribution Role Taxonomy (CRediT) \citep{hosseini2026credit}; the \textit{conceptualization}, \textit{methodology}, \textit{writing - review \& editing} of this work has been carried out by BS, CV, PH, CM and SF. \textit{Data curation}, \textit{formal analysis}, \textit{software} and \textit{writing - original draft} by BS. \textit{Supervision} by CV, PH, CM. \textit{Funding Acquisition} by PH and CM. 
Funding for the project was provided by Cancer Research UK Scotland Centre (CTRQQR\-2021\text{/}100006).
Authors acknowledge early support of Nevena Tsolova and Andreas Bender on initial conceptual ideas and dataset sourcing.\\
{\it Conflict of Interest}: None declared.

\bibliographystyle{biorefs}
\bibliography{refs}

\clearpage
\setcounter{section}{0}
\renewcommand{\thesection}{S\arabic{section}}

\setcounter{figure}{0}
\renewcommand{\thefigure}{S\arabic{figure}}

\setcounter{table}{0}
\renewcommand{\thetable}{S\arabic{table}}

\begingroup
\raggedbottom

\section{Supplementary Material}\label{sec::suplementary}
\subsection{Additional Results}
\subsubsection{Missigness}

\vfill

\begin{figure}[!ht]
    \centering
    \includegraphics[width=\textwidth]{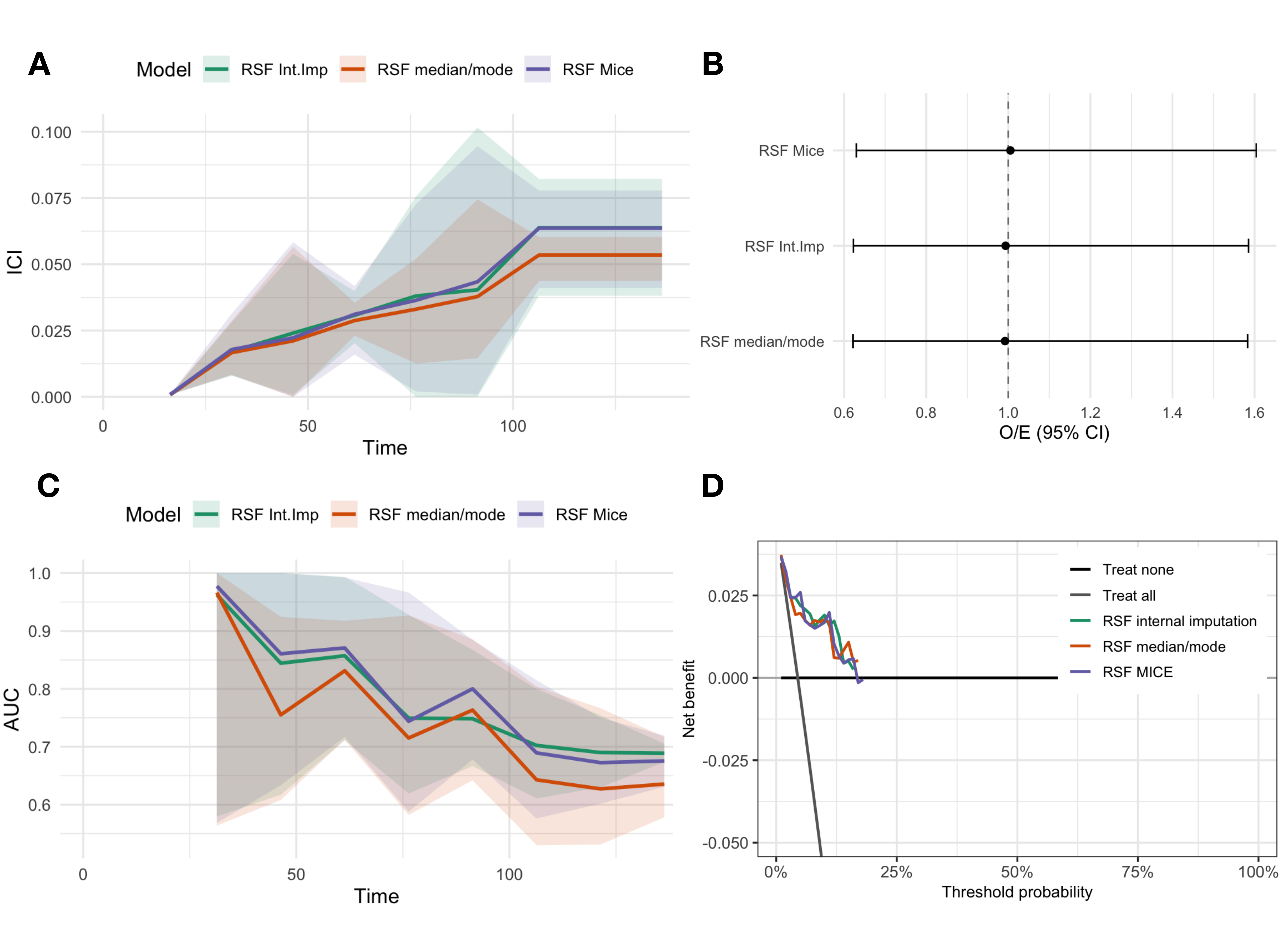}
    \vspace{-7mm}
    \caption{PBC data imputing results. Evaluation of $t$-year metrics on PBC dataset for RSF under different imputation methods; "Int.Imp" refers to the internal RSF handling of missing data, "median/mode" to the univariate imputation based on median and mode, and "Mice" to Multiple Imputation with Chained Equations (MICE). A) Calibration summary: Integrated Calibration Index (ICI) over time,  where dotted vertical lines indicate the time grid at which the calibration plots were computed, and subsequently ICI derived from. B) O/E ratio across models for cause of interest at $t = 5$ years, where deviations from 1 indicate miscalibration at that specific time point. C) Time-dependent Area Under the Survival Curve (AUC) over time D) Clinical utility plot for all the models at $t = 5$ years. Net benefit curves over \textit{treat all} and \textit{treat none} indicate positive net benefit at the given threshold probability. }
    \label{fig:pbc_missigness}
\end{figure}

\vfill
\subsubsection{Optimisation criterion}


\begin{figure}[H]
    \centering
    \includegraphics[width=\textwidth]{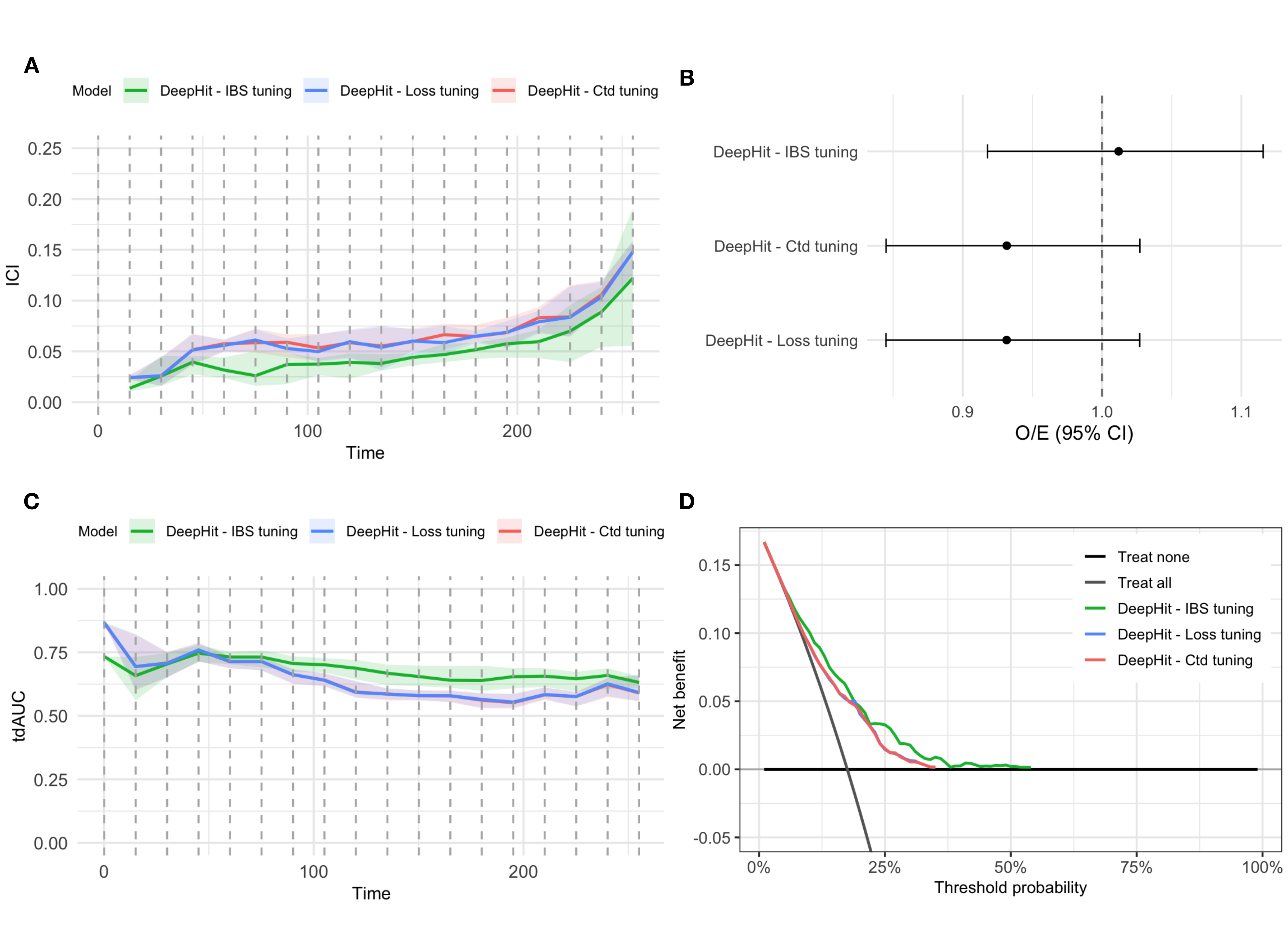}
    \caption{Evaluation of $t$-year metrics on METABRIC dataset (cause 1) of DeepHit models based on different hyper-parameter selection criterion: $C_{td}$, IBS, or the model loss. Where the models have been trained on a "fixed" number of epochs ($e= 300$). A) Calibration summary: Integrated Calibration Index (ICI) over time,  where dotted vertical lines indicate the time grid at which the calibration plots were computed, and subsequently ICI derived from. B) O/E ratio across models for cause of interest at $t = 5$ years, where deviations from 1 indicate miscalibration at that specific time point. C) Time-dependent Area Under the Survival Curve (AUC) over time D) Clinical utility plot for all the models at $t = 5$ years. Net benefit curves over \textit{treat all} and \textit{treat none} indicate positive net benefit at the given threshold probability. }
    \label{fig:deephit_loss_comparison}
\end{figure}

\begin{table}[H]
\vspace{-6mm}
\caption{\label{tab:table_deephit_losses}Out-of-sample performance (cause = 1) in METABRIC for 5 outer-folds and 3 inner-folds. Global measures are calculated over a time grid ($\mathcal{T}$), where the last time point is $T^*$. IBS, $C^{td}$ and $C^{\tau}$ are calculated up until $T^* = 21$ years, where $\tau = 5$ years.}
\centering
\small

\begin{minipage}{\textwidth}
\centering
\begin{threeparttable}

\makebox[\textwidth][c]{
\begin{tabular}{lcccccc}
\toprule

\multicolumn{7}{c}{\bf $t$-year prediction measures ($t=5$ years)} \\

\midrule

& \multicolumn{3}{c}{BS}
& \multicolumn{3}{c}{tdAUC} \\

\cmidrule(lr){2-4} \cmidrule(lr){5-7}

Model & Mean & Median & 95\% CI & Mean & Median & 95\% CI \\

\midrule

\rowcolor{rowblue}
DeepHit (IBS)
& \bf{0.1311} & \bf{0.1315} & \bf{[0.1243, 0.1379]}
& \bf{0.7279} & \bf{0.7319} & \bf{[0.7035, 0.7523]} \\

DeepHit ($C_{td}$)
& 0.1358 & 0.1384 & [0.1282, 0.1433]
& 0.7107 & 0.7133 & [0.6871, 0.7344] \\

DeepHit (Loss)
& 0.1357 & 0.1382 & [0.1282, 0.1433]
& 0.7118 & 0.7143 & [0.6881, 0.7355] \\

\bottomrule
\end{tabular}
}

\vspace{0.1cm}

\makebox[\textwidth][c]{
\begin{tabular}{lccccccccc}
\toprule

\multicolumn{10}{c}{\bf Global measures} \\

\midrule

& \multicolumn{3}{c}{IBS}
& \multicolumn{3}{c}{$C^{\tau}$}
& \multicolumn{3}{c}{$C^{td}$} \\

\cmidrule(lr){2-4}
\cmidrule(lr){5-7}
\cmidrule(lr){8-10}

Model
& Mean & Median & 95\% CI
& Mean & Median & 95\% CI
& Mean & Median & 95\% CI \\

\midrule

\cellcolor{rowblue}DeepHit (IBS)
& \cellcolor{rowblue}\bf{0.1557} & \cellcolor{rowblue}\bf{0.1548} & \cellcolor{rowblue}\bf{[0.1516, 0.1598]}
& \cellcolor{rowblue}\bf{0.6239} & \cellcolor{rowblue}\bf{0.6222} & \cellcolor{rowblue}\bf{[0.6011, 0.6468]}
& 0.6263 & 0.6346 & [0.5900, 0.6598] \\

DeepHit ($C_{td}$)
& 0.1651 & 0.1640 & [0.1627, 0.1675]
& 0.5999 & 0.6065 & [0.5846, 0.6152]
& \cellcolor{rowblue}0.6832 & \cellcolor{rowblue}\bf{0.6914} & \cellcolor{rowblue}[0.6436, 0.7088] \\

DeepHit (Loss)
& 0.1650 & 0.1639 & [0.1626, 0.1675]
& 0.6007 &0.6071 & [0.5859, 0.6156]
& \cellcolor{rowblue}\bf{0.6842} & \cellcolor{rowblue}0.6911 & \cellcolor{rowblue}\bf{[0.6427, 0.7104]} \\

\bottomrule
\end{tabular}
}

\end{threeparttable}
\end{minipage}
\end{table}

\subsubsection{Early stopping in Deep Learning models}\label{supp:earlystopping}



\begin{figure}[H]
    \centering

    \includegraphics[width=\textwidth]{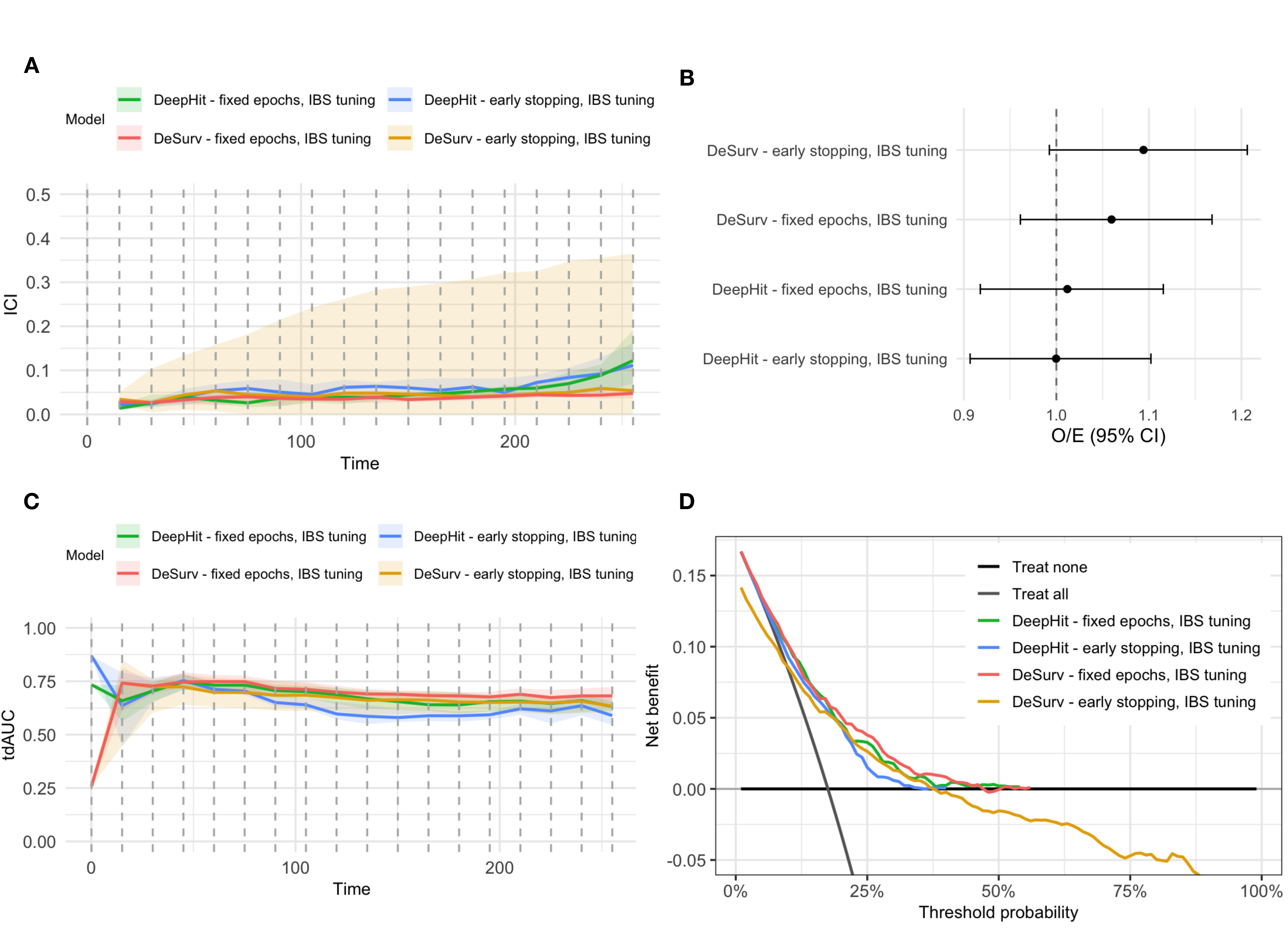}
    \vspace{-5mm}
    \caption{Evaluation of $t$-year metrics on METABRIC dataset (cause 1) of DeepHit and DeSurv models under early stopping or fixed number of epochs (i.e., $e = 300$). The early stopping includes patience as parameter determining the number of consecutive training epochs allowed without improvement in the validation loss before training termination. As default the patience is defined to be 20 as in DeSurv \citep{danks2022PMLR}.  Where the models have been trained based on IBS hyper-parameter tuning criterion. A) Calibration summary: Integrated Calibration Index (ICI) over time,  where dotted vertical lines indicate the time grid at which the calibration plots were computed, and subsequently ICI derived from. B) O/E ratio across models for cause of interest at $t = 5$ years, where deviations from 1 indicate miscalibration at that specific time point. C) Time-dependent Area Under the Survival Curve (AUC) over time D) Clinical utility plot for all the models at $t = 5$ years. Net benefit curves over \textit{treat all} and \textit{treat none} indicate positive net benefit at the given threshold probability. }
    \label{fig:early_stopping}
\end{figure}

\begin{table}[H]
\vspace{-6mm}
\caption{\label{tab:table_earlystop_fixed} Out-of-sample performance (cause = 1) in METABRIC for 5 outer-folds and 3 inner-folds for DeepHit and DeSurv under early stopping ("Early stop") and fixed training ("Fixed" with $e = 300$ epochs). Global measures are calculated over a time grid ($\mathcal{T}$), where the last time point is $T^*$. IBS, $C^{td}$ and $C^{\tau}$ are calculated up until $T^* = 21$ years, where $\tau = 5$ years.}
\centering
\small

\begin{minipage}{\textwidth}
\centering
\begin{threeparttable}

\makebox[\textwidth][c]{
\begin{tabular}{lcccccc}
\toprule

\multicolumn{7}{c}{\bf $t$-year prediction measures ($t=5$ years)} \\

\midrule

& \multicolumn{3}{c}{BS}
& \multicolumn{3}{c}{tdAUC} \\

\cmidrule(lr){2-4} \cmidrule(lr){5-7}

Model & Mean & Median & 95\% CI & Mean & Median & 95\% CI \\

\midrule

\rowcolor{rowblue}DeSurv (Fixed)
& \bf{0.1291} & \bf{0.1272} & \bf{[0.1191, 0.1391]}
& \bf{0.7376} & \bf{0.7490} & \bf{[0.6959, 0.7792]} \\

DeepHit (Fixed)
& 0.1311 & 0.1315 & [0.1243, 0.1379]
& 0.7279 & 0.7319 & [0.7035, 0.7523] \\

DeSurv (Early stop)
& 0.1474 & 0.1395 & [0.1162, 0.1786]
& 0.6985 & 0.6976 & [0.6297, 0.7672] \\

DeepHit (Early stop)
& 0.1352 & 0.1378 & [0.1267, 0.1436]
& 0.7184 & 0.7120 & [0.6977, 0.7391] \\

\bottomrule
\end{tabular}
}

\vspace{0.1cm}

\makebox[\textwidth][c]{
\begin{tabular}{lccccccccc}
\toprule

\multicolumn{10}{c}{\bf Global measures} \\

\midrule

& \multicolumn{3}{c}{IBS}
& \multicolumn{3}{c}{$C^{\tau}$}
& \multicolumn{3}{c}{$C^{td}$} \\

\cmidrule(lr){2-4}
\cmidrule(lr){5-7}
\cmidrule(lr){8-10}

Model
& Mean & Median & 95\% CI
& Mean & Median & 95\% CI
& Mean & Median & 95\% CI \\

\midrule

\rowcolor{rowblue}DeSurv (Fixed)
& \bf{0.1519} & \bf{0.1521} & \bf{[0.1480, 0.1559]}
& \bf{0.6411} & \bf{0.6408} & \bf{[0.6209, 0.6613]}
& \bf{0.6790} & \bf{0.6893} & \bf{[0.6594, 0.6919]} \\

DeepHit (Fixed)
& 0.1557 & 0.1548 & [0.1516, 0.1598]
& 0.6239 & 0.6222 & [0.6011, 0.6468]
& 0.6263 & 0.6346 & [0.5900, 0.6598] \\

DeSurv (Early stop)
& 0.1837 & 0.1564 & [0.1283, 0.2390]
& 0.6088 & 0.6201 & [0.5665, 0.6510]
& 0.6507 & 0.6589 & [0.6026, 0.6849] \\

DeepHit(Early stop)
& 0.1633 & 0.1628 & [0.1592, 0.1673]
& 0.6072 & 0.6071 & [0.5806, 0.6337]
& 0.6692 & 0.6776 & [0.6245, 0.6976] \\

\bottomrule

\end{tabular}
}

\end{threeparttable}
\end{minipage}
\end{table}

\endgroup

\subsection{Supplementary Figures}

\begin{figure}[htbp]
    \centering
    \includegraphics[
        width=\textwidth,
        trim={0 10cm 0 9cm},
        clip
    ]{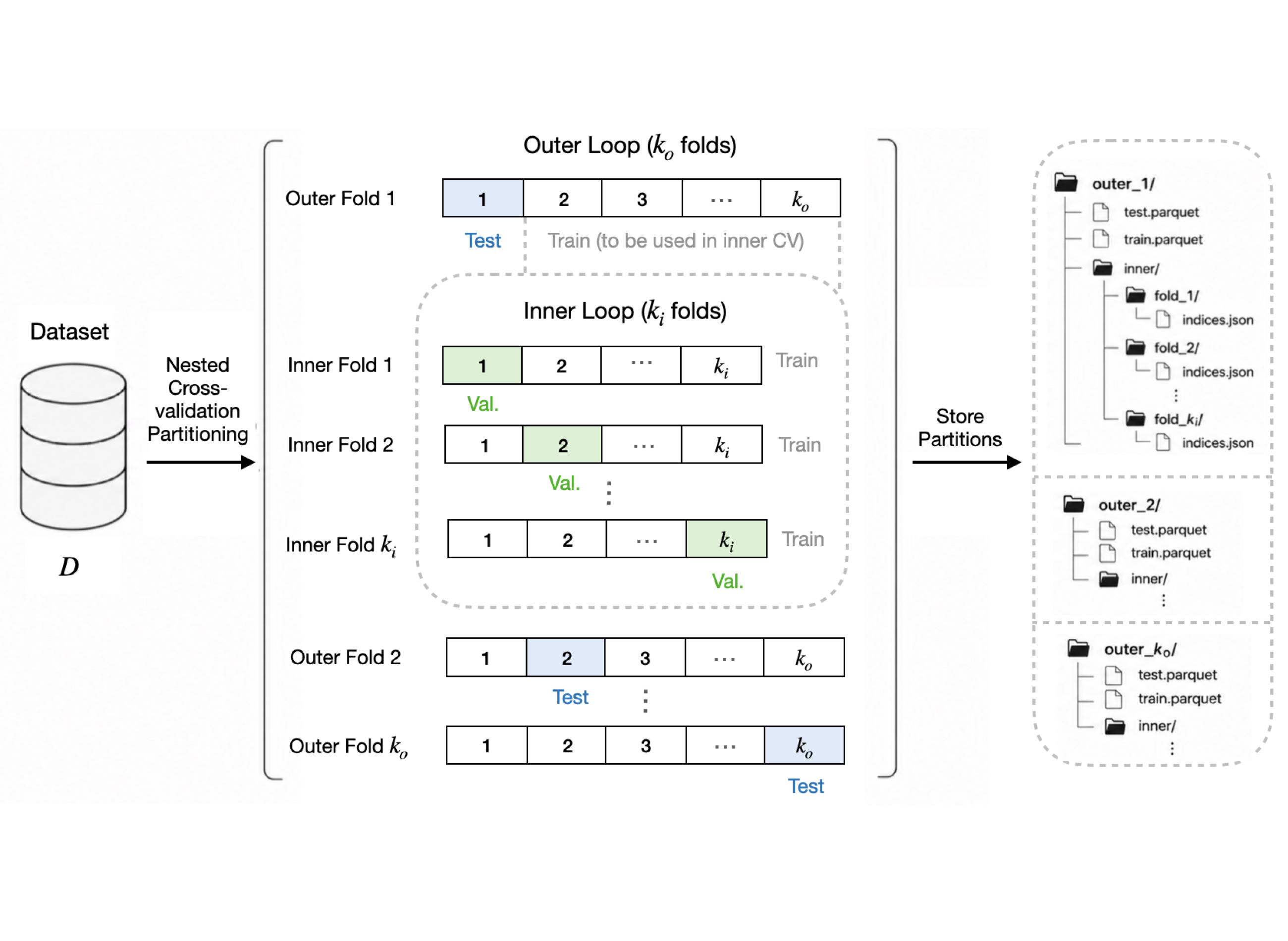}
    \caption{
    Nested cross-validation partitioning and data storage structure used in the benchmarking framework. The dataset is divided into $k_o$ outer folds, with each outer fold used once as the independent test partition. The corresponding outer training partition is further divided into $k_i$ inner folds for hyper-parameter optimisation. All train and test partitions are stored separately for each outer and inner fold to ensure reproducibility and prevent information leakage between model selection and final performance evaluation.
    }
    \label{fig:supp_partitioning}
\end{figure}

\begin{figure}[htbp]
    \centering
    \includegraphics[
        width=\textwidth
    ]{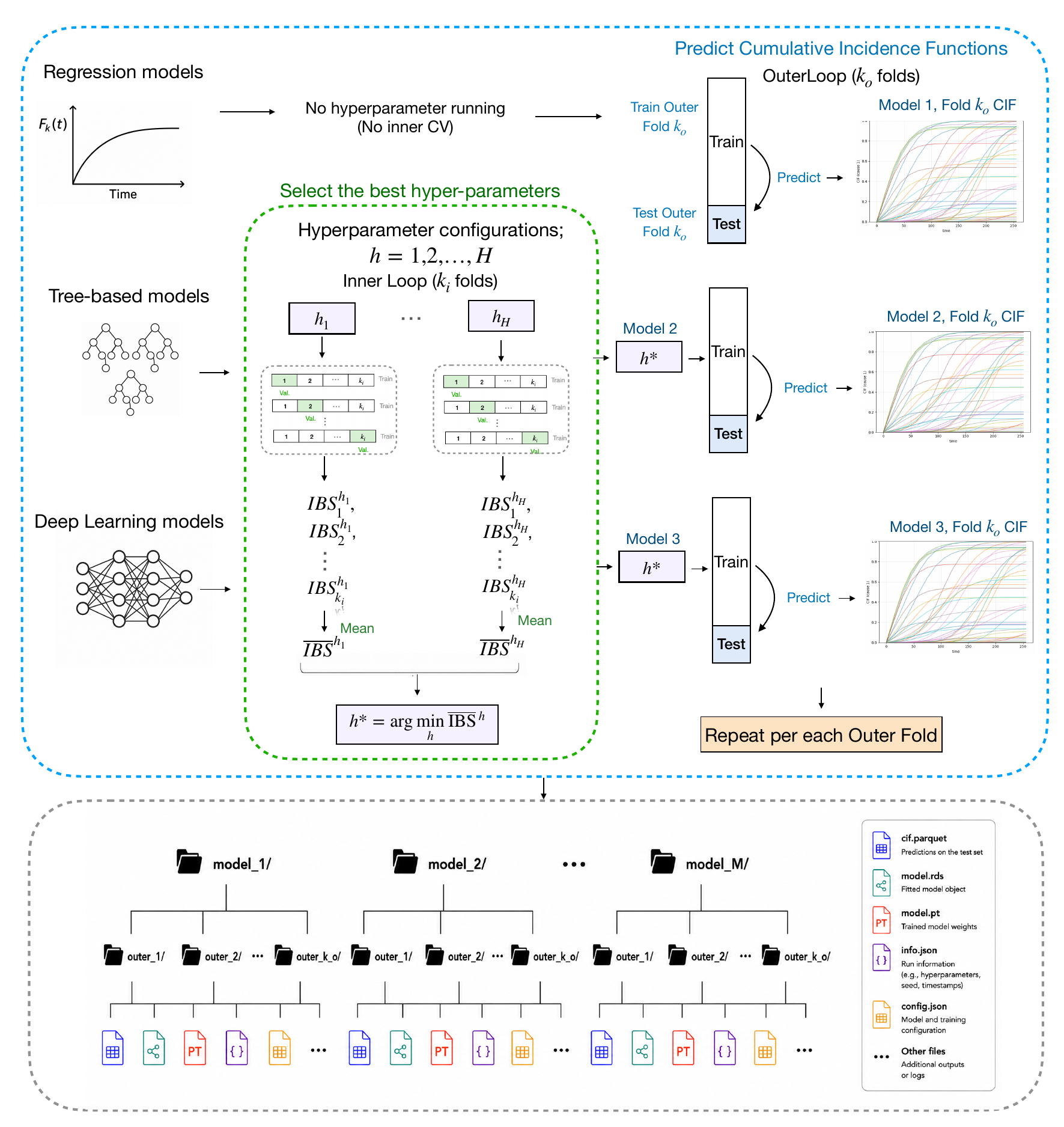}
    \caption{
    Model training, hyper-parameter optimisation and prediction workflow within each outer cross-validation fold. Models without tunable hyper-parameters are fitted directly on the outer training partition. For models requiring optimisation, candidate hyper-parameter configurations are evaluated across the inner folds using the selected optimisation criterion, and the best-performing configuration is then used to refit the model on the complete outer training partition. The fitted model is evaluated on the corresponding outer test partition to generate cause-specific cumulative incidence function predictions. This process is repeated across all outer folds and models, with fitted models, predictions, hyperparameters and performance outputs stored for subsequent evaluation and interpretability analyses.
    }
    \label{fig:supp_model_workflow}
\end{figure}

\begin{figure}[H]

    \centering
    \includegraphics[width=\linewidth]{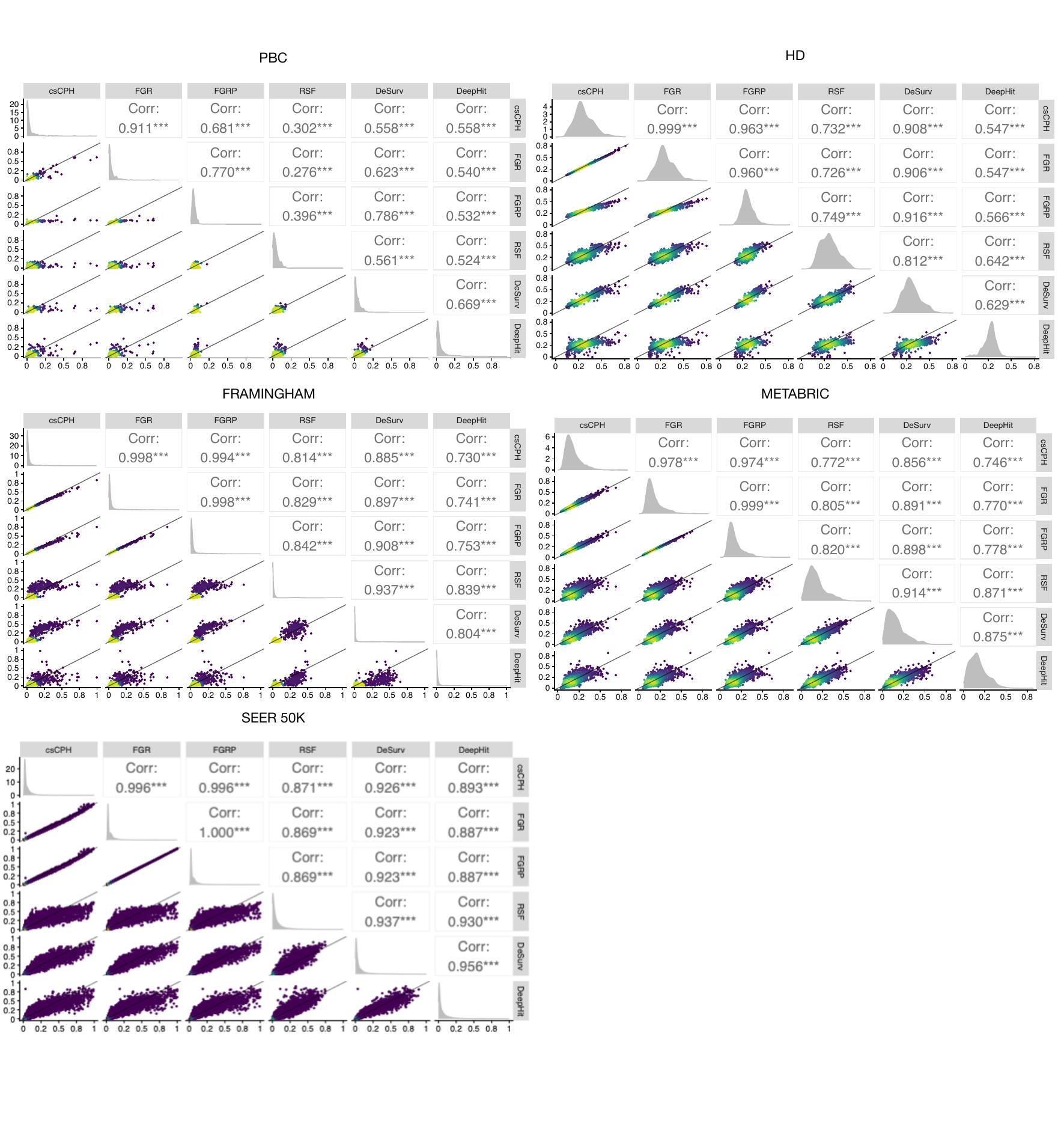}
    \vspace{-15mm}
    \caption{Pairwise comparison of individual-level predicted CIF at a specific time $t =5$ years across benchmarked models for cause 1 for collected datasets. Lower panels show scatterplots of the predicted risks for individuals from the outer test folds, while diagonal panels display the marginal distributions of predictions for each model. Upper panels report the pairwise Pearson correlation coefficients between model predictions; significance stars indicate the corresponding p-value (i.e., $*** \text{ $p$-value} < 0.001)$}
    \label{fig:sup:consistency_cifs}
\end{figure}

\begin{figure}[H]

    \centering
    \includegraphics[width=\linewidth]{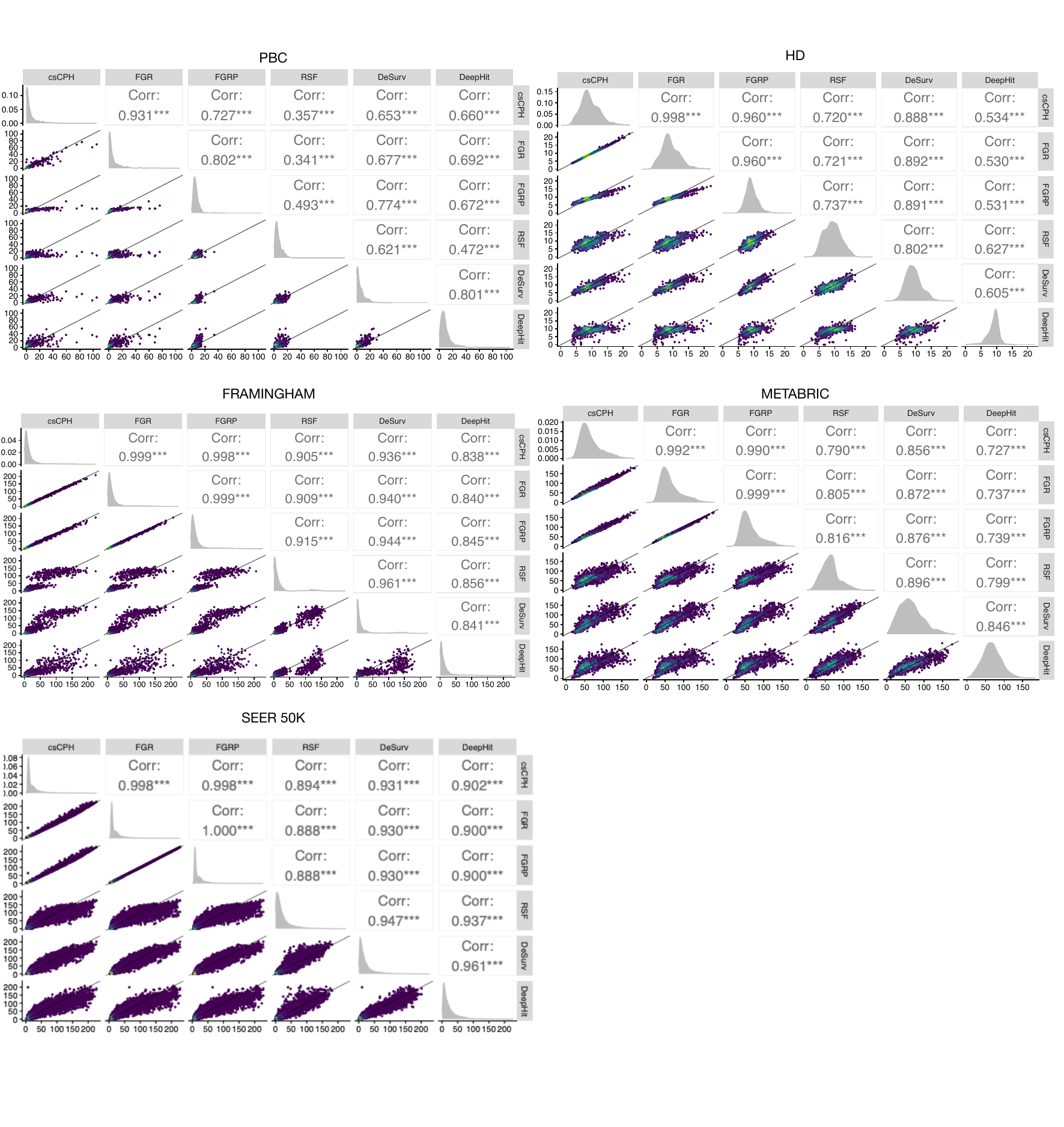}
    \vspace{-15mm}
    \caption{Pairwise comparison of individual-level predicted RMTL years across benchmarked models for cause 1 for collected datasets. Lower panels show scatterplots of the predicted risks for individuals from the outer test folds, while diagonal panels display the marginal distributions of predictions for each model. Upper panels report the pairwise Pearson correlation coefficients between model predictions; significance stars indicate the corresponding p-value (i.e., $*** \text{ $p$-value} < 0.001)$}
    \label{fig:sup:consistency_rmlt}
\end{figure}

\begin{figure}[H]
    \centering
    \includegraphics[width=1\linewidth]{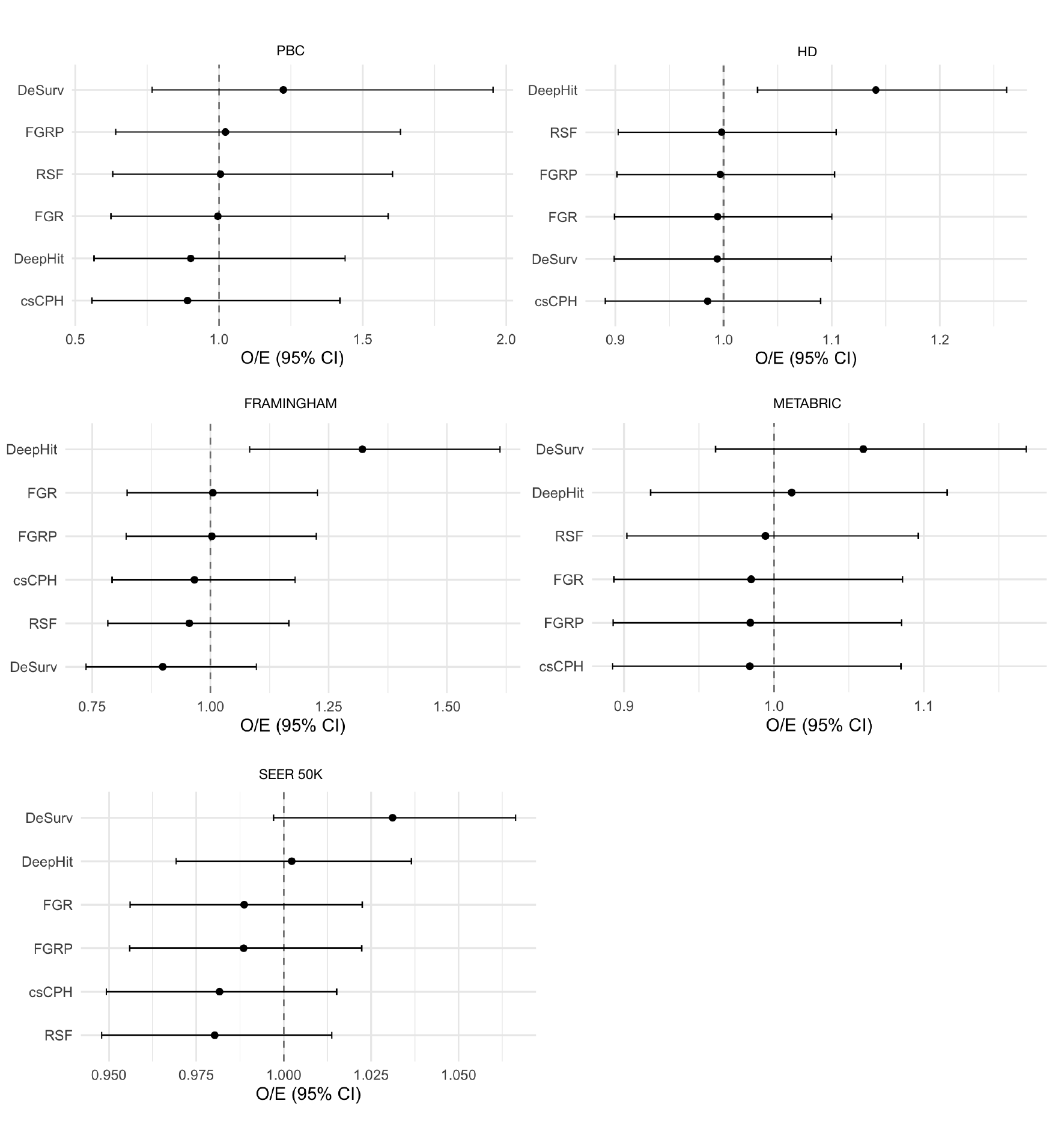}
    \caption{Observed to Expected (O/E) ratio for gathered metrics across models for cause of interest (cause 1) at $t = 5$ years, where deviations from 1 indicate miscalibration at that specific time point. }
    \label{fig:sup:eos}
\end{figure}

\begin{figure}[H]
    \centering
    \includegraphics[width=1\linewidth]{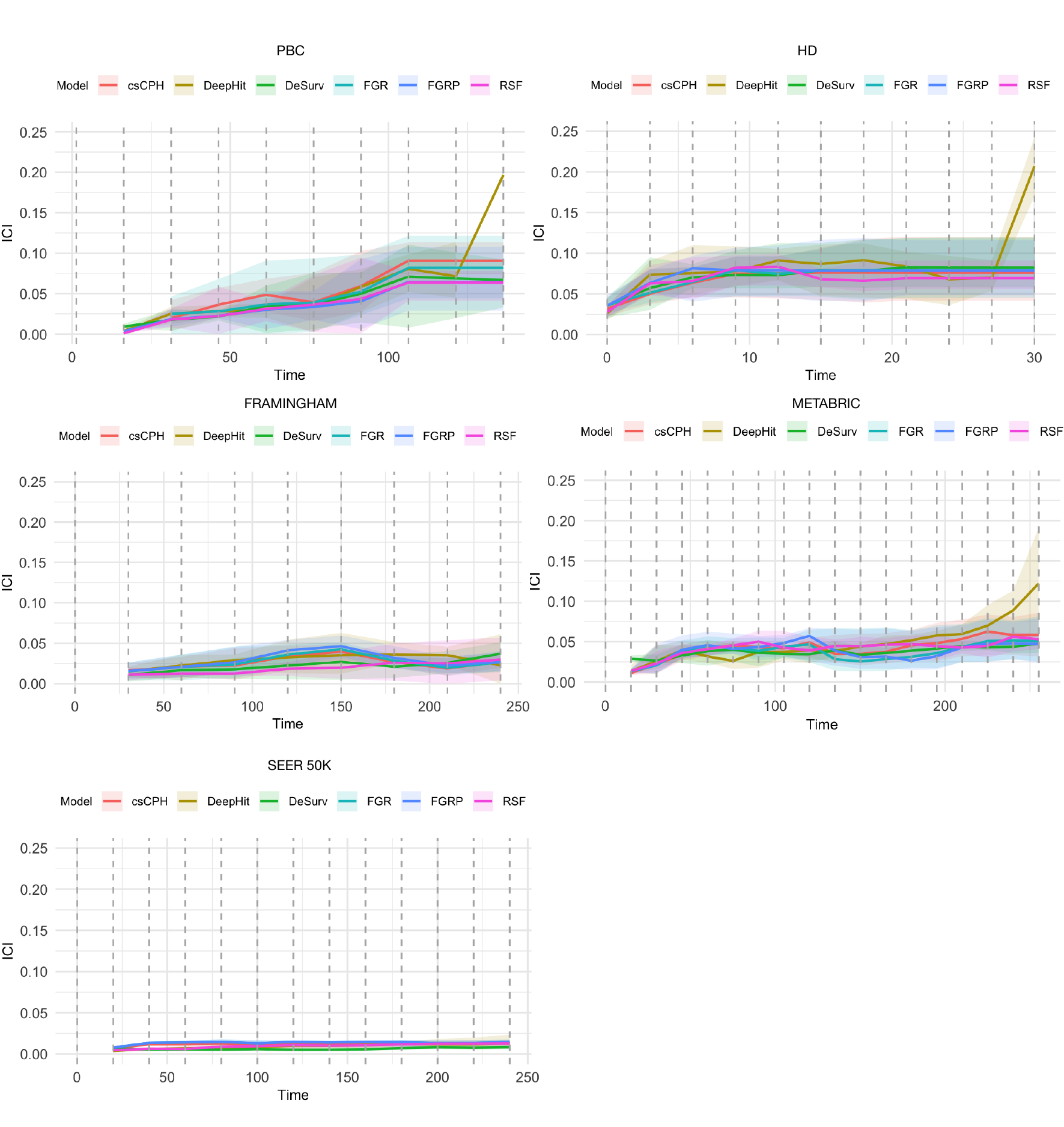}
    \caption{Integrated Calibration Index (ICI) over time for gathered datasets and across models for cause of interest (cause 1). Dotted vertical lines indicate the time grid at which the calibration plots were computed, and subsequently ICI derived from.}
    \label{fig:sup:icis}
\end{figure}

\begin{figure}[H]
    \centering
    \includegraphics[width=1\linewidth]{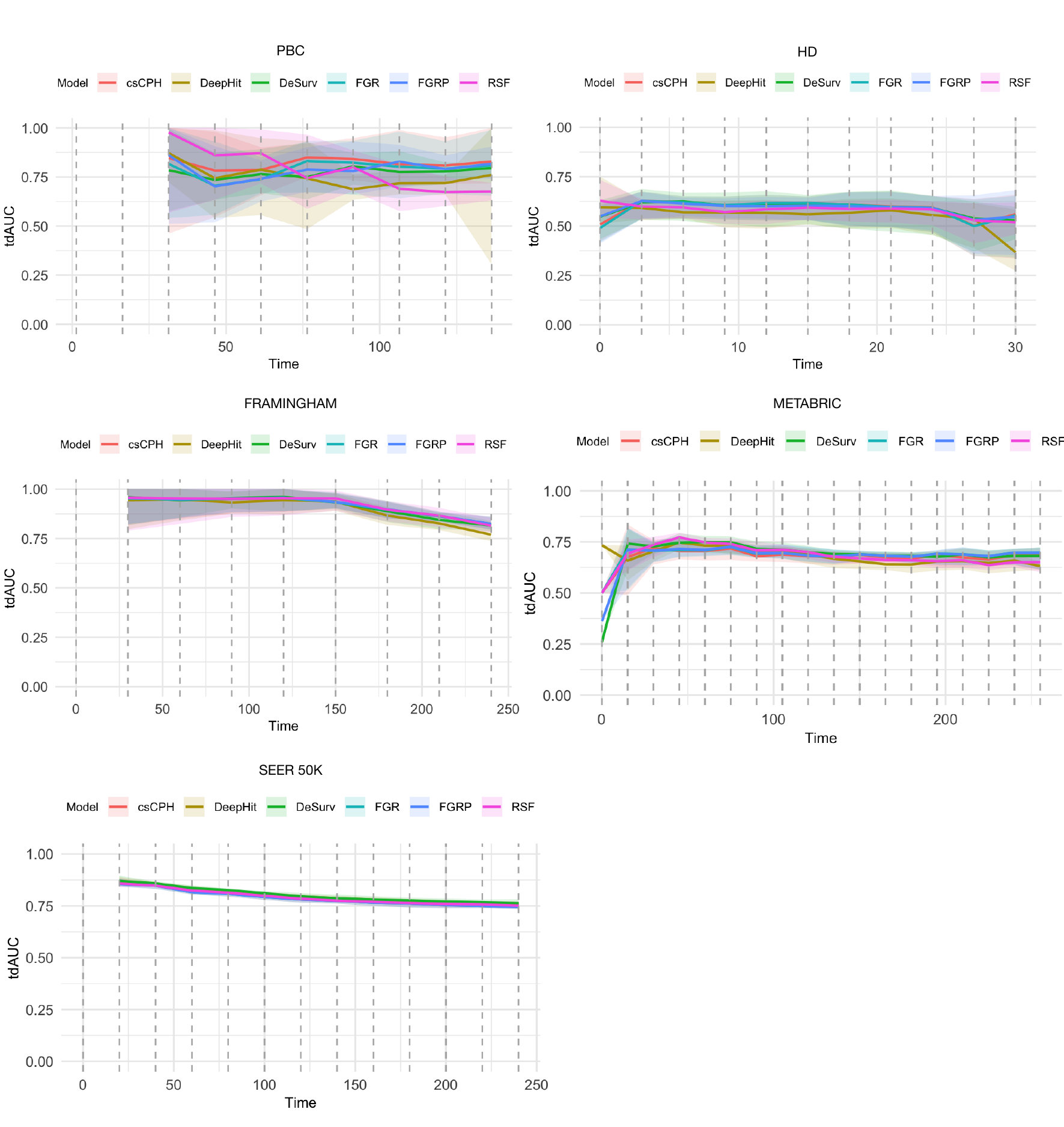}
    \caption{Time-dependent Area Under the Survival Curve (tdAUC) over time across gathered datasets and models for cause of interest (cause 1).}
    \label{fig:sup:aucs}
\end{figure}

\begin{figure}[H]
    \centering
    \includegraphics[width=\textwidth]{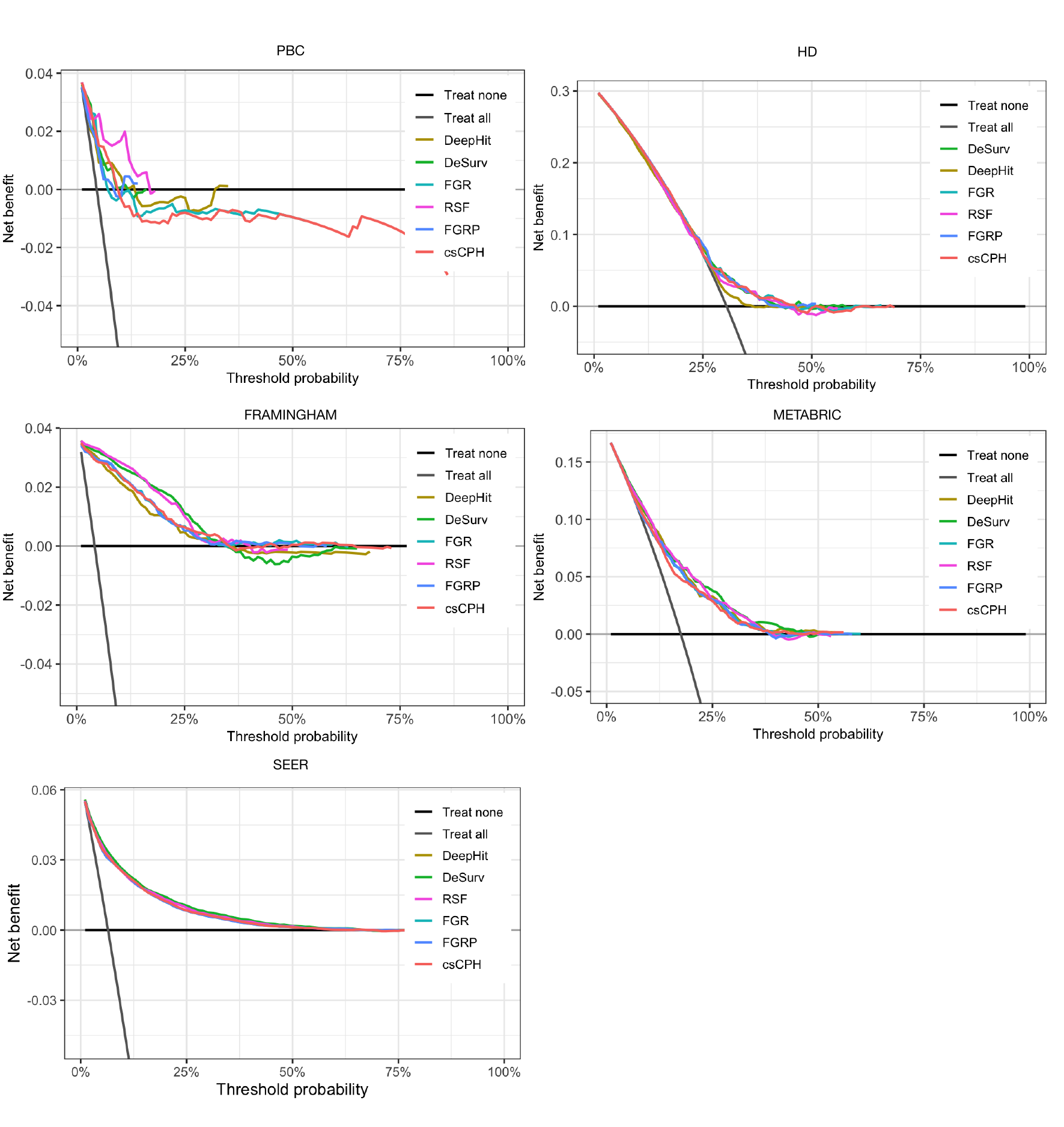}
    \caption{Evaluation of clinical utility at $t$-year metrics of benchmarked models on gathered datasets for cause of interest (cause 1). Net benefit curves over \textit{treat all} and \textit{treat none} indicate positive net benefit at the given threshold probability at a specific time $t=5$.}
    \label{fig:clinical_utilities}
\end{figure}

\begin{figure}[H]
    \centering
    \includegraphics[width=1\textwidth]{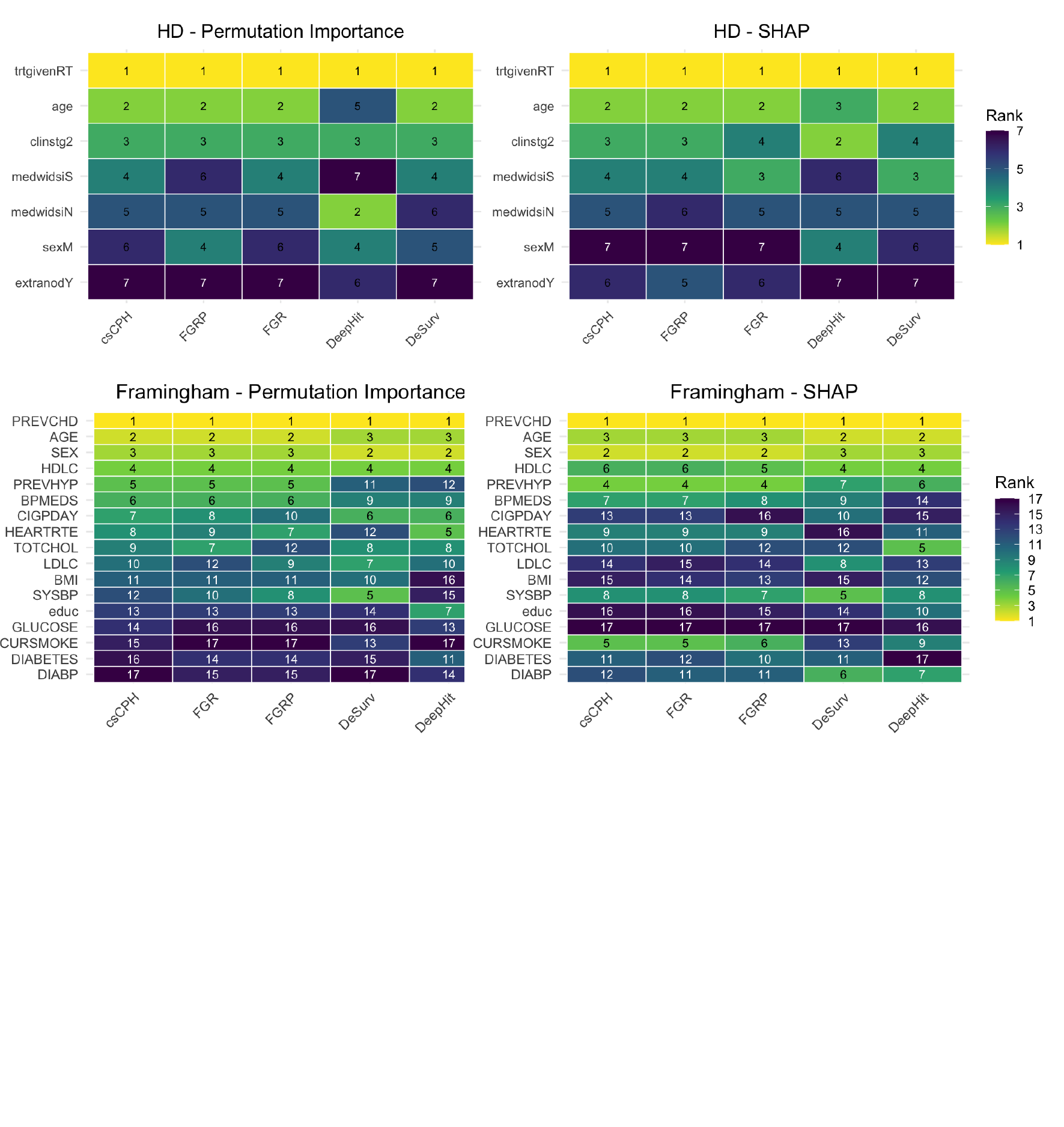}
    \vspace{-60mm}
    \caption{Covariate ranks across models on FRAMINGHAM and HD datasets (rank $=1$ indicates the most important variable or higher contribution). Left figures represent the average permutation importance across folds based on (Integrated Brier Score) IBS. B) Covariate ranks across models based on the absolute values of CifSHAP(t) computed as the mean across individuals, integration across time and subsequent mean across folds.}
    \label{fig:addinterp}
\end{figure}

\begin{figure}[H]
    \centering
    \includegraphics[width=\textwidth]{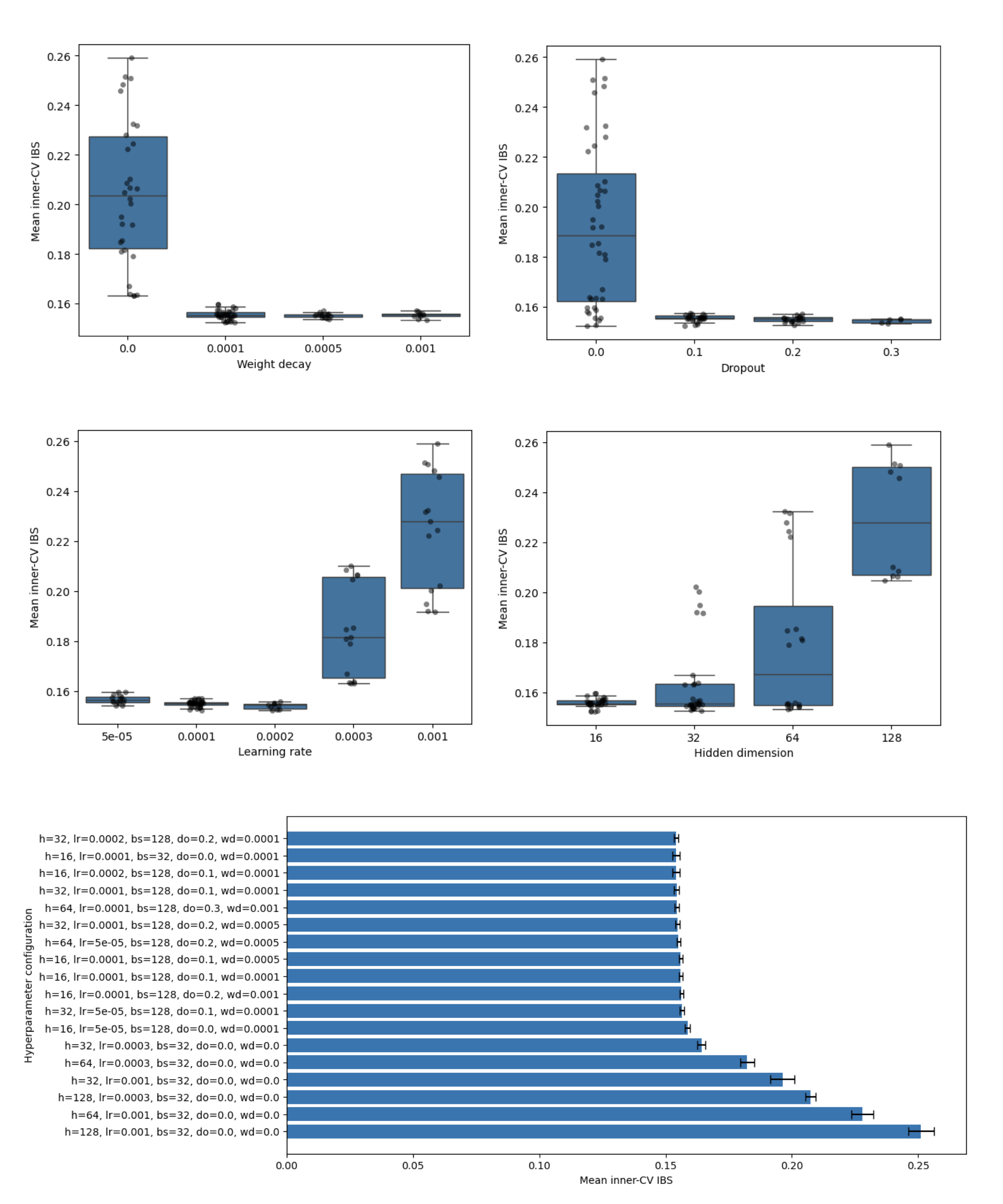}
    \caption{DeSurv performance in inner cross-validation (CV) based on mean IBS across inner folds across different hyper-parameter configurations an fixed number of epochs (i.e., $e = 300$), for METABRIC dataset trained for cause of interest (cause 1). Where $h$ indicates hidden dimensions, $lr$ learning rate, $bs$ batch size, $do$ dropout and $wd$ weight decay.}
    \label{fig:desurv_hyperp}
\end{figure}


\begin{figure}[H]
    \centering
    \includegraphics[width=\textwidth]{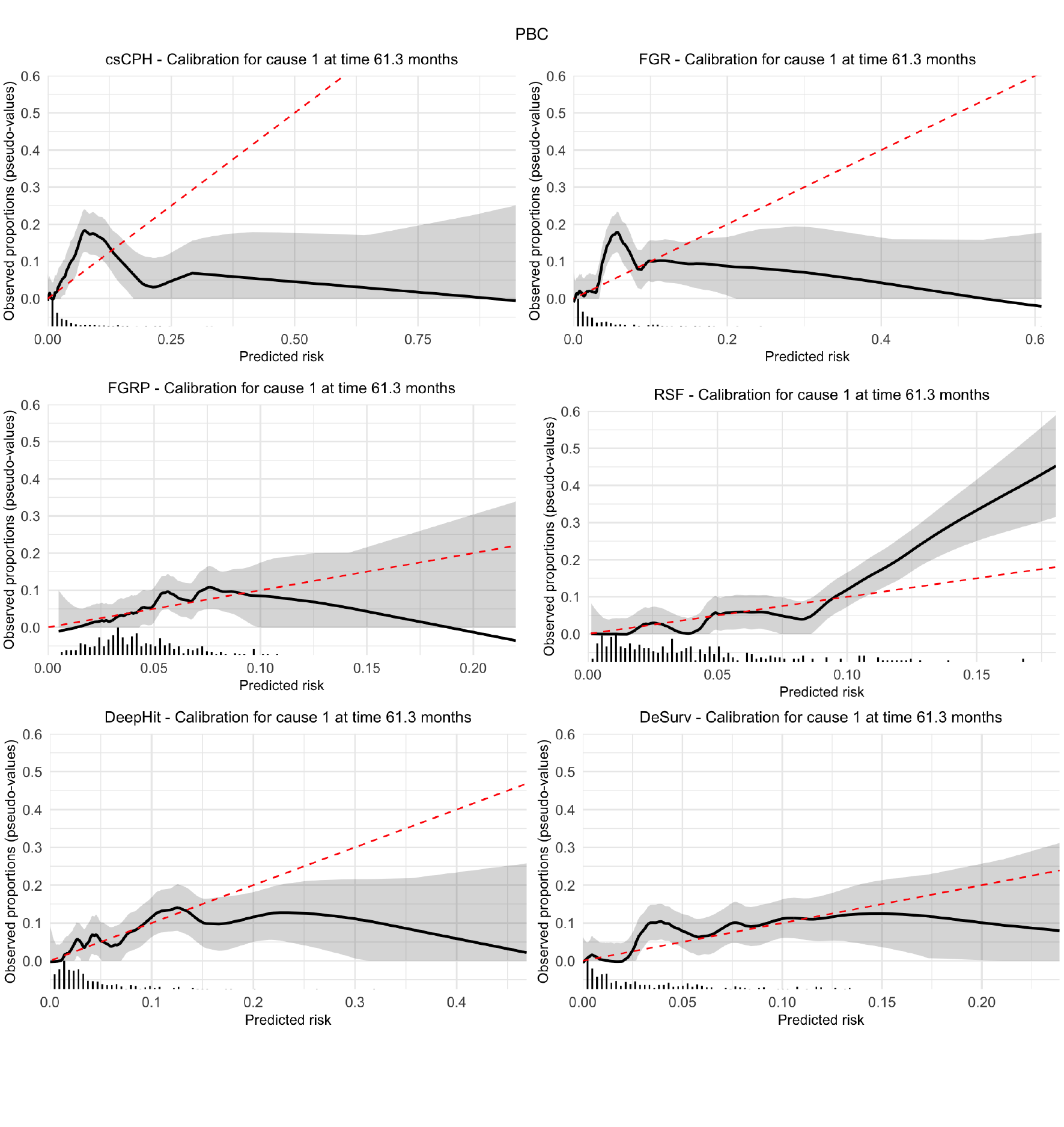}
    \caption{Calibration across models for dataset PBC}
    \label{fig:calpbc}
\end{figure}

\begin{figure}[H]
    \centering
    \includegraphics[width=\textwidth]{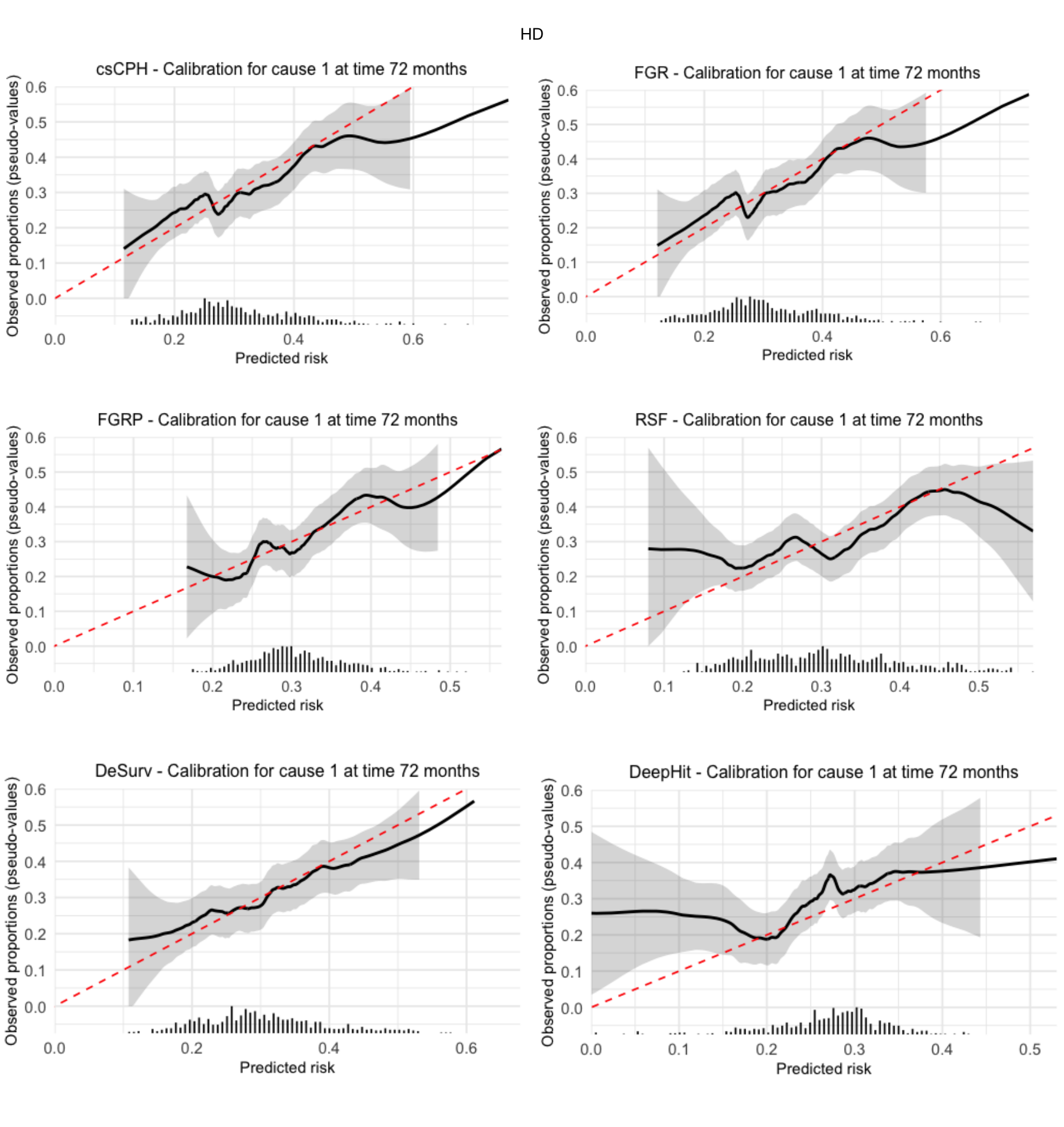}
    \caption{Calibration across models for dataset HD}
    \label{fig:calHD}
\end{figure}

\begin{figure}[H]
    \centering
    \includegraphics[width=\textwidth]{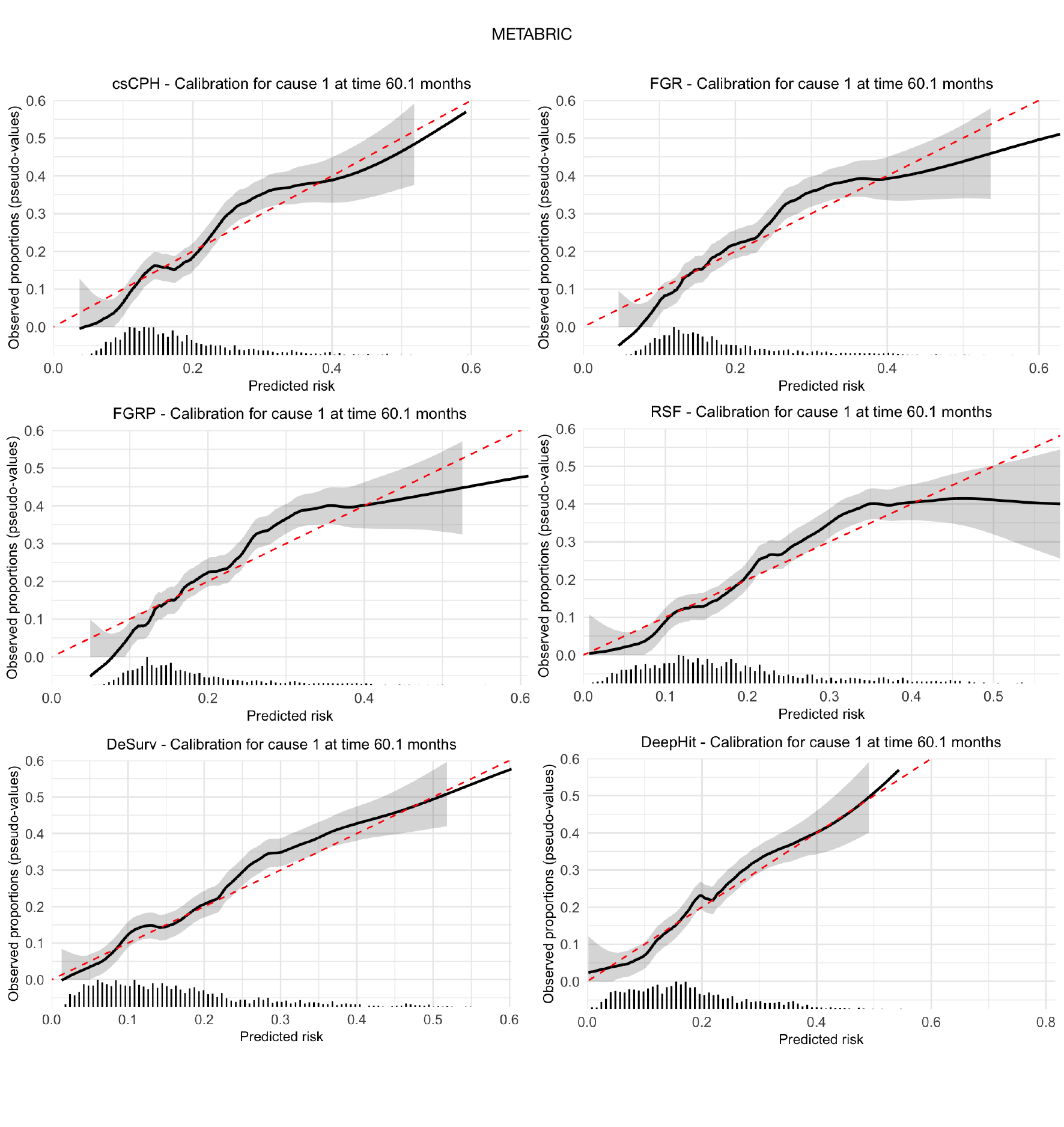}
    \caption{Calibration across models for dataset METABRIC}
    \label{fig:calmet}
\end{figure}

\begin{figure}[H]
    \centering
    \includegraphics[width=\textwidth]{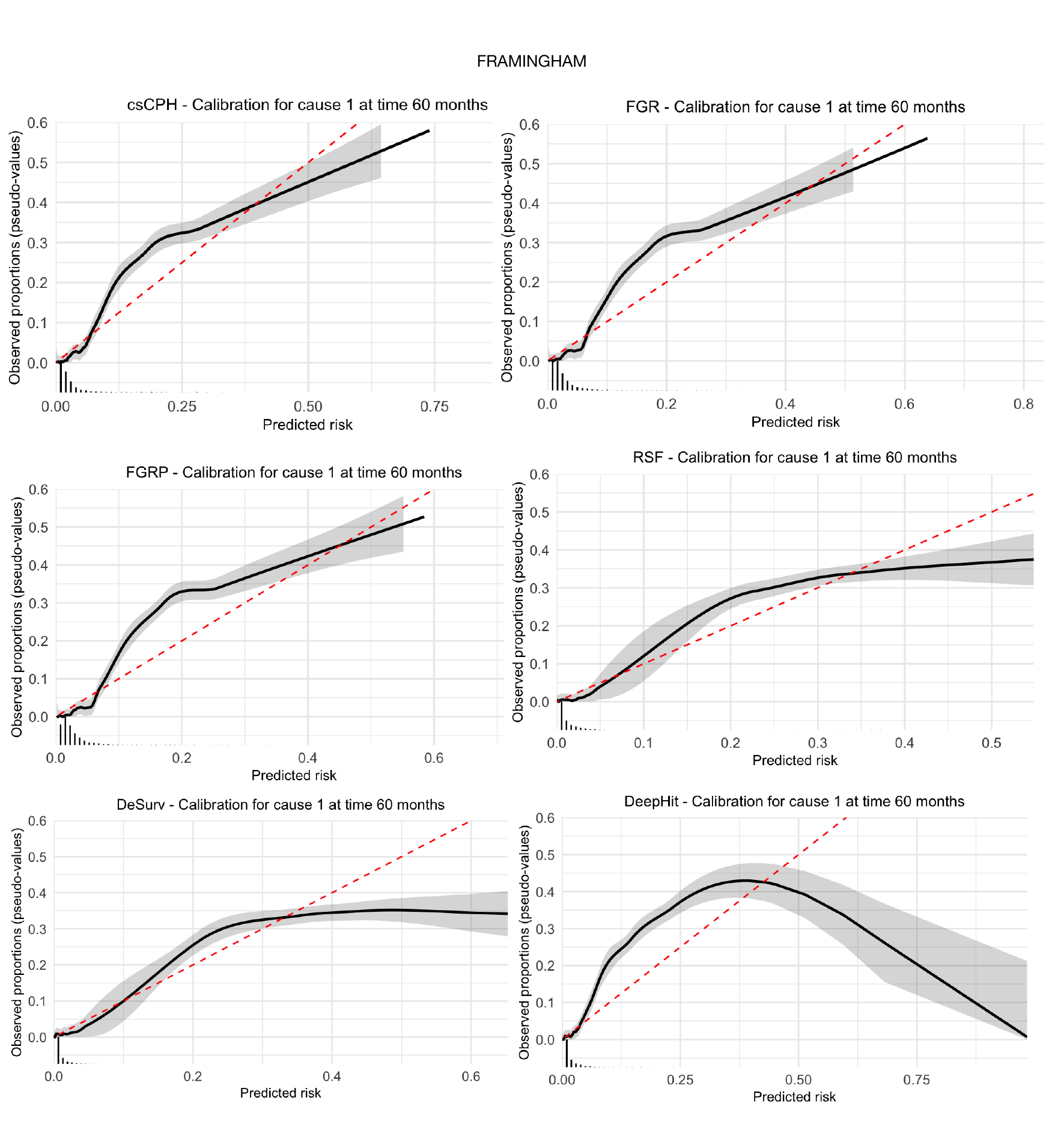}
    \caption{Calibration across models for dataset Framingham}
    \label{fig:calfram}
\end{figure}

\begin{figure}[H]
    \centering
    \includegraphics[width=\textwidth]{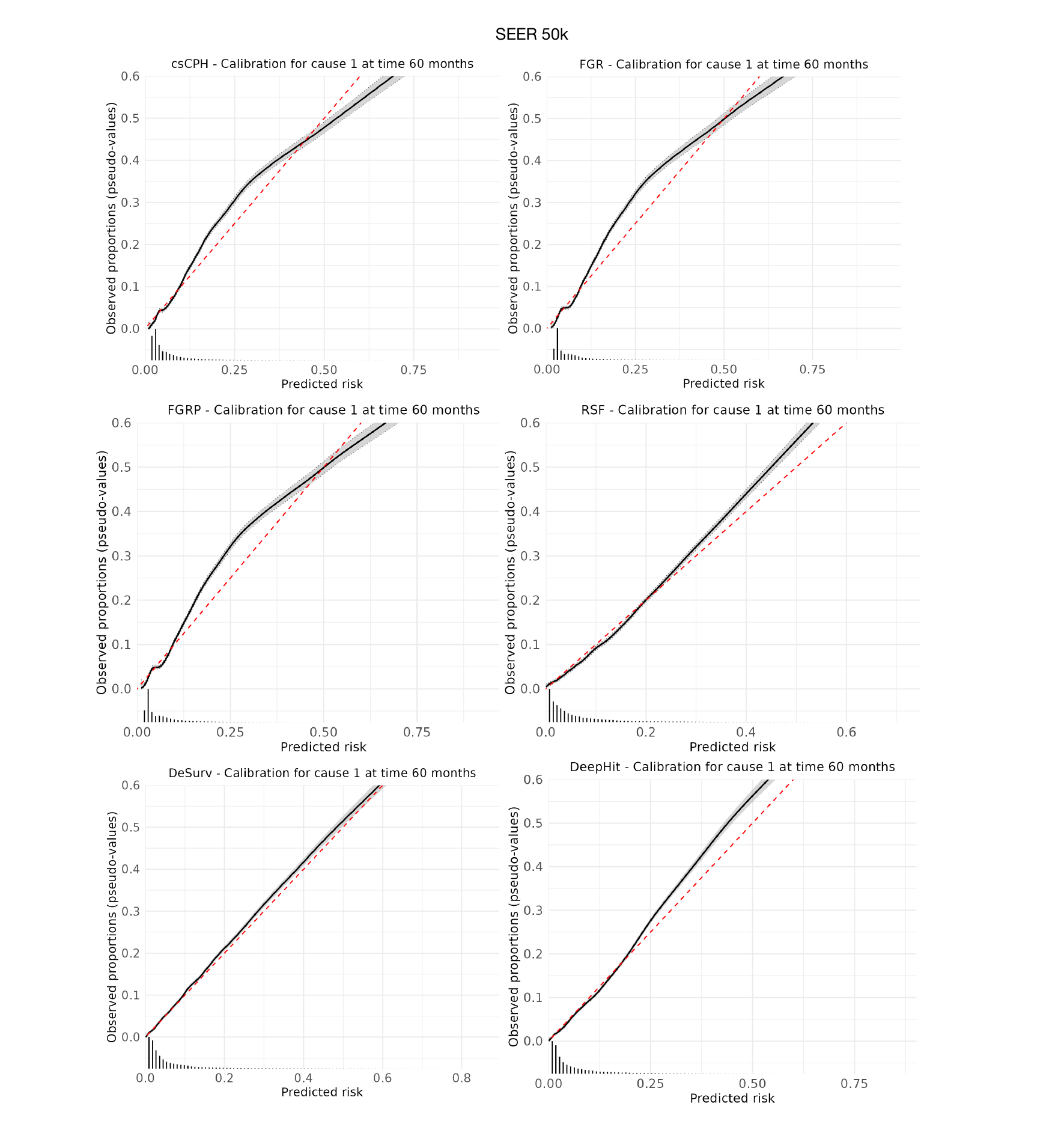}
    \caption{Calibration across models for dataset SEER subset of 50,000 individuals.}
    \label{fig:calseer}
\end{figure}

\subsection{Model Hyper-parameters}\label{supp:hyperp}

The hyper-parameters listed in Table \ref{tab:hp_grids} control model complexity, optimisation, and regularisation. \\
For the penalised subdistribution hazard model, $\lambda$ denotes the ridge regularisation parameter, where larger values produce greater coefficient shrinkage. In Random Survival Forests, \texttt{ntree} specifies the number of trees grown in the ensemble, while \texttt{nodesize} determines the minimum terminal node size, controlling the depth of individual trees. \\
For the neural network models (DeSurv and DeepHit), \texttt{hidden\_dim} specifies the width of the hidden representation, \texttt{lr} is the learning rate used by the optimiser, and \texttt{batch\_size} defines the number of observations processed at each optimisation step. Additional regularisation is provided through \texttt{dropout}, which randomly deactivates hidden units during training, and \texttt{weight\_decay}, corresponding to $L_2$ regularisation of the network weights.\\
For DeepHit, \texttt{num\_layers\_shared} and \texttt{num\_layers\_indiv} specify the number of hidden layers shared across competing events and the number of cause-specific hidden layers, respectively, while \texttt{hidden\_dim\_indiv} defines the width of the cause-specific hidden layers when specified. Batch normalisation (\texttt{batch\_norm}) stabilises training by normalising layer activations, and \texttt{optimizer} determines the optimisation algorithm (Adam or AdamWR). The parameter $\alpha$ controls the balance between the likelihood and pairwise ranking components of the DeepHit loss function, whereas $\sigma$ controls the smoothness of the ranking loss. For the AdamWR optimiser, \texttt{decoupled\_weight\_decay} specifies the regularisation strength independently of the learning rate, while \texttt{cycle\_len}, \texttt{cycle\_multiplier}, and \texttt{cycle\_eta\_multiplier} define the warm-restart schedule by controlling the initial cycle length, the increase in cycle duration, and the reduction in the peak learning rate after each restart.

\renewcommand{\arraystretch}{1.15}
\small

\begin{longtable}{
@{}l
>{\RaggedRight\arraybackslash}p{0.42\linewidth}
>{\RaggedRight\arraybackslash}p{0.48\linewidth}
@{}
}

\caption{Hyper-parameter grids for Random Survival Forest, Penalised Subdistribution Hazard, DeSurv, and DeepHit models. ($^*$) Indicating the fixed hyper-parameter configuration used during runtime measurements.}
\label{tab:hp_grids}
\\

\toprule
ID & \textbf{Main hyper-parameters} & \textbf{Additional parameters} \\
\midrule
\endfirsthead

\caption[]{Hyper-parameter grids for Random Survival Forest, DeSurv, and DeepHit models (continued).}
\\

\toprule
ID & \textbf{Main hyper-parameters} & \textbf{Additional parameters} \\
\midrule
\endhead

\midrule
\multicolumn{3}{r@{}}{\footnotesize Continued on next page}
\\
\endfoot

\bottomrule
\endlastfoot


\multicolumn{3}{@{}l}{\textbf{Penalised Subdistribution Hazard (R)}} \\
\midrule

P1--P30 &
\texttt{$\lambda \in [10^{-4},\,1]$}  (ridge penalisation)&
--- \\


\midrule
\multicolumn{3}{@{}l}{\textbf{Random Survival Forest (R)}} \\
\midrule

R1 &
\texttt{ntree=300,\ nodesize=5} &
\texttt{splitrule=logrankCR} \\

R2 &
\texttt{ntree=300,\ nodesize=15} &
\texttt{splitrule=logrankCR} \\

R3 &
\texttt{ntree=600,\ nodesize=5} &
\texttt{splitrule=logrankCR} \\

R4 ($^*$) &
\texttt{ntree=600,\ nodesize=15} &
\texttt{splitrule=logrankCR} \\


\midrule
\multicolumn{3}{@{}l}{\textbf{DeSurv (Python)}} \\
\midrule

D1 &
\texttt{hidden\_dim=32,\ lr=3e-4,\ batch\_size=32} &
--- \\

D2 &
\texttt{hidden\_dim=32,\ lr=1e-3,\ batch\_size=32} &
--- \\

D3 &
\texttt{hidden\_dim=64,\ lr=3e-4,\ batch\_size=32} &
--- \\

D4 &
\texttt{hidden\_dim=64,\ lr=1e-3,\ batch\_size=32} &
--- \\

D5 &
\texttt{hidden\_dim=128,\ lr=3e-4,\ batch\_size=32} &
--- \\

D6 &
\texttt{hidden\_dim=128,\ lr=1e-3,\ batch\_size=32} &
--- \\

D7 &
\texttt{hidden\_dim=16,\ lr=1e-4,\ batch\_size=32} &
\texttt{dropout=0.0,\ weight\_decay=1e-4} \\

D8 &
\texttt{hidden\_dim=16,\ lr=5e-5,\ batch\_size=128} &
\texttt{dropout=0.0,\ weight\_decay=1e-4} \\

D9 &
\texttt{hidden\_dim=16,\ lr=1e-4,\ batch\_size=128} &
\texttt{dropout=0.1,\ weight\_decay=1e-4} \\

D10 &
\texttt{hidden\_dim=16,\ lr=2e-4,\ batch\_size=128} &
\texttt{dropout=0.1,\ weight\_decay=1e-4} \\

D11 &
\texttt{hidden\_dim=16,\ lr=1e-4,\ batch\_size=128} &
\texttt{dropout=0.1,\ weight\_decay=5e-4} \\

D12 &
\texttt{hidden\_dim=16,\ lr=1e-4,\ batch\_size=128} &
\texttt{dropout=0.2,\ weight\_decay=1e-3} \\

D13 &
\texttt{hidden\_dim=32,\ lr=5e-5,\ batch\_size=128} &
\texttt{dropout=0.1,\ weight\_decay=1e-4} \\

D14 &
\texttt{hidden\_dim=32,\ lr=1e-4,\ batch\_size=128} &
\texttt{dropout=0.1,\ weight\_decay=1e-4} \\

D15 &
\texttt{hidden\_dim=32,\ lr=1e-4,\ batch\_size=128} &
\texttt{dropout=0.2,\ weight\_decay=5e-4} \\

D16 &
\texttt{hidden\_dim=32,\ lr=2e-4,\ batch\_size=128} &
\texttt{dropout=0.2,\ weight\_decay=1e-4} \\

D17 &
\texttt{hidden\_dim=64,\ lr=5e-5,\ batch\_size=128} &
\texttt{dropout=0.2,\ weight\_decay=5e-4} \\

D18 ($^*$)&
\texttt{hidden\_dim=64,\ lr=1e-4,\ batch\_size=128} &
\texttt{dropout=0.3,\ weight\_decay=1e-3} \\


\midrule
\multicolumn{3}{@{}l}{\textbf{DeepHit (Python)}} \\
\midrule

H1 &
\texttt{hidden\_dim=32,\ lr=3e-4,\ batch\_size=128} &
\texttt{num\_layers\_shared=2,\ num\_layers\_indiv=1,\ batch\_norm=True,\ dropout=0.1,\ optimizer=adam,\ weight\_decay=1e-4,\ alpha=0.1,\ sigma=0.1} \\

H2 &
\texttt{hidden\_dim=32,\ lr=1e-3,\ batch\_size=128} &
\texttt{num\_layers\_shared=2,\ num\_layers\_indiv=1,\ batch\_norm=True,\ dropout=0.1,\ optimizer=adam,\ weight\_decay=1e-4,\ alpha=0.3,\ sigma=0.1} \\

H3 &
\texttt{hidden\_dim=64,\ lr=3e-4,\ batch\_size=128} &
\texttt{num\_layers\_shared=2,\ num\_layers\_indiv=1,\ batch\_norm=True,\ dropout=0.2,\ optimizer=adam,\ weight\_decay=5e-4,\ alpha=0.5,\ sigma=0.1} \\

H4 &
\texttt{hidden\_dim=64,\ lr=1e-3,\ batch\_size=128} &
\texttt{num\_layers\_shared=2,\ num\_layers\_indiv=1,\ batch\_norm=True,\ dropout=0.2,\ optimizer=adam,\ weight\_decay=5e-4,\ alpha=0.7,\ sigma=0.1} \\

H5 ($^*$) &
\texttt{hidden\_dim=64,\ lr=1e-4,\ batch\_size=128} &
\texttt{num\_layers\_shared=3,\ num\_layers\_indiv=1,\ batch\_norm=True,\ dropout=0.3,\ optimizer=adam,\ weight\_decay=1e-3,\ alpha=0.3,\ sigma=0.1} \\

H6 &
\texttt{hidden\_dim=32,\ lr=3e-4,\ batch\_size=128} &
\texttt{num\_layers\_shared=2,\ num\_layers\_indiv=1,\ batch\_norm=True,\ dropout=0.3,\ optimizer=adam,\ weight\_decay=1e-3,\ alpha=0.7,\ sigma=0.1} \\

H7 &
\texttt{hidden\_dim=64,\ lr=3e-4,\ batch\_size=128} &
\texttt{num\_layers\_shared=2,\ num\_layers\_indiv=1,\ batch\_norm=True,\ dropout=0.2,\ optimizer=adam,\ weight\_decay=5e-4,\ alpha=0.5,\ sigma=0.05} \\

H8 &
\texttt{hidden\_dim=64,\ lr=3e-4,\ batch\_size=128} &
\texttt{num\_layers\_shared=2,\ num\_layers\_indiv=1,\ batch\_norm=True,\ dropout=0.2,\ optimizer=adam,\ weight\_decay=5e-4,\ alpha=0.5,\ sigma=0.2} \\

H9 &
\texttt{hidden\_dim=32,\ lr=1e-3,\ batch\_size=128} &
\texttt{num\_layers\_shared=2,\ num\_layers\_indiv=1,\ batch\_norm=True,\ dropout=0.2,\ optimizer=adam,\ weight\_decay=1e-4,\ alpha=0.9,\ sigma=0.1} \\

H10 &
\texttt{hidden\_dim=54,\ lr=1e-2,\ batch\_size=32} &
\texttt{num\_layers\_shared=2,\ num\_layers\_indiv=1,\ hidden\_dim\_indiv=32,\ batch\_norm=True,\ dropout=0.1,\ optimizer=adamwr,\ decoupled\_weight\_decay=1e-2,\ cycle\_len=1,\ cycle\_multiplier=2,\ cycle\_eta\_multiplier=0.8,\ alpha=0.2,\ sigma=0.1} \\

\end{longtable}

\subsection{Metrics}\label{supp:metrics}

Performance measures must account for right-censoring and competing risks during model evaluation \citep{vanGeloven2022BMJ}. To account for censoring, often metrics utilise either pseudo-observations or inverse probability of censoring weights (IPCW). Pseudo-observations replace the unknown observed outcome with an estimated subject-specific contribution, which is derived from the jackknife estimator of the CIF (e.g., commonly the Aalen-Johansen estimator). The result is an approximation of the outcome target, which can be directly compared against the model's risk predictions \citep{Andersen2010SMMR}. 
For instance, this method is used for the generation of calibration plots. \\
In a different manner, IPCW correct for the right-censoring bias by up-weighting those individuals that have had an event despite a high probability of being censored. The censoring distribution $G(t) = Pr(C > t)$, that is, the probability of censoring time ($C$) to happen before $t$, is commonly estimated via the Kaplan-Meier estimator of the censoring indicator. IPCW are subsequently estimated by computing the inverse of $G(t)$. For instance some versions of the C-index utilise IPCW \citep{sierra2025cindexmultiverse} or in the Brier Score \citep{gerds2006BiometricalJournal}. 

\subsubsection{Calibration measures: Plots and numerical summaries}\label{supp:calib:measures}

Calibration aims to assess whether the estimated risk matches the observed event proportions. Calibration evaluates whether a model's predicted CIF, $F_k(t)$, or risk estimates, agree with the observed event proportions at different time points $t$ and cause $k$. For instance, suppose a model predicts a 5-year cumulative incidence of 20\% for the $k$-th event type. If the model is well calibrated, approximately a 20\% of the individuals would experience the event $k$ by 5 years. 

To assess calibration across the whole range of predictions at a specific time point $t$, calibration plots are commonly used. Observed event proportions are plotted against predicted risks, where a well-calibrated model produces a curve that follows the diagonal line and deviations indicate miscalibration. 

We can formally define individual true event indicators occurring before a specific time $t$ as $Y_i(t) = I(T_i \leq t, Z_i = k)$, where $Y_i(t) = 1$ if the event $k$ occurred by time $t$ and $Y_i(t) = 0$, otherwise, where true event status at time $t$ for censored patients is unknown. Given a model $M$, a strong calibration is defined as $M(t | \text{X}) = P(Y(t) = 1 | \text{X})$ for all $\text{X}$. Meaning that the predicted risk is equal to the observed outcome proportions in the data.



One can approximate $Y_i(t)$ with the jackknife pseudo-values approach (\cite{Andersen2010SMMR}), which can give a proxy of the observed event indicators at the individual level, including censored patients. This approach can be defined as follows: 

\begin{equation}
    \tilde Y_i(t) = n \hat F_k(t) - (n-1)\hat F_k^{-i}(t)
\end{equation}

where $\hat F_k(t)$ is the Aalen-Johansen estimator of the CIF for cause $k$ and time $t$, and $\hat F_k^{-i}(t)$ is the same estimator computed after leaving out individual $i$. Therefore, this method approximates the unobserved individual contributions to solve censoring limitations and taking into consideration the presence of competing risks. Additionally, a smooth curve can be estimated non-parametrically (e.g \cite{Gerds2014SM}, \cite{vanGeloven2022BMJ})

More detailed numerical summaries can be derived from calibration curves. One such measure is the Integrated Calibration Index (ICI), defined as the average absolute difference between predicted risks and the corresponding observed outcome proportions obtained from the calibration curve. Formally, if $x$ denotes the predicted risks and $x_c$ the corresponding values on the calibration curve, the ICI is given by $\text{ICI}(t) = \mathbb{E}[|x - x_c|]$, the mean absolute error between predicted risks and values of the calibration curve. In addition, summary statistics such as the median absolute error (E50) or other percentiles can be used to describe the distribution of calibration errors across individuals. These summaries will depend on the smoothing applied to obtain the estimated risks. 

A method to summarise the calibration in large by a particular time point $t$ is the ratio of observed and expected outcomes, $O/E$ ratio. Calibration in the large is defined as $E\{M(t|\text{X})\} = P(Y(t)=1)$, meaning that the average predicted risk equals the overall observed event proportion. While calibration plots compare observed and predicted risks across the range of predicted probabilities (i.e., conditional on the risk level), the O/E ratio provides a global summary by comparing the overall observed event proportion, estimated via the Aalen–Johansen estimator, to the average predicted risk across all individuals. 

\subsubsection{Discrimination}

A popular measure of discrimination is the concordance index (C-index) that summarises the model's ability to correctly rank pairs of subjects according to their risk over time. Given a pair of subjects $(i,j)$, observed survival times $(T_i, T_j)$, and covariates $(\mathbf{x}_i, \mathbf{x}_j)$, the C-index can be defined as:

\begin{equation}
    C = P(M(\mathbf{x}_i) > M(\mathbf{x}_j) | T_i < T_j)
\end{equation}

where $M(\cdot)$ are model risk predictions which often are a summary of risk at a specific time point $t$ ($M_t(\mathbf{x}_i)$). However, just as in single-risk settings, the presence of censoring requires the adjustment via IPCW and therefore the truncation of the estimator by $\tau$ to avoid exploding weights \citep{Uno2011}. Therefore, the C-index for competing risk $k$ can be defined as follows: 

\begin{equation}
C_{\tau}
  = P\Bigl(
      M(\mathbf{x}_i) > M(\mathbf{x}_j)
      \,\Big|\,
      Z_i = k,\;
      T_i < \tau,\;
      (T_i < T_j \text{ or } Z_j \notin (0,k))
    \Bigr).
\end{equation}

Alternatively, Antolini's C-index $C_{td}$ formulation, \citep{antolini2005}, can be easily extended for competing risks: 

\begin{equation}
 \label{eq:antolini}
C_{td} = P(F_k(T_i | \mathbf{x}_i) > F_k(T_i |\mathbf{x}_j) | T_i < T_j).  
\end{equation}

Note, that it does not require a summary of the CIF at a specific time point $t$, and utilises CIF distribution. Additionally, it does not adjust for censoring.

As shown by \cite{Blanche2019Biostatistics}, the C-index is not a proper measure for evaluating $t$-year predicted risk because it ranks event times rather than the binary outcome at time $t$, $Y_i(t) = I(T_i \leq t)$. Consequently, accurate $t$-year risk predictions do not necessarily results in a higher C-index. Instead, \cite{Blanche2019Biostatistics} proposed the time-dependent AUC (tdAUC), which directly evaluates discrimination for the binary outcome at time $t$ and can be extended to competing risks as follows:

\begin{equation}
    tdAUC = P(M(\mathbf{x}_i) > M(\mathbf{x}_j) | T_i \leq t, T_j > t)
\end{equation} 








Beyond its lack of properness, the C-index is prone to manipulation ("C-hacking") and may yield contradictory rankings when predicted survival curves overlap \cite{sonabend2022avoiding}. As highlighted by \cite{Lillelund2025}, relying solely on the C-index overemphasises discrimination while ignoring other aspects of predictive performance. Furthermore, recent work has demonstrated a "C-index multiverse", whereby conceptually equivalent estimators implemented in different software produce different results \cite{sierra2025cindexmultiverse}. Although demonstrated in single-risk settings, these inconsistencies also extend to competing risks.


\subsubsection{Cumulative Incidence Function Transformation}\label{supp:cif_transform}


To rank subjects, alternative transformations to $t$-year predicted risk $M_t(\mathbf{x}_i)$ have been proposed to produce a summary that aligns with discrimination across the whole follow-up. For instance, \cite{Ishwaran2008} defined the expected mortality as $M(\mathbf{x}_i) = \sum_{t \in \mathcal{T}} -\log(S(t | \mathbf{x}_i))$ where $\mathcal{T}$ is a set of observed times, and was advocated by \cite{sonabend2022avoiding} to be used as alternative input to the C-index. Given numerical instabilities of expected mortality for high risk individuals at later time points (i.e $S(t) \approx\ 0$), Restricted Mean Survival Time (RMST) has been proposed \cite{sierra2025cindexmultiverse}, i.e.: \begin{equation} \label{eq:RMST}
M(\mathbf{x}_i) = - \text{RMST}_k(\mathbf{x}_i;T^*) = -
\mathbb{E} \left[\min (\tilde{T}_i, T^*) \mid  \mathbf{x}_i \right] = - \int_0^{T^*} S(t \mid \mathbf{x}_i) \, dt.
\end{equation}

Similarly, in the case of competing risks, we could define a $t$-year predicted risk as $M_{t, k}(\mathbf{x}_{i}) = F_{t,k} (t | \mathbf{x}_{i})$, since higher cumulative incidence is indicative of higher risk. However, $F_{t,k} (t | \mathbf{x}_{i})$ has the same limitations as $M_t(\mathbf{x}_i)$ in single risk settings and it is not "proper". An analogous transformation can be computed from the CIF in the competing risk setting named Restricted Mean Time Lost (RMTL), i.e: 


\begin{equation}
\label{eq:RMTL_comp}
M_k(\mathbf{x}_i)
=
\operatorname{RMTL}_k(\mathbf{x}_i;T^*)
=
\mathbb{E}\!\left[
(T^*-T_i)
\,\mathbf{1}\{T_i\leq T^*,\,Z_i=k\}
\mid \mathbf{x}_i
\right]
=
\int_{0}^{T^*}
F_k(t\mid\mathbf{x}_i)\,dt .
\end{equation}

where $F_k(t \mid \mathbf{x}_i)$ is the cause–$k$ cumulative incidence function,
$\mathcal{T}$ is a grid of evaluation times.




\subsubsection{Overall Prediction Error}\label{supp:bs}

The prediction error at a given time point $t$ reflects how closely a model's predicted event probabilities match the observed outcome. In survival analysis, a common choice is the Brier Score (BS), which measures the mean squared difference between predicted probabilities and event indicators. 
In the presence of right-censoring and competing risks, the weighted brier score can be defined as follows:

\begin{equation}
\label{eq:wbs}
\mathrm{BS}_k(t) = \frac{1}{n}\sum_{i=1}^{n}
\left[
I(T_i > t)\,
\frac{\hat{F}_k(t \mid \mathbf{x}_i)^2}{\hat{G}(t)}
+
I(T_i \le t,\, Z_i = k)\,
\frac{\big(1 - \hat{F}_k(t \mid \mathbf{x}_i)\big)^2}{\hat{G}(T_i^{-})}
+
I(T_i \le t,\, Z_i \notin \{0,k\})\,
\frac{\hat{F}_k(t \mid \mathbf{x}_i)^2}{\hat{G}(T_i^{-})}
\right],
\end{equation}

Here, $\hat{F}_k(t \mid \mathbf{x}_i)$ denotes the model-predicted CIF for cause $k$ by time $t$ for individual $i$, and $\hat{G}(t)$ denotes the estimated probability of remaining uncensored beyond time $t$. The inverse terms $1/\hat{G}(t)$ and $1/\hat{G}(T_i^-)$ are IPCW that correct for right-censoring. The Integrated Brier Score (IBS) is the time-averaged version of the weighted Brier Score, obtained by integrating the prediction error over a time interval $[ 0, t_{max}]$, formally:

\begin{equation}
\label{eq:ibs}
\mathrm{IBS}_k = \frac{1}{t_{\max}} \int_{0}^{t_{\max}} \mathrm{BS}_k(t)\, dt
\end{equation}

In practise, since the BS is evaluated at discrete time points, numerical integration via trapezoidal rule is used to approximate the integral. The IBS effectively summarises the model's average overall prediction performance over the follow-up period into a single scalar measure. 



\subsubsection{Clinical Utility}\label{supp:clinicalutility}

To assess clinical utility for the competing risk of interest $k$, the net benefit at $\tau$ can be estimated for a prediction model based on its predicted risk at a threshold probability ($p$). When there is a range of $p$, it is called decision curve. 

The net benefit can be defined as follows: 

\begin{equation}
\mathrm{NB}_k(\tau) = \frac{\mathrm{TP}_k(\tau)}{n} -
\frac{\mathrm{FP}_k(\tau)}{n} \cdot \Bigl( \frac{p}{1-p} \Bigr)
\end{equation}

where $\mathrm{TP}_k(\tau)$ is the true positives and $\mathrm{FP}_k(\tau)$ the false positives. To estimate them, we need to define $\delta_i(\tau, p) = \mathbf{1}\{ \hat{F}_k(\tau \mid x_i) \ge p \}$, where $\delta_i = 1$ when the models predicted risk for that individual is higher than the threshold probability $p$. Then the quantity $\hat{P}(\delta = 1)$, represents the proportion of individuals classified as $\delta = 1$. Whereas $\hat{F}_k(\tau \mid \delta = 1)$ represents the Aalen-Johansen estimator of the cumulative incidence of $k$ for the population in which the predicted risk is higher than the threshold probability $p$. Then the quantities can be defined as follows:

\begin{equation}
\widehat{\mathrm{TP}}_k(\tau) = \hat{F}_k(\tau \mid \delta = 1) \cdot \hat{P}(\delta = 1) \cdot n
\end{equation}

\begin{equation}
\widehat{\mathrm{FP}}_k(\tau) = \left[1 - \hat{F}_k(\tau \mid \delta = 1)\right] \cdot \hat{P}(\delta = 1) \cdot n
\end{equation}

The net benefit of \textit{treat all} and \textit{treat none} are often plotted together with net benefit of the prediction model for comparison. Where \textit{treat all} is given by $\hat{P}(\delta = 1) = 1$, and where \textit{treat none} is given by $\hat{P}(\delta = 1) = 0$.

\subsection{Interpretability}

\subsubsection{Permutation importance}

Permutation importance is based on the change in predictive performance that is observed after randomly permuting the values of each covariate. Here we select the overall prediction error as a target metric. Specifically, we consider the IBS for a cause $k$ defined in Equation \eqref{eq:ibs}. This would capture an aggregate change over time and across all individuals. Permutation importance for a feature $j$ for cause $k$ can be written as follows: 


\begin{equation}
\mathrm{PI}_{j}^{(k)}=\mathrm{IBS}_{k}\!\left(\widehat{F}_{k}^{\pi_j}\right)
-\mathrm{IBS}_{k}\!\left(\widehat{F}_{k}\right)=\Delta\mathrm{IBS}_{k,j}.
\end{equation}


where $\widehat{F}_{k}(t \mid \mathbf{x}_i)$ denotes the predicted CIF for cause $k$, and
$\widehat{F}_{k}^{\pi_j}(t \mid \mathbf{x}_i)$ denotes the corresponding predicted CIF after randomly permuting covariate $j$. Consequently,
$\Delta\mathrm{IBS}_{k,j}$ denotes the increase in the IBS induced by permuting covariate $j$.

\subsubsection{SurvSHAP(t) extension to competing risks: CifSHAP(t)}\label{cifhsap}

\cite{lundberg2017shaplibrary} introduced SHAP (SHapley Additive exPlanations) as a class of additive feature attribution methods to explain model predictions. The method consists on creating a explanation model that decomposes an individual prediction into a baseline prediction and a feature-specific contribution.

Let $f$ be a prediction model and $f(\mathbf{x}_i)$ be the prediction provided to individual $i$ based on observed covariate values $\mathbf{x}_i$. We assume the following linear decomposition: 
\begin{equation}
\label{eq:shap1}
f(\mathbf{x}_i)
=
\phi_0
+
\sum_{j=1}^{p}\phi_j^{(i)}.
\end{equation}

where $p$ is the total number of features (i.e., covariates), $\phi_0$ is the baseline prediction (i.e., corresponding to the average model prediction over a background dataset, often a subset of the training set), and $\phi_j$ is the Shapely value for feature $j$, representing its contribution to the deviation of the prediction for the individual $i$ from the baseline. The SHAP value feature $j$ is defined as:

\begin{equation}
\phi_j = \sum_{S \subseteq \{1,\dots,p\}\setminus \{j\}} 
\frac{|S|!(p-|S|-1)!}{p!} \left[ v(S \cup \{j\}) - v(S) \right]
\end{equation}

where $S$ denotes a subset (i.e., coalition) of features excluding feature $j$. The function $v(S)$ represents the expected prediction of the original model when the features in $S$ are fixed to their observed values for the individual of interest, while the remaining features are averaged over a background dataset. Consequently, $v(S \cup \{j\}) - v(S)$ quantifies the marginal contribution of feature $j$ to the model prediction given that the features in $S$ are known. The weighting term averages this marginal contribution over all possible subsets of features or coalitions, yielding an attribution that satisfies the Shapley axioms. 

Recently, \cite{KRZYZINSKI2023KBS} extended SHAP for survival outcomes, where the decomposition is obtained as a function of time. For a survival function $S(t \mid \mathbf{x}_i)$, this is defined as

\begin{equation}
S(t \mid \mathbf{x}_i) = \phi_0(t) + \sum_{j=1}^p \phi_j(t)
\end{equation}

where a feature contribution $\phi_j(t)$, is the time-dependent Shapley value for feature $j$. These are defined as:

\begin{equation}
\phi_j(t) = \sum_{S \subseteq \{1,\dots,p\}\setminus \{j\}} 
\frac{|S|!(p-|S|-1)!}{p!} 
\left[ v_t(S \cup \{j\}) - v_t(S) \right]
\end{equation}

where

\begin{equation}
v_t(S) = \mathbb{E}_{X_{\bar S}} \left[ S(t \mid \mathbf{x}_S, \mathbf{X}_{\bar S}) \right]
\end{equation}

where $x_{S}$ are the observed values of the features of the coalition $S$ for the individual being explained, and $X_{\bar S}$ denotes the remaining features not included in $S$. The expectation averages the model prediction over these remaining features, thereby quantifying the expected prediction when only features in $S$ are known.

In the CR setting, we consider the target for the decomposition to be the cause-specific cumulative incidence function (CIF). We therefore define the proposed extension, denoted CifSHAP$(t)$, as

\begin{equation}
F_k(t \mid \mathbf{x}) = \phi_0^{(k)}(t) + \sum_{j=1}^p \phi_j^{(k)}(t)
\end{equation}

where $k$ is the competing risk of interest. The $\phi_0^{(k)}(t)$ is the expected cause-specific cumulative incidence function over a background dataset of size $M$, and $\phi_j^{(k)}(t)$ represents the time-dependent Shapley value representing the contribution of feature $j$ to the predicted CIF of cause $k$ at time $t$. 
Then, the time-dependent SHAP value for feature $j$ and cause $k$ is defined as

\begin{equation}
\phi_j^{(k)}(t) = \sum_{S \subseteq \{1,\dots,p\}\setminus \{j\}} 
\frac{|S|!(p-|S|-1)!}{p!} 
\left[ v_t^{(k)}(S \cup \{j\}) - v_t^{(k)}(S) \right]
\end{equation}

where 

\begin{equation}
v_t^{(k)}(S) = \mathbb{E}_{X_{\bar S}} \left[ F_k(t \mid \mathbf{x}_S, \mathbf{X}_{\bar S}) \right]
\end{equation}

In practice, these expectations are approximated using a single background dataset $B$ containing $M$ observations ($B = \{\mathbf{x}^{(1)}, \mathbf{x}^{(2)}, \dots, \mathbf{x}^{(M)}\}$), typically sampled from the training set. For each observation $m$ in the background dataset, the features in coalition $S$ are fixed to the values of the individual being explained, while the remaining features are replaced by those of the $m$-the background observation, creating hybrid instances; $ (\mathbf{x}_S, \mathbf{x}_{\bar S}^{(1)}),\;
(\mathbf{x}_S, \mathbf{x}_{\bar S}^{(2)}),\; \ldots,\; (\mathbf{x}_S, \mathbf{x}_{\bar S}^{(M)})$. The value function is then approximated by

\begin{equation}
    v_t^{(k)}(S) \approx \frac{1}{M} \sum_{m=1}^M F_k(t \mid \mathbf{x}_S, \mathbf{x}_{\bar S}^{(m)})
\end{equation}

and 

\begin{equation}
v_t^{(k)}(S \cup \{j\}) \approx
\frac{1}{M} \sum_{m=1}^M F_k(t \mid \mathbf{x}_{S \cup \{j\}}, \mathbf{x}_{\bar S}^{(m)})
\end{equation}

where $\mathbf{x}_{\bar S}^{(m)}$ denotes the values of the features outside coalition $S$ for the $m$-th individual in the background dataset. 

Computing the exact Shapley values requires evaluating the value function for all $2^p$ possible feature coalitions, which becomes computationally expensive as the number of feature increases. Therefore, we use Kernel SHAP \citep{lundberg2017shaplibrary}, which approximates the Shapley values by sampling a subset of coalitions. 
For each sampled coalition $S$, a hybrid covariate vector is constructed $\tilde{\mathbf{x}}_{S}^{(i,m)} = \left(\mathbf{x}_{S}^{(i)},\mathbf{x}_{\bar S}^{(m)}\right),$
Model predictions across hybrid covariate vectors are averaged to approximate the value function $v_t^{(k)}(S)$. 
Rather than directly averaging the marginal contributions over all possible coalitions, Kernel SHAP estimates the Shapley values by fitting a weighted linear regression. Each sampled coalition is represented by a binary indicator vector of the features present, while the approximated value function serves as a response. The regression is weighted according to the SHAP kernel
\begin{equation}
\pi(S)=\frac{p-1}{\binom{p}{|S|}\,|S|\,(p-|S|)}
\end{equation}
where $|S|$ denotes the number of features in the coalition $S$. As a result, the weight assigned to a coalition depends of its size (i.e., assigning greater weight to very small and very large coalitions, and lower weight to intermediate sizes). In practise, $L$ number of coalitions are sampled where $L \ll 2^{p}$, reducing the computational cost. 

The decomposition itself is not theoretically constrained to produce individual SHAP values within $[0,1]$, since SHAP values are additive feature contributions rather than probabilities. Instead, the probability constraints apply to the reconstructed CIF obtained by summing the baseline prediction and all feature contributions,

\begin{equation}
\underbrace{F_k(t \mid \mathbf{x}_i)}_{\text{model-predicted CIF}}
\approx
\underbrace{
\phi_0^{(k)}(t)
+
\sum_{j=1}^{p}\phi_{j}^{(k)}(t)
}_{\text{CifSHAP reconstruction}}.
\end{equation}

Consequently, the CifSHAP$(t)$ reconstruction should satisfy the same probability constraints as the original model prediction,

\begin{equation}
0 \leq F_k(t \mid \mathbf{x}_i) \leq 1,
\end{equation}

and 

\begin{equation}
\sum_{k=1}^{K} F_k(t \mid \mathbf{x}_i) \leq 1.
\end{equation}

These constrains were empirically verified across all explained patients of METABRIC dataset for DeepHit model in GitHub repository \url{https://github.com/BBolosSierra/CompRisksBenchmark/blob/bego/python/CifSHAP_constrains.ipynb}






To obtain global explanations, SHAP values can be aggregated across individuals either by averaging signed contributions,

$$\bar{\phi}_{j,k}(t) = \frac{1}{N} \sum_{i=1}^{N} \phi_{j,k}^{(i)}(t)$$ 

or by averaging absolute contributions,

$$\bar{\phi}_{j,k}(t) = \frac{1}{N} \sum_{i=1}^{N} |\phi_{j,k}^{(i)}(t)| $$

where the latter quantifies the overall magnitude of feature importance independently of direction. The absolute SHAP values were used for feature ranking.

To summarize feature importance over time, the aggregated SHAP trajectories were integrated over the time horizon. For feature $j$, cause $k$, and outer cross-validation fold $f$, the integrated importance was defined as

$$I_{j,k}^{(f)} = \int \bar{\phi}_{j,k}(t)\,dt$$

Since predictions were evaluated on a discrete time grid, the integral was numerically approximated using the trapezoidal rule.
Integrated SHAP importance values were computed independently for each outer fold,
$\left(
I_{j,k}^{(1)},
I_{j,k}^{(2)},
\dots,
I_{j,k}^{(F)}
\right)$,
and subsequently averaged across folds to obtain a global estimate of feature importance:

$$\bar{I}_{j,k} = \frac{1}{F} \sum_{f=1}^{F} I_{j,k}^{(f)}$$

The variability across folds was further used to compute confidence intervals for the integrated SHAP importance estimates.

In the competing risks setting, the background dataset used to approximate the conditional expectations contains the events in the same proportions as in the training set. Similarly, the set of individuals selected to be explained are in the same proportions as the test set. In this work, the number of individuals to be explained was fixed to 50, with 100 coalitions, while the background dataset is composed by a 15\% of the original dataset. However, a higher percentage can be used to decrease uncertainty. Additionally, the subjects IDs are stored for both the background dataset and explained patients for reproducibility across R and python.

\subsection{Datasets}\label{supp:datasets}

\subsubsection{Open-source datasets}

Loading and preprocessing of each dataset is available in \url{https://github.com/BBolosSierra/CompRisksBenchmark/blob/main/Scripts/download_datasets.Rmd}

The Mayo Clinic Primary Biliary Cholangitis Data (PBC) dataset from the \pkg{survival} R package, consists of 418 patients with primary biliary cirrhosis (Table \ref{tab:tab:pbc}), including 276 complete cases \citep{Therneau2000Modeling}. Individuals entered follow-up at study registration and were observed until liver transplantation, death, or administrative censoring. Within the competing risks framework, liver transplantation was defined as the event of interest (Cause 1), whereas death without prior transplantation was treated as the competing event (Cause 2). Follow-up time was defined as the number of days from registration to the first observed endpoint.

To account for missing data, multiple imputation by chained equations (MICE) was performed using all available covariates. Following imputation, highly collinear variables (\textit{ascites}, \textit{stage}, \textit{hepato}, and \textit{spiders}) were removed prior to model fitting.\\
Imputation was performed independently within each iteration of the nested cross-validation procedure to avoid information leakage. After splitting the original dataset into the outer training and test partitions, the training partition was imputed using MICE, generating $m=3$ completed training datasets after 20 iterations. The fitted imputation models were then applied to the corresponding test partition to obtain three completed test datasets. The same procedure was repeated independently within each inner cross-validation fold used for hyperparameter optimisation.

 The Hodgkin's Disease (HD) dataset originally described in \cite{pintilie2006competing} and directly downloaded from \cite{Monterrubio2024BiometricalJournal}, consists of 865 patients diagnosed with early-stage disease (clinical stage I or II) and treated with either radiation therapy alone (RT) or combined with chemotherapy (CMT). Patients with early-stage HD, typically experience favourable long-term survival. Therefore, the occurrence of late events such as second malignancies becomes particularly relevant. In this dataset, the event of interest is the development of a second malignancy (Cause 1), whereas death without a prior second malignancy constitutes the competing event (Cause 2). The summary statistics for the dataset are included in Table \ref{tab:hd_summary}.
 
The Molecular Taxonomy of Breast Cancer International Consortium (METABRIC) dataset, available through \pkg{cBioPortal} and accessed via the \pkg{cbioPortalData} API \citep{Ramos2020JCO}, consists of 1,936 patients with primary breast cancer (Table \ref{tab:metabric_summary}) \citep{curtis2012}. The dataset is commonly used to benchmark single-risk survival models \citep[e.g.,][]{Katzman2018BMC,DeepHit2018AAAI,danks2022PMLR}, typically without accounting for competing risks. Individuals entered follow-up at diagnosis and were observed until death or administrative censoring. Within the competing risks framework, death due to breast cancer was defined as the event of interest (Cause 1), whereas death from other causes was treated as the competing event (Cause 2). Patients who remained alive at the end of follow-up were considered censored. Follow-up time was defined as the overall survival time, measured in months, from diagnosis to death or censoring.
The analysis included five clinical variables and four gene expression features (\textit{MKI67}, \textit{PGR}, \textit{EGFR}, and \textit{ERBB2}), following the variable selection proposed by \cite{Katzman2018BMC} and subsequently adopted in benchmarking studies implemented in \pkg{pycox}. 






The FRAMINGHAM Heart Study dataset, available through \pkg{riskCommunicator} R package, consists of 2,236 participants with complete covariate information after preprocessing (Table \ref{tab:framingham_summary}). The original dataset contains multiple cardiovascular endpoints that may occur within the same individual, including coronary heart disease (CHD), stroke, and death. Since competing risks analysis requires mutually exclusive first events, a composite competing risks outcome was constructed. 

Cardiovascular disease (CVD) was defined as the event of interest (Cause 1), whereas death without a prior CVD event was treated as the competing event (Cause 2). Follow-up time was defined as the time from study entry to the first occurrence of a CVD event, death, or administrative censoring. When both a CVD event and death were recorded, the earliest event determined the outcome. Individuals experiencing both events on the same day were classified as Cause 1, assuming death was attributable to the cardiovascular event. Participants without either event were administratively censored at their last available follow-up time. 

The competing risks outcome was derived from the original \textit{TIMECVD}, \textit{CVD}, \textit{TIMEDTH}, and \textit{DEATH} variables. Three variables with very low prevalence (\textit{PREVAP}, \textit{PREVMI}, and \textit{PREVSTRK}) were removed prior to benchmarking to improve model stability. Follow-up time was converted from days to months for consistency with the remaining analyses.

\begin{table}[H]
\centering
\small
\caption{
Summary statistics for the PBC dataset. Cause 1 corresponds to liver transplantation, while Cause 2 corresponds to death. Variables marked with ($^*$) were removed prior to benchmarking. Continuous variables are reported as mean (SD), and categorical variables as n (\%).
}
\label{tab:tab:pbc}

\renewcommand{\arraystretch}{0.9}

\begin{tabular}{lccc}
\toprule
\textbf{Variable} &
\makecell{\textbf{Censored}\\(\textbf{$N = 232$}, 55.5\%)} &
\makecell{\textbf{Cause 1}\\(\textbf{$N = 25$}, 5.9\%)} &
\makecell{\textbf{Cause 2}\\(\textbf{$N = 161$}, 38.5\%)}\\
\midrule

\multicolumn{4}{l}{\textbf{Continuous variables}}\\
\midrule

time        & 2,333 (995) & 1,546 (753) & 1,377 (1,049)\\
age         & 50 (10) & 42 (6) & 54 (10)\\
bili        & 1.6 (1.9) & 3.6 (3.6) & 5.5 (5.8)\\
chol        & 327 (166) & 440 (336) & 416 (275)\\
\hspace{3mm}Unknown & 80 & 7 & 47\\
albumin     & 3.59 (0.36) & 3.49 (0.46) & 3.36 (0.47)\\
copper      & 67 (57) & 124 (100) & 135 (98)\\
\hspace{3mm}Unknown & 65 & 6 & 37\\
alk.phos    & 1,578 (1,633) & 1,535 (838) & 2,594 (2,677)\\
\hspace{3mm}Unknown & 64 & 6 & 36\\
ast         & 107 (53) & 130 (37) & 142 (58)\\
\hspace{3mm}Unknown & 64 & 6 & 36\\
trig        & 112 (48) & 134 (71) & 140 (79)\\
\hspace{3mm}Unknown & 81 & 7 & 48\\
platelet    & 261 (89) & 310 (103) & 242 (108)\\
\hspace{3mm}Unknown & 5 & 0 & 6\\
protime     & 10.45 (0.92) & 10.36 (0.54) & 11.19 (1.05)\\
\hspace{3mm}Unknown & 1 & 0 & 1\\

\midrule
\multicolumn{4}{l}{\textbf{Categorical variables}}\\
\midrule

\textbf{trt} &&&\\
\hspace{3mm}1 & 83 (49\%) & 10 (53\%) & 65 (52\%)\\
\hspace{3mm}2 & 85 (51\%) & 9 (47\%) & 60 (48\%)\\
\hspace{3mm}Unknown & 64 & 6 & 36\\

\addlinespace[0.4em]

\textbf{sex} &&&\\
\hspace{3mm}m & 17 (7.3\%) & 3 (12\%) & 24 (15\%)\\
\hspace{3mm}f & 215 (93\%) & 22 (88\%) & 137 (85\%)\\

\addlinespace[0.4em]

\textbf{ascites ($^*$)} &&&\\
\hspace{3mm}0 & 167 (99\%) & 19 (100\%) & 102 (82\%)\\
\hspace{3mm}1 & 1 (0.6\%) & 0 (0\%) & 23 (18\%)\\
\hspace{3mm}Unknown & 64 & 6 & 36\\

\addlinespace[0.4em]

\textbf{hepato} ($^*$) &&&\\
\hspace{3mm}0 & 108 (64\%) & 7 (37\%) & 37 (30\%)\\
\hspace{3mm}1 & 60 (36\%) & 12 (63\%) & 88 (70\%)\\
\hspace{3mm}Unknown & 64 & 6 & 36\\

\addlinespace[0.4em]

\textbf{spiders} ($^*$) &&&\\
\hspace{3mm}0 & 135 (80\%) & 14 (74\%) & 73 (58\%)\\
\hspace{3mm}1 & 33 (20\%) & 5 (26\%) & 52 (42\%)\\
\hspace{3mm}Unknown & 64 & 6 & 36\\

\addlinespace[0.4em]

\textbf{edema} &&&\\
\hspace{3mm}0   & 216 (93\%) & 22 (88\%) & 116 (72\%)\\
\hspace{3mm}0.5 & 15 (6.5\%) & 3 (12\%) & 26 (16\%)\\
\hspace{3mm}1   & 1 (0.4\%) & 0 (0\%) & 19 (12\%)\\

\addlinespace[0.4em]

\textbf{stage} ($^*$) &&&\\
\hspace{3mm}1 & 19 (8.3\%) & 0 (0\%) & 2 (1.3\%)\\
\hspace{3mm}2 & 64 (28\%) & 5 (20\%) & 23 (15\%)\\
\hspace{3mm}3 & 97 (42\%) & 10 (40\%) & 48 (31\%)\\
\hspace{3mm}4 & 50 (22\%) & 10 (40\%) & 84 (54\%)\\
\hspace{3mm}Unknown & 2 & 0 & 4\\

\bottomrule
\end{tabular}
\end{table}

\begin{table}[H]
\centering
\small
\caption{
Summary statistics for the Hodgkin's Disease (HD) dataset. Cause 1 corresponds to the occurrence of a second malignancy, while Cause 2 corresponds to death without a prior second malignancy. Continuous variables are reported as mean (SD), and categorical variables as n (\%).
}
\label{tab:hd_summary}

\renewcommand{\arraystretch}{1}

\begin{tabular}{lccc}
\toprule
\textbf{Variable} &
\makecell{\textbf{Censored}\\(\textbf{$N = 439$}, 50.8\%)} &
\makecell{\textbf{Cause 1}\\(\textbf{$N = 291$}, 33.6\%)} &
\makecell{\textbf{Cause 2}\\(\textbf{$N = 135$}, 15.6\%)}\\
\midrule

\multicolumn{4}{l}{\textbf{Continuous variables}}\\
\midrule

age  & 30 (10) & 38 (17) & 47 (17)\\
time & 20 (6) & 3 (3) & 13 (7)\\

\midrule
\multicolumn{4}{l}{\textbf{Categorical variables}}\\
\midrule

\textbf{sex} &&&\\
\hspace{3mm}F & 223 (51\%) & 132 (45\%) & 47 (35\%)\\
\hspace{3mm}M & 216 (49\%) & 159 (55\%) & 88 (65\%)\\

\addlinespace[0.4em]

\textbf{trtgiven} &&&\\
\hspace{3mm}CMT & 146 (33\%) & 61 (21\%) & 42 (31\%)\\
\hspace{3mm}RT  & 293 (67\%) & 230 (79\%) & 93 (69\%)\\

\addlinespace[0.4em]

\textbf{medwidsi} &&&\\
\hspace{3mm}L & 67 (15\%) & 36 (12\%) & 10 (7.4\%)\\
\hspace{3mm}N & 201 (46\%) & 171 (59\%) & 92 (68\%)\\
\hspace{3mm}S & 171 (39\%) & 84 (29\%) & 33 (24\%)\\

\addlinespace[0.4em]

\textbf{extranod} &&&\\
\hspace{3mm}N & 398 (91\%) & 263 (90\%) & 125 (93\%)\\
\hspace{3mm}Y & 41 (9.3\%) & 28 (9.6\%) & 10 (7.4\%)\\

\addlinespace[0.4em]

\textbf{clinstg} &&&\\
\hspace{3mm}1 & 141 (32\%) & 97 (33\%) & 58 (43\%)\\
\hspace{3mm}2 & 298 (68\%) & 194 (67\%) & 77 (57\%)\\

\bottomrule
\end{tabular}
\end{table}

\begin{table}[H]
\small
\centering
\caption{
Summary statistics for the METABRIC dataset. Cause 1 corresponds to breast cancer-specific death, while Cause 2 corresponds to death from other causes during follow-up. MKI67, PGR, EGFR, and ERBB2 denote gene expression measurements. Treatment variables include chemotherapy, hormone therapy, and radiotherapy. ER\_IHC denotes estrogen receptor status assessed by immunohistochemistry. Continuous variables are reported as mean (SD), and categorical variables as n (\%).
}
\label{tab:metabric_summary}

\renewcommand{\arraystretch}{1}

\begin{tabular}{lccc}
\toprule
\textbf{Variable} & 
\makecell{\textbf{Censored} \\ (\textbf{$N = 812$}, 41.9\%)} & 
\makecell{\textbf{Cause 1} \\ (\textbf{$N = 640$}, 33.0\%)} & 
\makecell{\textbf{Cause 2} \\ (\textbf{$N = 484$}, 25.0\%)} \\
\midrule

\multicolumn{4}{l}{\textbf{Continuous variables}} \\
\midrule

MKI67                  & 5.83 (0.35) & 5.95 (0.34) & 5.82 (0.30) \\
PGR                    & 6.31 (1.06) & 6.03 (0.88) & 6.40 (1.08) \\
EGFR                   & 6.26 (0.82) & 6.29 (0.98) & 6.02 (0.69) \\
ERBB2                  & 10.64 (1.35) & 10.96 (1.55) & 10.70 (1.04) \\
AGE\_AT\_DIAGNOSIS     & 56 (11) & 60 (14) & 70 (10) \\
OS\_MONTHS             & 159 (71) & 78 (60) & 130 (70) \\

\midrule
\multicolumn{4}{l}{\textbf{Categorical variables}} \\
\midrule

\textbf{CHEMOTHERAPY} & & & \\
\hspace{3mm} 0 & 623 (77\%) & 441 (69\%) & 461 (95\%) \\
\hspace{3mm} 1 & 189 (23\%) & 199 (31\%) & 23 (4.8\%) \\

\addlinespace[0.4em]

\textbf{ER\_IHC} & & & \\
\hspace{3mm} 0 & 193 (24\%) & 186 (29\%) & 60 (12\%) \\
\hspace{3mm} 1 & 619 (76\%) & 454 (71\%) & 424 (88\%) \\

\addlinespace[0.4em]

\textbf{HORMONE\_THERAPY} & & & \\
\hspace{3mm} 0 & 319 (39\%) & 252 (39\%) & 160 (33\%) \\
\hspace{3mm} 1 & 493 (61\%) & 388 (61\%) & 324 (67\%) \\

\addlinespace[0.4em]

\textbf{RADIO\_THERAPY} & & & \\
\hspace{3mm} 0 & 271 (33\%) & 252 (39\%) & 249 (51\%) \\
\hspace{3mm} 1 & 541 (67\%) & 388 (61\%) & 235 (49\%) \\

\bottomrule
\end{tabular}
\end{table}

\begin{table}[H]
\small
\centering
\caption{
Summary statistics for the Framingham dataset. Cause 1 corresponds to cardiovascular disease, while Cause 2 corresponds to death from other causes. Variables marked with ($^*$) were removed prior to benchmarking due to a prevalence below 3\%. Continuous variables are reported as mean (SD), and categorical variables as n (\%).
}
\label{tab:framingham_summary}

\renewcommand{\arraystretch}{1}
\small
\begin{tabular}{lccc}
\toprule
\textbf{Variable} &
\makecell{\textbf{Censored}\\(\textbf{$N = 1500$}, 67.1\%)} &
\makecell{\textbf{Cause 1}\\(\textbf{$N = 523$}, 23.4\%)} &
\makecell{\textbf{Cause 2}\\(\textbf{$N = 213$}, 9.5\%)} \\
\midrule

\multicolumn{4}{l}{\textbf{Continuous variables}}\\
\midrule

TOTCHOL   & 238 (44) & 239 (48) & 230 (45)\\
AGE        & 58 (8) & 64 (8) & 64 (9)\\
SYSBP      & 135 (21) & 148 (23) & 144 (23)\\
DIABP      & 80 (11) & 83 (12) & 81 (13)\\
CIGPDAY    & 7 (12) & 7 (11) & 9 (13)\\
BMI         & 25.7 (3.8) & 26.2 (4.0) & 25.3 (4.3)\\
HEARTRTE   & 77 (12) & 79 (13) & 80 (13)\\
GLUCOSE    & 87 (23) & 95 (42) & 90 (26)\\
HDLC        & 50 (16) & 45 (15) & 50 (16)\\
LDLC        & 182 (46) & 188 (48) & 175 (48)\\
CR\_time    & 288 (4) & 157 (86) & 223 (40)\\

\midrule
\multicolumn{4}{l}{\textbf{Categorical variables}}\\
\midrule

\textbf{SEX} &&&\\
\hspace{3mm}1 & 546 (36\%) & 321 (61\%) & 104 (49\%)\\
\hspace{3mm}2 & 954 (64\%) & 202 (39\%) & 109 (51\%)\\

\addlinespace[0.4em]

\textbf{CURSMOKE} &&&\\
\hspace{3mm}0 & 985 (66\%) & 345 (66\%) & 126 (59\%)\\
\hspace{3mm}1 & 515 (34\%) & 178 (34\%) & 87 (41\%)\\

\addlinespace[0.4em]

\textbf{DIABETES} &&&\\
\hspace{3mm}0 & 1,441 (96\%) & 446 (85\%) & 191 (90\%)\\
\hspace{3mm}1 & 59 (3.9\%) & 77 (15\%) & 22 (10\%)\\

\addlinespace[0.4em]

\textbf{BPMEDS} &&&\\
\hspace{3mm}0 & 1,332 (89\%) & 403 (77\%) & 178 (84\%)\\
\hspace{3mm}1 & 168 (11\%) & 120 (23\%) & 35 (16\%)\\

\addlinespace[0.4em]

\textbf{educ} &&&\\
\hspace{3mm}1 & 509 (34\%) & 231 (44\%) & 99 (46\%)\\
\hspace{3mm}2 & 488 (33\%) & 157 (30\%) & 59 (28\%)\\
\hspace{3mm}3 & 294 (20\%) & 69 (13\%) & 28 (13\%)\\
\hspace{3mm}4 & 209 (14\%) & 66 (13\%) & 27 (13\%)\\

\addlinespace[0.4em]

\textbf{PREVCHD} &&&\\
\hspace{3mm}0 & 1,449 (97\%) & 348 (67\%) & 196 (92\%)\\
\hspace{3mm}1 & 51 (3.4\%) & 175 (33\%) & 17 (8.0\%)\\

\addlinespace[0.4em]

\textbf{PREVAP ($^*$)} &&&\\
\hspace{3mm}0 & 1,457 (97\%) & 410 (78\%) & 196 (92\%)\\
\hspace{3mm}1 & 43 (2.9\%) & 113 (22\%) & 17 (8.0\%)\\

\addlinespace[0.4em]

\textbf{PREVMI ($^*$)} &&&\\
\hspace{3mm}0 & 1,497 (100\%) & 414 (79\%) & 210 (99\%)\\
\hspace{3mm}1 & 3 (0.2\%) & 109 (21\%) & 3 (1.4\%)\\

\addlinespace[0.4em]

\textbf{PREVSTRK ($^*$)} &&&\\
\hspace{3mm}0 & 1,500 (100\%) & 478 (91\%) & 213 (100\%)\\
\hspace{3mm}1 & 0 (0\%) & 45 (8.6\%) & 0 (0\%)\\

\addlinespace[0.4em]

\textbf{PREVHYP} &&&\\
\hspace{3mm}0 & 746 (50\%) & 114 (22\%) & 68 (32\%)\\
\hspace{3mm}1 & 754 (50\%) & 409 (78\%) & 145 (68\%)\\

\bottomrule
\end{tabular}
\end{table}

\newpage
\subsubsection{SEER breast cancer dataset}\label{supp:seer}

The breast cancer cohort was obtained from SEER using SEER*Stat. The database used was: \textit{Incidence - SEER Research Data, 17 Registries, Nov 2024 Sub (2000--2022) - Linked To County Attributes - Time Dependent (1990--2023) Income/Rurality, 1969--2023 Counties}, released April 2025 and based on the November 2024 submission. The data were downloaded using a case-listing session.


\begin{enumerate}
    \item Open SEER*Stat and create a new \textit{Case Listing} session.
    \item Select the database: \textit{Incidence - SEER Research Data, 17 Registries, Nov 2024 Sub (2000--2022)}.
    \item In the \textit{Selection} tab, restrict the cohort to:
    \begin{itemize}
        \item age at diagnosis: 15 to $>90$ years;
        \item sex: male and female;
        \item year of diagnosis: 2000--2022;
        \item site: breast using \textit{Site recode ICD-O-3/WHO 2008}.
    \end{itemize}
    \item In the \textit{Table} tab, export patient identifiers, demographic variables, tumour characteristics, staging variables, treatment variables, survival time, cause-of-death variables, income, and rurality variables.
    \item Export the case-listing output for downstream processing.
\end{enumerate}

The initial export contained 1,365,310 patients. Missing, unknown, borderline/unknown, unstaged, and equivalent values were recoded as missing. The analysis cohort was then restricted to patients diagnosed between 2004 and 2015 with follow-up available until 2020.

Covariate selection was performed before complete-case filtering. Variables were removed if they were redundant, highly missing, highly imbalanced, or had excessive category granularity. Sex was removed because male patients represented only 0.7\% of the cohort. AJCC T, N, M and stage components were removed in favour of a summary staging variable. Collaborative Stage variables were also removed because they were highly missing and largely duplicated information captured by other staging variables.

Additional redundant variables were removed when they were constant, duplicated another field, or were not suitable for modelling. These included site recode variables, duplicated primary-site fields, histologic type due to class imbalance, HER2 status due to missingness, and cause-specific death classification variables not used as predictors. 

Several variables were simplified. Age was grouped into three categories: 15--40, 41--69, and 70+ years. Laterality was reduced to left and right. Marital status was reduced to married versus not married. Surgery was recoded as yes/no. Tumour size, regional nodes positive, and surgery variables were converted from SEER recodes into simplified numeric or binary variables where possible. Income intervals were converted to approximate numeric values using interval midpoints, and rural-urban codes were recoded from 1 (most urban) to 5 (most rural).

Duplicate patient IDs occurred because multiple tumours may be recorded for the same patient. The total number of in situ/malignant tumours was retained as a summary of prior or co-occurring tumours, and duplicated patients were reduced by keeping the last recorded occurrence. After removing rows with missing values in the retained variables, the final complete-case dataset contained 702,003 patients.

\renewcommand{\arraystretch}{1}
\small
\begin{longtable}{p{0.43\linewidth}ccc}
\caption{
Summary statistics for the SEER breast cancer dataset. Cause 1 corresponds to breast cancer-specific death, while Cause 2 corresponds to death from other causes. Continuous variables are reported as mean (SD), and categorical variables as n (\%).}
\label{tab:seer_summary}\\
\toprule
\textbf{Variable} &
\makecell{\textbf{Censored}\\(\textbf{$N = 471{,}902$}, 67.2\%)} &
\makecell{\textbf{Cause 1}\\(\textbf{$N = 81{,}636$}, 11.6\%)} &
\makecell{\textbf{Cause 2}\\(\textbf{$N = 148{,}465$}, 21.1\%)}\\
\midrule
\endfirsthead

\toprule
\textbf{Variable} &
\makecell{\textbf{Censored}\\(\textbf{$N = 471{,}902$}, 67.2\%)} &
\makecell{\textbf{Cause 1}\\(\textbf{$N = 81{,}636$}, 11.6\%)} &
\makecell{\textbf{Cause 2}\\(\textbf{$N = 148{,}465$}, 21.1\%)}\\
\midrule
\endhead

\midrule
\multicolumn{4}{r}{\textit{Continued on next page}}\\
\endfoot

\bottomrule
\endlastfoot

\multicolumn{4}{l}{\textbf{Continuous variables}}\\
\midrule

Tumor Size Over Time Recode (1988+) 
& 19 (15) 
& 33 (22) 
& 20 (16)\\

Regional nodes positive (1988+) 
& 1 (2) 
& 4 (6) 
& 1 (3)\\

Survival months 
& 141 (60) 
& 66 (49) 
& 102 (61)\\

Total number of in situ/malignant tumors for patient 
& 1.28 (0.58) 
& 1.34 (0.62) 
& 1.56 (0.78)\\

Median household income inflation adj to 2023 
& 83{,}598 (19{,}885) 
& 81{,}237 (19{,}708) 
& 81{,}257 (19{,}892)\\

\midrule
\multicolumn{4}{l}{\textbf{Categorical variables}}\\
\midrule

\textbf{Age recode with $<$1 year olds and 90+} &&&\\
\hspace{3mm}15--40 
& 67{,}316 (14\%) 
& 12{,}995 (16\%) 
& 3{,}488 (2.3\%)\\
\hspace{3mm}41--69 
& 330{,}445 (70\%) 
& 46{,}109 (56\%) 
& 56{,}052 (38\%)\\
\hspace{3mm}70+ 
& 74{,}141 (16\%) 
& 22{,}532 (28\%) 
& 88{,}925 (60\%)\\

\addlinespace[0.4em]

\textbf{Race and origin recode} &&&\\
\hspace{3mm}Hispanic (All Races) 
& 54{,}667 (12\%) 
& 9{,}151 (11\%) 
& 10{,}499 (7.1\%)\\
\hspace{3mm}Non-Hispanic American Indian/Alaska Native 
& 1{,}956 (0.4\%) 
& 413 (0.5\%) 
& 641 (0.4\%)\\
\hspace{3mm}Non-Hispanic Asian or Pacific Islander 
& 45{,}522 (9.6\%) 
& 5{,}260 (6.4\%) 
& 7{,}458 (5.0\%)\\
\hspace{3mm}Non-Hispanic Black 
& 40{,}017 (8.5\%) 
& 11{,}215 (14\%) 
& 13{,}305 (9.0\%)\\
\hspace{3mm}Non-Hispanic White 
& 329{,}740 (70\%) 
& 55{,}597 (68\%) 
& 116{,}562 (79\%)\\

\addlinespace[0.4em]

\textbf{Laterality} &&&\\
\hspace{3mm}Left - origin of primary 
& 238{,}304 (50\%) 
& 42{,}043 (52\%) 
& 75{,}804 (51\%)\\
\hspace{3mm}Right - origin of primary 
& 233{,}598 (50\%) 
& 39{,}593 (48\%) 
& 72{,}661 (49\%)\\

\addlinespace[0.4em]

\textbf{Summary stage 2000 (1998--2017)} &&&\\
\hspace{3mm}Distant 
& 2{,}542 (0.5\%) 
& 7{,}265 (8.9\%) 
& 1{,}478 (1.0\%)\\
\hspace{3mm}Localized 
& 334{,}754 (71\%) 
& 26{,}587 (33\%) 
& 101{,}753 (69\%)\\
\hspace{3mm}Regional 
& 134{,}606 (29\%) 
& 47{,}784 (59\%) 
& 45{,}234 (30\%)\\

\addlinespace[0.4em]

\textbf{Rural-Urban Continuum Code} &&&\\
\hspace{3mm}1 
& 293{,}279 (62\%) 
& 48{,}728 (60\%) 
& 83{,}371 (56\%)\\
\hspace{3mm}2 
& 99{,}270 (21\%) 
& 16{,}579 (20\%) 
& 31{,}237 (21\%)\\
\hspace{3mm}3 
& 34{,}123 (7.2\%) 
& 6{,}829 (8.4\%) 
& 13{,}857 (9.3\%)\\
\hspace{3mm}4 
& 27{,}054 (5.7\%) 
& 5{,}734 (7.0\%) 
& 11{,}896 (8.0\%)\\
\hspace{3mm}5 
& 18{,}176 (3.9\%) 
& 3{,}766 (4.6\%) 
& 8{,}104 (5.5\%)\\

\addlinespace[0.4em]

\textbf{PR Status Recode Breast Cancer (1990+)} &&&\\
\hspace{3mm}0 
& 125{,}131 (27\%) 
& 35{,}880 (44\%) 
& 42{,}458 (29\%)\\
\hspace{3mm}1 
& 346{,}771 (73\%) 
& 45{,}756 (56\%) 
& 106{,}007 (71\%)\\

\addlinespace[0.4em]

\textbf{ER Status Recode Breast Cancer (1990+)} &&&\\
\hspace{3mm}0 
& 78{,}732 (17\%) 
& 24{,}930 (31\%) 
& 23{,}909 (16\%)\\
\hspace{3mm}1 
& 393{,}170 (83\%) 
& 56{,}706 (69\%) 
& 124{,}556 (84\%)\\

\addlinespace[0.4em]

\textbf{RX Summ--Surg Prim Site (1998+)} &&&\\
\hspace{3mm}0 
& 2{,}144 (0.5\%) 
& 1{,}833 (2.2\%) 
& 840 (0.6\%)\\
\hspace{3mm}1 
& 469{,}758 (100\%) 
& 79{,}803 (98\%) 
& 147{,}625 (99\%)\\

\addlinespace[0.4em]

\textbf{Married} &&&\\
\hspace{3mm}Married 
& 306{,}104 (65\%) 
& 43{,}830 (54\%) 
& 68{,}360 (46\%)\\
\hspace{3mm}Not married 
& 165{,}798 (35\%) 
& 37{,}806 (46\%) 
& 80{,}105 (54\%)\\

\end{longtable}

\subsubsection{Datasets Time Grid}

\begin{table}[H]
\centering
\caption{Time grid characteristics for benchmarked datasets. ($^*$) Marks when the initial time $T_0 = 0$, which is pre-processed automatically when creating the cross-validation splits into $T_0 = 0.01$ to avoid issues downstream.}
\label{datasets_time_grid}
\begin{tabular}{lccccc}
\toprule
Dataset & Unit & $T_0$ & $T^{*}$ & $|\mathcal{T}|$ & $\Delta t$\\
\midrule
PBC & Months & 1.35 & 136.35 & 10 & 15  \\
METABRIC & Months & 0.1 & 255.1 & 18 & 15 \\
FRAMINGHAM ($^*$) & Months & 0.01 & 240.01 & 9 & 30 \\
HD & Months & 0.003 & 30.003 & 11 & 3 \\
SEER & Months & 0.01 & 240 & 13 & 20 \\

\bottomrule
\end{tabular}
\end{table}

\end{document}